%% file: main.tex
\pdfoutput=1
\documentclass[11pt]{article}
\usepackage[margin=1.4in]{geometry}
\usepackage{blindtext}

\usepackage{booktabs}
\usepackage{multirow}
\usepackage{caption}
\usepackage{floatrow}
\usepackage[table]{xcolor}

\usepackage{ulem}

\definecolor{darkgreen}{rgb}{0.0, 0.4, 0.0}

\usepackage{amssymb}
\usepackage{natbib}
\usepackage[ddmmyyyy]{datetime}
\bibpunct{(}{)}{;}{a}{,}{,}

\usepackage[utf8]{inputenc} 
\usepackage[T1]{fontenc}    
\usepackage{url}            
\usepackage{booktabs}       
\usepackage{amsfonts}       
\usepackage{nicefrac}       
\usepackage{microtype}      
\usepackage{xcolor}         
\usepackage{makecell}
\usepackage{adjustbox}
\usepackage{rotating}
\usepackage{tabularx}
\usepackage{multirow}
\usepackage{pifont}
\usepackage{float}
\usepackage{amsmath}
\usepackage{graphicx}
\usepackage{grffile}  
\usepackage{bmpsize}
\usepackage{subcaption}

\usepackage{etoolbox,refcount}
\usepackage{multicol}

\newcounter{countitems}
\newcounter{nextitemizecount}
\newcommand{\setupcountitems}{%
  \stepcounter{nextitemizecount}%
  \setcounter{countitems}{0}%
  \preto\item{\stepcounter{countitems}}%
}
\makeatletter
\newcommand{\computecountitems}{%
  \edef\@currentlabel{\number\c@countitems}%
  \label{countitems@\number\numexpr\value{nextitemizecount}-1\relax}%
}
\newcommand{\nextitemizecount}{%
  \getrefnumber{countitems@\number\c@nextitemizecount}%
}
\newcommand{\previtemizecount}{%
  \getrefnumber{countitems@\number\numexpr\value{nextitemizecount}-1\relax}%
}
\makeatother    
\newenvironment{AutoMultiColItemize}{%
\ifnumcomp{\nextitemizecount}{>}{3}{\begin{multicols}{2}}{}%
\setupcountitems\begin{itemize}}%
{\end{itemize}%
\unskip\computecountitems\ifnumcomp{\previtemizecount}{>}{3}{\end{multicols}}{}}

\usepackage{tcolorbox}
\usepackage{soul}
\definecolor{highlighteryellow}{RGB}{255,255,102}
\definecolor{highlightergreen}{RGB}{144,238,144} 
\definecolor{highlighterred}{RGB}{255,178,178}
\definecolor{highlighterblue}{RGB}{135, 206, 235}

\usepackage{arydshln}

\usepackage{hyperref}       
\hypersetup{
  colorlinks   = true, 
  urlcolor     = purple, 
  linkcolor    = blue, 
  citecolor    = red, 
  pdfencoding  = auto, 
  unicode      = true  
}
\newcommand{\cmark}{\ding{51}}%
\newcommand{\xmark}{\ding{55}}%
\newcommand*\rot{\rotatebox{90}}

\makeatletter
\renewcommand{\@makefntext}[1]{%
  \parindent 1em%
  \noindent
  \hb@xt@1.8em{\hss\@makefnmark\ }#1%
}
\makeatother

\renewcommand{\cite}[1]{\citep{#1}}

\newcommand{\SMA}{\textbf{SMA}}

\usepackage{lastpage}

\makeatletter
\newcommand{\@authorone}{Romain Mussard$^*$}
\newcommand{\@authoremailone}{romain.mussard@univ-rouen.fr}
\newcommand{\@authoraffilone}{LITIS UR 4108, Univ Rouen Normandie, F-76000 Rouen, France}

\newcommand{\@authortwo}{Aurélien Gauffre$^*$}
\newcommand{\@authoremailtwo}{aurelien.gauffre@univ-grenoble-alpes.fr}
\newcommand{\@authoraffiltwo}{Université Grenoble Alpes, CNRS, LIG, 38400 Saint-Martin-d'Hères, France}

\newcommand{\@authorthree}{Ihsan Ullah}
\newcommand{\@authoremailthree}{ihsan.ullah@chalearn.org}
\newcommand{\@authoraffilthree}{ChaLearn, California, USA}

\newcommand{\@authorfour}{Thanh Gia Hieu Khuong}
\newcommand{\@authoremailfour}{thanh-gia-hieu.khuong@universite-paris-saclay.fr}
\newcommand{\@authoraffilfour}{TAU, LISN, Université Paris-Saclay, 91190 Gif-sur-Yvette, France}

\newcommand{\@authorfive}{Massih-Reza Amini}
\newcommand{\@authoremailfive}{massih-reza.amini@univ-grenoble-alpes.fr}
\newcommand{\@authoraffilfive}{Université Grenoble Alpes, CNRS, LIG, 38400 Saint-Martin-d'Hères, France}

\newcommand{\@authorsix}{Isabelle Guyon}
\newcommand{\@authoremailsix}{guyon@chalearn.org}
\newcommand{\@authoraffilsix}{Université Paris-Saclay, Gif-sur-Yvette, France and ChaLearn, California, USA}

\newcommand{\@authorseven}{Lisheng Sun-Hosoya}
\newcommand{\@authoremailseven}{lisheng.sun@universite-paris-saclay.fr}
\newcommand{\@authoraffilseven}{TAU, LISN, Université Paris-Saclay, 91190 Gif-sur-Yvette, France}

\renewcommand{\@maketitle}{%
  \newpage
  \vspace*{-1.5cm}%
  \begin{center}%
  \let \footnote \thanks
    {\LARGE \@title \par}%
    \vskip 0.8em%
    \lineskip .3em%
    \begin{center}%
    \normalsize\textbf{\@authorone} \\
    \small\textsc{\@authoremailone} \\
    \small\textit{\@authoraffilone}
    \par\vspace{0.05in}
    \normalsize\textbf{\@authortwo} \\
    \small\textsc{\@authoremailtwo} \\
    \small\textit{\@authoraffiltwo}
    \par\vspace{0.05in}
    \normalsize\textbf{\@authorthree} \\
    \small\textsc{\@authoremailthree} \\
    \small\textit{\@authoraffilthree}
    \par\vspace{0.05in}
    \normalsize\textbf{\@authorfour} \\
    \small\textsc{\@authoremailfour} \\
    \small\textit{\@authoraffilfour}
    \par\vspace{0.05in}
    \normalsize\textbf{\@authorfive} \\
    \small\textsc{\@authoremailfive} \\
    \small\textit{\@authoraffilfive}
    \par\vspace{0.05in}
    \normalsize\textbf{\@authorsix} \\
    \small\textsc{\@authoremailsix} \\
    \small\textit{\@authoraffilsix}
    \par\vspace{0.05in}
    \normalsize\textbf{\@authorseven} \\
    \small\textsc{\@authoremailseven} \\
    \small\textit{\@authoraffilseven}
    \end{center}%
    \vskip 0.5em%
    \@thanks
  \end{center}%
  \par
  \vskip 0.8em}
\makeatother

\begin{document}

\title{Stylized Meta-Album: Group-bias injection with style transfer to study robustness against distribution shifts}
\makeatletter
\author{%
\@authorone\thanks{\@authoraffilone. Email: \@authoremailone} \and
\@authortwo\thanks{\@authoraffiltwo. Email: \@authoremailtwo} \and
\@authorthree\thanks{\@authoraffilthree. Email: \@authoremailthree} \and
\@authorfour\thanks{\@authoraffilfour. Email: \@authoremailfour} \and
\@authorfive\thanks{\@authoraffilfive. Email: \@authoremailfive} \and
\@authorsix\thanks{\@authoraffilsix. Email: \@authoremailsix} \and
\@authorseven\thanks{\@authoraffilseven. Email: \@authoremailseven}
}
\makeatother
\maketitle
\def\thefootnote{*}\footnotetext{These authors contributed equally to this work.}\def\thefootnote{\arabic{footnote}}

\input{sections/abstract}

\vskip 0.2in

\input{sections/introduction}

\input{sections/related_work}

\input{sections/contributions_and_recommended_use}

\input{sections/StylizedMetaAlbum}

\input{sections/benchmarks}

\input{sections/domain_adaptation}
\input{sections/discussion_and_limitations}
\input{sections/discussion}

\section*{Acknowledgments and Disclosure of Funding}
This work originated as a student project in the M1 class ``Creation of an AI Challenge'' offered at University Paris Saclay, edition 2023. We extend our gratitude to all students of that class for their diligent contributions, with special thanks to Haolin Chen, Mahdi Ranjbar, Boubacar Sow, Ayoub Hammal, Quentin Le Tellier, Benedictus Kent Rachmat, Alex-Răzvan Ispas, and Junior Cedric Tonga for their invaluable efforts in data preparation. 
We are thankful to Birhanu Hailu Belay, Gabriel Lauzzana, Ahmad Nasser, and Sergio Escalera for coaching students.

We also received useful input from many members of the TAU team of the LISN laboratory. 
We would like to thank Dustin Carrión-Ojeda for his useful input, Adrien Pavao for his help in configuring GPUs for the experiments, Anne-Catherine Letournel and Obada Haddad for setting up the dataset backup bucket, and Romain Egele for introducing us to DeepHyper.
We are grateful to Pieter Gijsbers and Joaquin Vanschoren from OpenML for the help in hosting Stylized Meta-Album on their platform. 

Some of the computations presented in this paper were performed using the GRICAD infrastructure (https://gricad.univ-grenoble-alpes.fr), which is supported by Grenoble research communities.

This work was supported by ChaLearn, the ANR (Agence Nationale de la Recherche, National Agency for Research) under AI chair of excellence HUMANIA, grant number ANR-19-CHIA-0022, and TAILOR EU Horizon 2020 grant 952215.

This project was provided with computing HPC and storage resources by GENCI at IDRIS thanks to the grant 20XX-AD011016658 on the supercomputer Jean Zay's A100/H100 partition.

\vskip 0.2in
\bibliography{bibliography.bib}

\appendix
\clearpage
\input{appendices/datasheet}

\clearpage
\input{appendices/glossary}

\clearpage
\input{appendices/style-dataset}

\clearpage
\input{appendices/quality-control}

\clearpage
\input{appendices/benchmark-fairness}

\clearpage
\input{appendices/UDA-detailed-results}

\clearpage
\input{appendices/license}

\clearpage
\input{appendices/links}

\end{document}

%% file: sections/abstract.tex
\begin{abstract}
We introduce \textbf{Stylized Meta-Album} (SMA), a new image classification meta-dataset comprising 24 datasets (12 content datasets, and 12 stylized datasets), designed to advance studies on out-of-distribution (OOD) generalization and related topics. Created using style transfer techniques from 12 subject classification datasets, SMA provides a diverse and extensive set of 4800 groups, combining various subjects (objects, plants, animals, human actions, textures) with multiple styles. 
SMA enables flexible control over groups and classes, allowing us to configure datasets to reflect diverse benchmark scenarios. While ideally, data collection would capture extensive group diversity, practical constraints often make this infeasible. SMA addresses this by enabling large and configurable group structures through flexible control over styles, subject classes, and domains—allowing datasets to reflect a wide range of real-world benchmark scenarios. This design not only expands group and class diversity, but also opens new methodological directions for evaluating model performance across diverse group and domain configurations—including scenarios with many minority groups, varying group imbalance, and complex domain shifts—and for studying fairness, robustness, and adaptation under a broader range of realistic conditions.
To demonstrate SMA's effectiveness, we implemented two benchmarks: (1) a novel OOD generalization and group fairness benchmark leveraging SMA's domain, class, and group diversity to evaluate existing benchmarks. Our findings reveal that while simple balancing and algorithms utilizing group information remain competitive as claimed in previous benchmarks, increasing group diversity significantly impacts fairness, altering the superiority and relative rankings of algorithms. We also propose to use \textit{Top-M worst group accuracy} as a new hyperparameter tuning metric, demonstrating broader fairness during optimization and delivering better final worst-group accuracy for larger group diversity. (2) An unsupervised domain adaptation (UDA) benchmark utilizing SMA's group diversity to evaluate UDA algorithms across more scenarios, offering a more comprehensive benchmark with lower error bars (reduced by 73\% and 28\% in closed-set setting and UniDA setting, respectively) compared to existing efforts. These use cases highlight SMA's potential to significantly impact the outcomes of conventional benchmarks. 

\end{abstract}

%% file: sections/introduction.tex
\begin{figure}[ht]
    \centering
    \includegraphics[width=\textwidth]{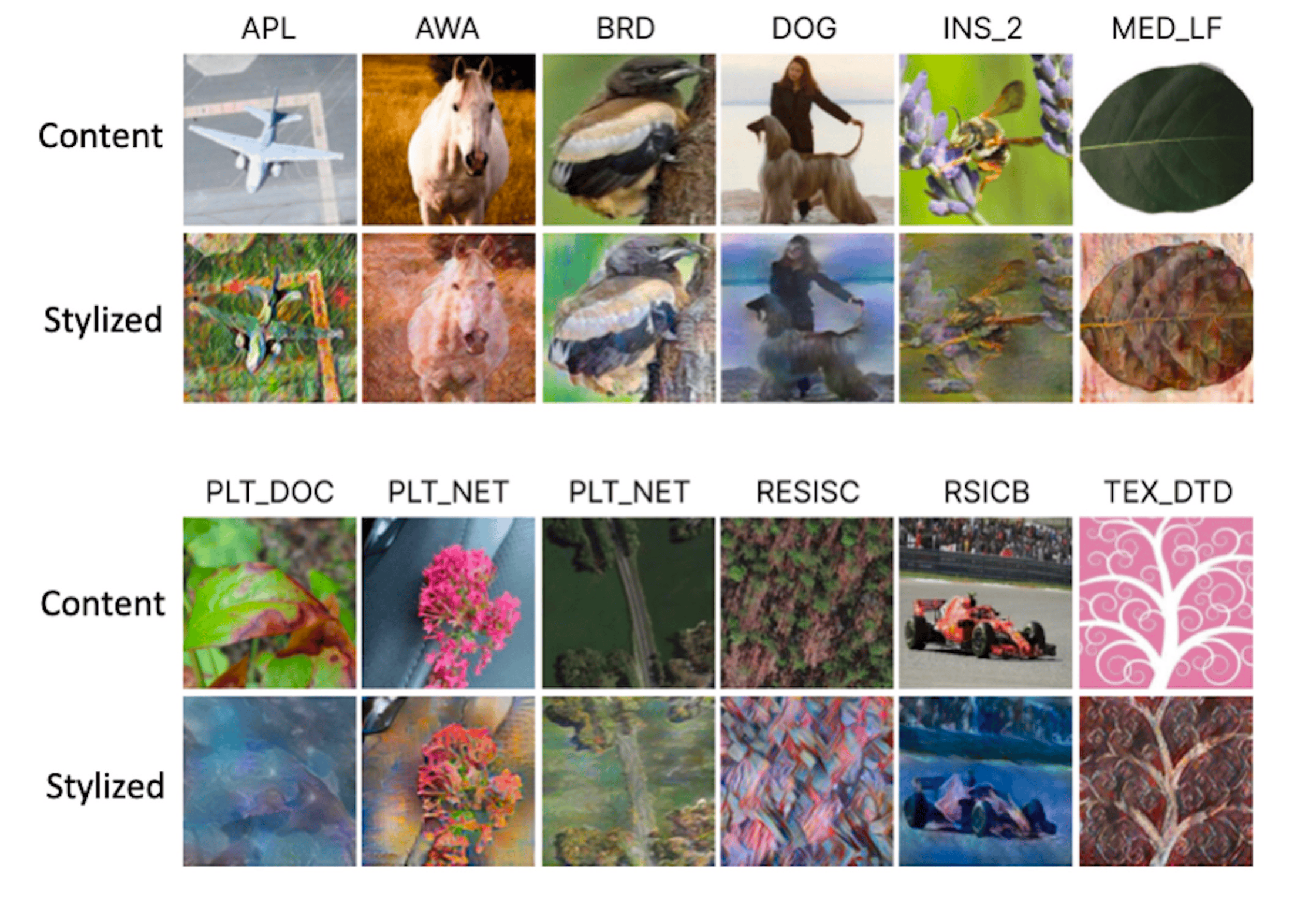}
    \caption{Stylized Meta-Album sample images. We show one content image and one stylized image for each dataset (content image stylized by a specific style image). Each column represents one dataset, ordered as in Table \ref{tab:datasets-summary}. The generation process is shown in Figure \ref{fig:SMA-flow}.}
    \label{fig:sample_images}

\end{figure}

\section{Introduction} 
\label{sec:introduction}

Predictive modeling problems in machine learning often involve data generative models of the form \(X = f(Y, Z, \epsilon)\).
Here, \(Y\) represents the target value to be predicted, \(Z\) is a nuisance factor (which may or may not be observed), and $\epsilon$ denotes unknown random noise. These problems can be challenging due to various reasons. Firstly, a change in the distribution of \(Z\) between the training and test stages, known as distribution shift, can pose difficulties. Additionally, \(Z\) may be known or unknown during training and/or testing. For instance, in fairness problems where \(Z\) represents protected variables, it may be prohibited to use them during testing. Another challenge arises when \(Z\) is correlated or dependent on \(Y\), resulting in "spurious" correlations or correlated noise. Moreover, \(Y\) and \(Z\) can be intertwined or entangled within \(X\). Furthermore, the availability of features representing \(Y\) and \(Z\) can vary. Any of the aforementioned factors, individually or in combination, can lead to performance degradation during model deployment.

To explore these complexities thoroughly, it is essential to utilize datasets that provide not only sufficient variability in \(Y\) and \(Z\) but also allow systematic control over their relationship. Several datasets have been developed for such analyses, particularly in computer vision. Examples include the Waterbird dataset, where \(Y\) represents bird species ("waterbird" or "land bird") and \(Z\) represents the background ("water" or "land"), and the CelebA dataset, where \(Y\) represents gender and \(Z\) represents hair color (see Section \ref{sec:related-work} for a detailed review). However, these datasets are limited by the small number of \(Z\) categories they encompass, which might influence benchmark outcomes compared to datasets with greater \(Z\) variability.

Addressing these limitations, we introduce the \textbf{Stylized Meta-Album (SMA)}, an extensive meta-dataset for computer vision derived from 12 subject classification datasets within the "Meta-Album" \citep{meta-album}. In SMA, \(X\)  denotes the pixel values of the input images. \(Y\) refers to the labels or target values associated with these images, encompassing categories such as objects, plants, animals, and human actions, collectively called "subjects". \(Z\) represents the various styles of the input images, such as different painting styles and natural scenes. 

The SMA dataset comprises 20 style classes, each with 40 images, paired with 20 content classes from each Meta-Album dataset, with 50 to 5000 images per class. Using style transfer techniques, we generate a "stylized dataset" for each subject dataset by pairing one image from each content class with one image from each style class, resulting in 400 unique groups per dataset, forming 4800 groups across the entire SMA. This provides substantial variability in both \(Z\) and \(Y\), facilitating straightforward re-sampling with respect to both. SMA's flexibility and modular nature make it highly adaptable for systematic experiments and future expansions by integrating more style and content images.

\begin{figure}[ht]
    \centering
    \includegraphics[width=0.8\linewidth]{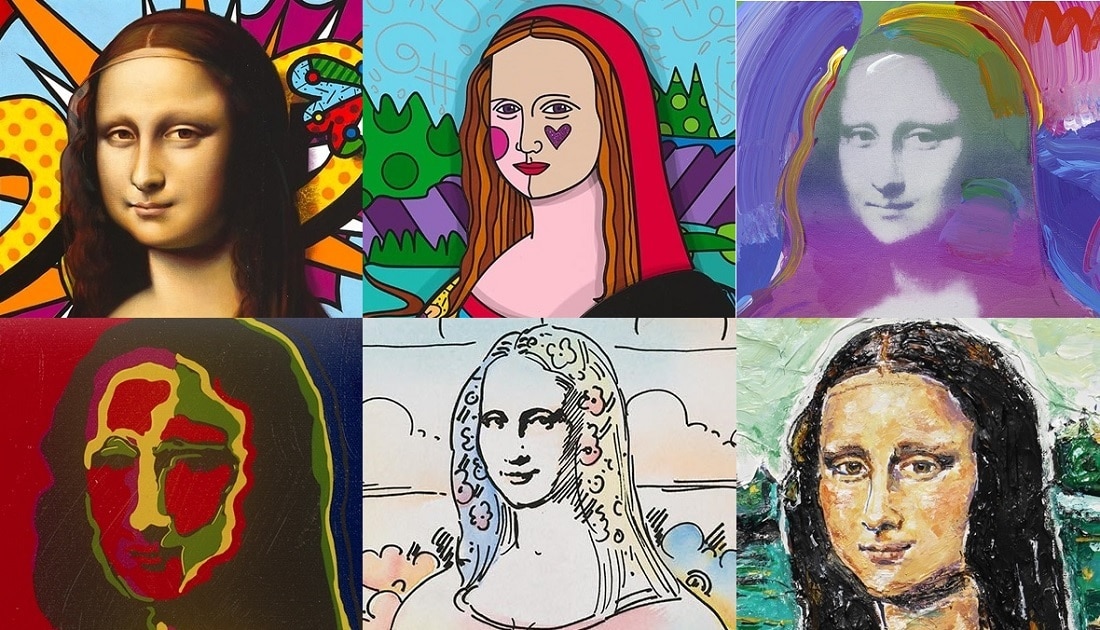}
    \caption{Variations in artistic style (Source: \href{https://www.parkwestgallery.com/six-different-artists-da-vinci-mona-lisa/}{parkwestgallery.com}).}
    \label{fig:moma-lisa}
\end{figure}

\begin{figure}[ht]
    \centering
    \includegraphics[width=0.8\linewidth]{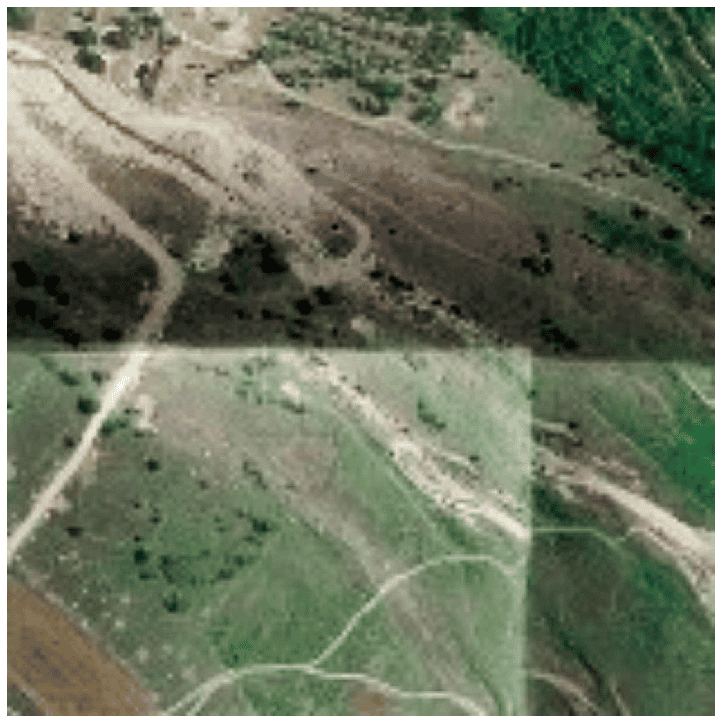}
    \caption{Remote sensing (Source: \href{https://www.mdpi.com/2072-4292/16/2/309}{StallStitch}).}
    \label{fig:remote}
\end{figure}

\begin{figure}[ht]
    \centering
    \includegraphics[width=0.8\linewidth]{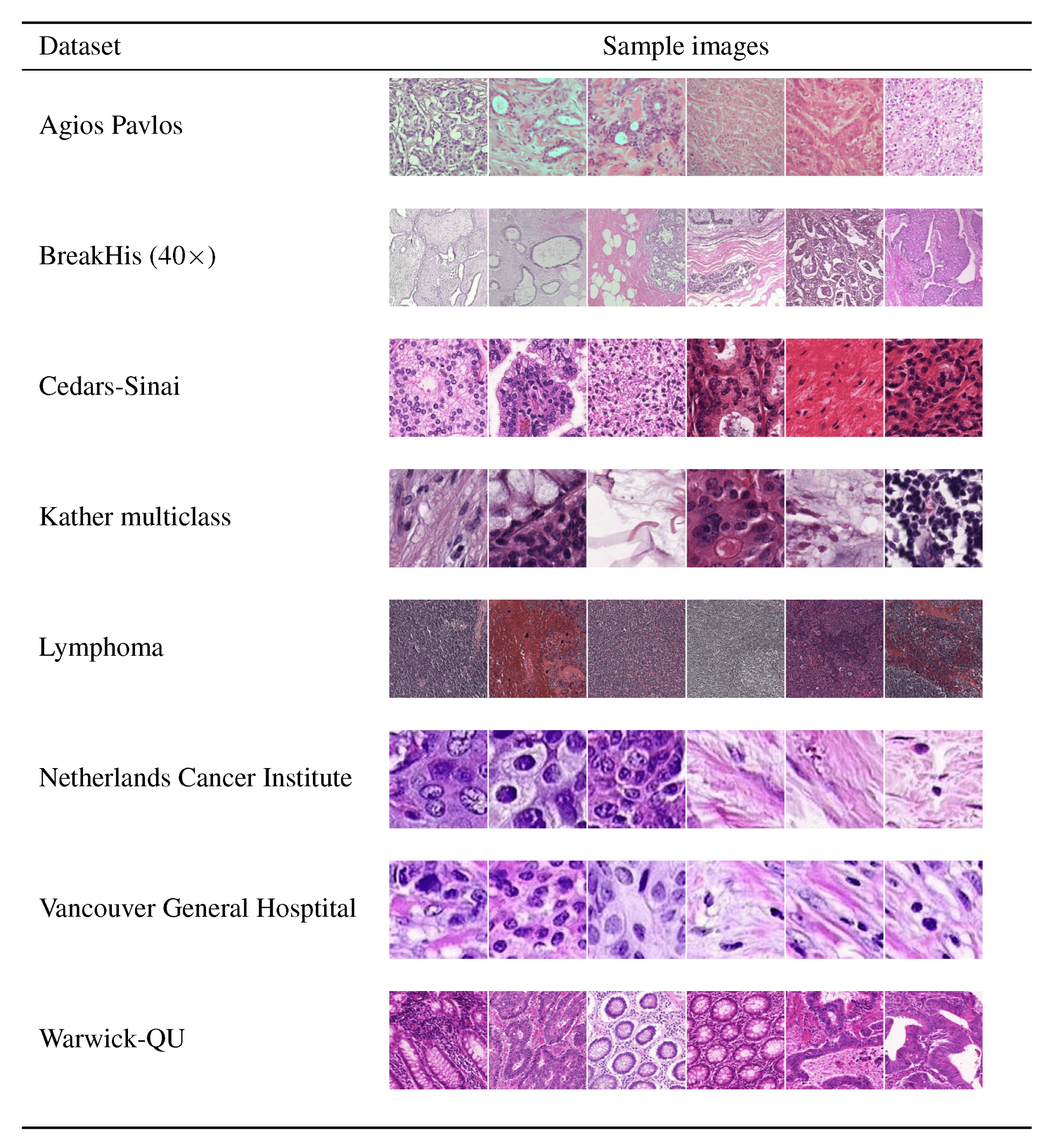}
    \caption{Remote sensing (Source: \href{https://www.mdpi.com/2072-6694/12/11/3337}{MDPI}).}
    \label{fig:bio-imaging}
\end{figure}

The choice of using ``style transfer" to trigger a distribution shift is justified as follows. Changes in style cause variations in images that alter secondary features such as color palette, texture, delineation (altogether representing the effect of $Z$). Such changes modify the appearance of an image $X$), without altering it sufficiently to prevent recognition of the main subject matter (e.g., if the subject matter in animal recognition, the animals remain recognizable). However, the introduces or modified features, which are not essential, may leak information about $Y$ by design, if sampled in a manner that certain chosen styles correlate with certain classes (e.g., wild animals associated with warm color palettes and domestic animals with cold color palettes). Changes in style occur naturally in art. The portrait of the same subject performed by different artists (Figure \ref{fig:moma-lisa}) can appear very different due to style changes and yet the person remains distinctly recognizable. The style $Z$ may leak information about the target $Y$, if, for example, the problem is to classify paintings as landscapes, portraits, and still lives, but the training data contains in category $Y$ predominantly a certain style $Z$ (e.g., more impressionist landscapes, more hyper-realist portraits, and more cubist still-lives).
In other domains, shifts in distribution that resemble changes in artistic style occur naturally. For example, in areal imagery, changes in sensors and recording conditions (time of day, season) may alter the appearance of a region, independent of its actual geography (Figure \ref{fig:remote}). Here the recording conditions play the role of $Z$. The data are biased if, for example, the problem is to classify urban areas {\em vs.} countryside (target $Y$) and urban areas are photographed in summer while countryside in winter (resulting in different lighting conditions, which may be wrongly used to predict $Y$, as opposed to geographic features). Other examples occur in biological or medical imagery (Figure \ref{fig:bio-imaging}). In that case, what plays the role of the style $Z$ results from changes in reagents, staining conditions, preparation procedure and acquisition system. Getting rid of such shifts in distribution (referred to as ``color deconvolution'') and is a major problem to avoid spurious correlations between data sources (e.g., hospitals where the data were recorded) and targeted diseases ($Y$variable). While, ideally, it would be preferable to use data plagued with natural variations to study robustness against distribution shifts, the sources of such variations are usually not controllable and/or unknown. Thus, in an effort to make available data amenable to conducting controlled benchmarks, we use the subterfuge of transferring style to mimic naturally occurring variations.

The paper is organized as follows: Section \ref{sec:introduction} reviews related datasets and benchmarks, comparing them with SMA and outlining its potential use cases. Section \ref{sec:SMA} details the generation, release, and licensing of SMA. Section \ref{sec:Benchmarks-using-SMA} presents use cases in group fairness and domain adaptation, showing how variability in 
\(Z\) can reshape benchmark conclusions. Section \ref{sec:Conclusion}
 discusses limitations, broader impact, and future directions. The datasheet for SMA is in Appendix \ref{appendix:datasheet}. For clarity, a glossary of key terms and concepts, including definitions of out-of-distribution (OOD) generalization, group fairness, robustness, unsupervised domain adaptation (UDA), meta-dataset, dataset, content dataset, style dataset, stylized dataset, and used metrics,  is included in Appendix \ref{appendix:glossary}.

SMA datasets are available on the SMA Website\footnote{SMA Website: \url{https://stylized-meta-album.github.io/}}. All experiment code is in the SMA GitHub Repository\footnote{SMA GitHub Repository: \url{https://github.com/ihsaan-ullah/stylized-meta-album}}.

%% file: sections/related_work.tex
\subsection{Related work} \label{sec:related-work}

\input{tables/comparison-of-out-of-distribution-benchmarks}

As discussed previously, challenges in prediction tasks often arise from the violation of the assumption that training and test sets are drawn from the same distribution. This distributional shift can occur in various dimensions of \(Z\) (nuisance factors, also called attributes, domains, sources, or styles), which might subsequently influence \(X\) (the input—pixels in images, in the context of computer vision) and induce a shift in \(X\). Moreover, distributional shifts can also occur in \(Y\) (prediction targets). To address these challenges, benchmark datasets for out-of-distribution generalization and multi-source/domain adaptation have been developed. 
These are sometimes collectively called out-of-distribution benchmark datasets \citep{liu2021towards}. Here, we categorize them according to their principal use cases.\\

\textbf{OOD generalization benchmarks:} Out-of-distribution (OOD) generalization refers to the capability of a model to accurately predict on data that vary in distribution from the training set. A computer vision OOD generalization benchmark can test this capability by introducing variations in background or style ($Z$ in previous sections) to original images (whose targets denoted as $Y$), thereby creating spurious correlations between $Y$ and $Z$ and varying the distribution of $Z$ between training and test stages.

Generally, these types of benchmarks exhibit large class and group imbalance \cite{idrissi2022simple}. The first popular benchmark was CelebA \cite{CelebA}, which contains spurious correlations between hair color and gender. However, lately, the most used benchmark of this type is Waterbirds \cite{Waterbirds}, which was artificially created by cutting images of birds from the Caltech-UCSD Birds-200-2011 (CUB) dataset \cite{CUB2011} and placing them on top of different backgrounds from the Places dataset \cite{Places}. Thus, this dataset presents a correlation between the type of bird (waterbird or land bird) and the type of background (water or land). Within this framework, our SMA highlights the spurious correlation between the content and style of a stylized image. Other popular benchmarks for this type of problem, but for natural language processing applications, are MultiNLI \cite{MultiNLI} and CivilComments \cite{CivilComments}. Recent studies \cite{li2023identification, li2024learning, gao2023out} have shown that the optimal selection of data or transformations can significantly enhance data augmentation, leading to improved overall performance and fairness of deep learning models. The use of data augmentation to improve fairness aligns these methods with our approach. In addition, these methods combined with style transfer could present possible leads to improve the SMA benchmark and more globally to improve fairness and robustness in deep learning models.

\textbf{Multi-source benchmarks:} This type of benchmark provides examples of classes influenced by multiple nuisance factors (\(Z\)), referred to as sources. For instance, within the class "bird", the dataset may contain images from different sources/styles such as photo, clipart, sketch, painting, and others. Consequently, these benchmarks are predominantly used for domain-adaptation studies. Domain adaptation refers to the process of adapting a model trained on one domain (source domain) to generalize to a different, but related domain (target domain), despite a distributional shift between the source and target domains. It is important to note that in the context of domain adaptation, \(Z\), or the source, is commonly called the domain. The term "source" is typically used to distinguish between the source domain where the model is trained and the target domain where the model will be tested.

The first well-known dataset in this category was Office-Home \cite{OfficeHome}, which includes 65 classes from 4 different sources (art, clipart, product, photo). Other benchmarks that compile images from multiple sources include PACS \cite{PACS}, DomainNet \cite{DomainNet}, and ImageNet-R \cite{ImageNet-R}. An alternative approach for creating multi-source datasets involves applying style transfer techniques to an existing dataset to generate various versions of the data. Stylized-ImageNet \cite{Geirhos2019} utilized this method by applying AdaIN style transfer \cite{AdaIN} to modify the original ImageNet \cite{ImageNet} images into the styles of various paintings. However, this dataset does not qualify as multi-source since each image is stylized only once. In contrast, in SMA, we employ the AdaIN style transfer technique to blend images from 20 distinct style classes with multi-domain images from the Meta-Album (see Section \ref{sub-sec:style-transfer}). This approach results in the creation of a multi-source, multi-domain benchmark dataset. 

Similarly, WILDS \cite{wilds} is a benchmark comprising 10 multi-domain datasets, including computer vision (CV), text, and graphs datasets. WILDS datasets capture spurious correlations ‘in the wild’, arising naturally from real-world data collection biases (e.g., in CAMELYON17-WILDS, tissue slides from different hospitals exhibit staining variations; in WILDCAM2020-WILDS, camera trap variations introduce biases). These correlations occur without intervention, offering no control over their characteristics. In contrast, SMA uses style transfer to synthetically introduce and precisely control spurious correlations, making it particularly well-suited for evaluating distribution shifts under controlled conditions.

While WILDS datasets reflect natural data variability, SMA complements this by introducing synthetic but systematic variations that are challenging to achieve in real-world datasets. Additionally, style transfer enables broader transformations than simple attribute modifications, such as changes to backgrounds or illumination. Unlike these localized alterations, style transformations affect global image features, including textures, which are known to strongly influence model decisions and often introduce spurious correlations \cite{Geirhos2019}. Together, SMA and WILDS form complementary resources for studying robustness and fairness in machine learning.

Tables~\ref{tab:comparison-of-out-of-distribution-benchmarks} presents a comparative analysis of SMA against existing benchmarks in OOD generalization and multi-source contexts, respectively. SMA distinguishes itself from other benchmarks through three primary aspects: : (1) it contains the largest number of sources (i.e. styles or domains in the context of domain-adaptation) which can be easily extended by collecting more style datasets; (2) it provides a sufficient number of images per class and per source, which facilitates studies of OOD generalization by enabling the easy creation of training sets with specific group imbalances; and (3) it inherits some advantages from Meta-Album such as the multi-domain feature and the possibility to grow the number of datasets continuously.

%% file: tables/comparison-of-out-of-distribution-benchmarks.tex
\begin{table}[t]
  \caption{\textbf{Comparison of SMA against related benchmark datasets}. The statistics are computed based on the complete datasets, including training, validation, and testing sets where applicable. It should be noted that each SMA dataset maintains the same number of classes and nuisances; however, the number of examples may vary from one dataset to another.}
  \label{tab:comparison-of-out-of-distribution-benchmarks}
  \centering
  \begin{adjustbox}{width=\linewidth}
  \begin{tabular}{l c c c c c c c c c c c}
    \toprule
    \makecell{\Large{Benchmark}} & 
    \multicolumn{1}{c}{\rot{\Large{\# of datasets}}} &
    \multicolumn{1}{c}{\rot{\makecell{\Large{\# of sources or}\\\Large{\# of nuisance factors}}}}  & 
    \rot{\Large{\# of classes}} & 
    \multicolumn{1}{c}{\rot{\Large{\# of instances}}} & 
    \multicolumn{1}{c}{\rot{\makecell{\Large{min/max instances}\\\Large{per class}}}}  & \multicolumn{1}{c}{\rot{\makecell{\Large{min/max instances}\\\Large{per group}}}} &
    \multicolumn{1}{c}{\rot{\makecell{\Large{min/max classes}\\\Large{per source}}}} &
    \multicolumn{1}{c}{\rot{\makecell{\Large{min/max instances}\\\Large{per source}}}} &
    \multicolumn{1}{c}{\rot{\makecell{\Large{multi-datasets}}}} & 
    \multicolumn{1}{c}{\rot{\makecell{\Large{multi-versions}}}} & 
    \multicolumn{1}{c}{\rot{\makecell{\Large{uniform image size}}}}
    \\
    \midrule
    \multicolumn{12}{c}{\textbf{Out-of-distribution generalization benchmarks}} \\
    \midrule
    CelebA \cite{CelebA} & 1 & 2 & 2 & $202\,599$ & $29\,983\,/\,172\,616$ & $1\,749\,/\,89\,931$ & 2/2 & $84\,434\,/\,118\,165$ & \xmark & \xmark & \cmark 
    \\
    Waterbirds \cite{Waterbirds} & 1 & 2 & 2 & $11\,788$ & $2\,663\,/\,9\,125$ & $831\,/\,6\,220$ & 2/2 & $4\,737\,/\,7\,051$ & \xmark & \xmark & \xmark
    \\
    CivilComments \cite{CivilComments} & 1 & 8 & 2 & $244\,436$ & $43\,197/201\,239$ & $1\,685/50\,581$ & 2/2 & $11\,030/58\,584$ & \xmark & \xmark & \xmark
    \\
    MultiNLI \cite{MultiNLI} & 1 & 2 & 3 & $412\,349$ & $137\,152/137\,841$ & $3\,020/134\,821$ & 3/3 & $29\,404/382\,945$ & \xmark & \xmark & \xmark
    \\
    \midrule
    \multicolumn{12}{c}{\textbf{Multi-source benchmarks}} \\
    \midrule
    Office-Home \cite{OfficeHome} & 1 & 4 & 65 & $15\,913$ & $142\,/\,363$ & 15/99 & $65\,/\,65$ & $2\,496/4\,503$ & \xmark & \xmark & \xmark 
    \\
    PACS \cite{PACS} & 1 & 4 & 7 & $9\,991$ & $943\,/\,1\,729$ & 80/816 & $7\,/\,7$ & $1\,670\,/\,3\,929$  & \xmark & \xmark & \cmark 
    \\
    DomainNet \cite{DomainNet} & 1 & 6 & 345 & $596\,006$ & $878\,/\,2\,799$ & 10/901 & $345\,/\,345$ & $48\,833\,/\,175\,327$ & \xmark & \xmark & \xmark
    \\
    ImageNet-R \cite{ImageNet-R} & 1 & 15 & 200 & $30\,000$ & $51\,/\,430$ & 1/204 & $106\,/\,200$ & $508\,/\,5\,673$ & \xmark & \xmark & \xmark 
    \\
    Office-31 \cite{saenko2010adapting} & 1 & 3 & $31$ & $4\,113$ & $68/164$ & $7/100$ & $31/31$ & $796/2\,818$ & \xmark & \xmark & \cmark 
    \\
    VisDA-C \cite{peng2017visda} & 1 & 3 & 12 & $280\,157$ & $15\,112\,/31\,235$ & $8\,079\,/17\,360$  & $12\,/\,12$ & $55\,388\,/\,152\,397$ & \xmark & \xmark & \xmark 
    \\
    \midrule
    \multicolumn{12}{c}{\textbf{Stylized Meta-Album (SMA)}} \\
    \midrule
    \textbf{SMA}\_\textit{\textbf{BRD} (Single Dataset)} & 1 & 20 & 20 & $90\,620$ & $4\,260/5\,460$ & $213/273$ & 20/20 & $4\,531/4\,531$ & \xmark & \cmark & \cmark
    \\
    \midrule
    \textbf{SMA} \textit{\textbf{Extended} (Meta-Dataset)} & 12 & 20 & 240 & $2\,859\,980$ & $1\,120/32\,900$ & $56/1\,645$ & 20/20 & $1\,395/30\,000$ & \cmark & \cmark & \cmark
    \\
    \bottomrule
  \end{tabular}
  \end{adjustbox}

\end{table}

%% file: sections/contributions_and_recommended_use.tex
\subsection{Contributions and recommended use} 

\label{sec:contributions}

\subsubsection{Contributions}

Our main contributions are:
\begin{enumerate}
    \item We introduce the Stylized Meta-Album (SMA), a \textbf{new computer vision meta-dataset} consisting of 24 uniformly formatted datasets across multiple domains. SMA is designed in a flexible, systematic, and controlled way to advance studies on out-of-distribution (OOD) generalization and related topics, such as learning under distribution shift (e.g., UDA), and addressing spurious correlations (e.g., group fairness). In particular, SMA enables users to define and manipulate groups (e.g., based on style and subject class), control the degree of distribution shift, and vary the proportions between majority and minority groups—facilitating rigorous, customizable OOD benchmarks.

    \item We provide two versions of each dataset: Mini and Extended, \textbf{catering to researchers with varying computational resources}.
    \item We conducted a comprehensive \textbf{group fairness benchmark} of state-of-the-art (SOTA) methods focusing on Group Fairness using 8 SMA datasets, leveraging SMA's extensive group diversity. Our findings show that simple balancing techniques remain competitive with growing number of groups. However, greater class and group diversity reduces bias implicitly and enhances other methods' performance. We also \textbf{propose a new optimization objective} that enhances the currently used worst-group accuracy, providing greater stability and fairness in the context of increased group diversity.
    
    \item To the best of our knowledge, we conduct the \textbf{first benchmark of SOTA algorithms in  Unsupervised Domain Adaptation (UDA) across diverse classification tasks}, evaluating performance in two settings: close-set (where the label sets of the source and target domains are identical) and UniDA (where the label sets of the source and target domains can differ). One notable challenge in UDA research is the variability in performance metrics observed in existing benchmarks, often due to their limited scenarios. This variability can make it challenging to draw reliable conclusions when comparing different algorithms. This variability is commonly measured by the standard error of the mean (SEM) and visualized as error bars in performance plots. To address this issue, we focus on five small to medium-sized SMA datasets, each with 25 scenarios, and highlight the diverse difficulty levels within the SMA datasets. This approach allows us to achieve an average reduction in the SEM by approximately 73\% (close-set) and 28\% (UniDA) compared to existing UDA datasets in the literature, ensuring better statistical significance.

    \item We provide a curated list of recommended uses for SMA, designed to inspire researchers \textbf{to leverage SMA for exploring a variety of topics that align with their interests}.
\end{enumerate}

\subsubsection{Recommended Use}

Stylized Meta-Album (SMA) is an appropriate meta-dataset for a wide variety of applications. Its core strength lies in the large and extensible number of subject-style combinations, which provide a high degree of control over distributional shifts. This makes SMA particularly well-suited for studying out-of-distribution (OOD) generalization \cite{liu2021towards} and group fairness \cite{idrissi2022simple}. Users can customize the number and definition of groups (e.g., based on style, subject class, and content domain), two-group to more complex multi-group scenarios. This flexibility enables controlled evaluations under varying degrees of domain or group imbalance—including user-defined proportions between majority and minority groups—making it ideal for probing robustness and generalization. SMA can also support research on shortcut learning\cite{geirhos2020shortcut}, domain adaptation, domain-incremental learning \cite{vandeven2019generative}, few-shot domain-incremental learning \cite{esaki2024oneshot}, and domain-incremental continual learning \cite{vandeven2019scenarios}. As a meta-dataset encompassing 12 distinct datasets, SMA is particularly conducive to investigating meta-learning with an emphasis on fairness, particularly in scenarios involving sensitive attributes \cite{zhao2021fairness}, enhancing domain adaptation via meta-learning techniques \cite{li2018learning}, and examining the impact of style/background elements as potential shortcuts in few-shot learning processes \cite{luo2021rectifying}. Furthermore, since SMA is created from \href{https://meta-album.github.io/}{Meta-Album} \cite{meta-album}, it can be used for conventional few-shot learning, meta-learning, continual learning, and transfer learning in the context of image classification. 
To illustrate its extensive utility, we present two benchmarks in Section \ref{sec:Benchmarks-using-SMA} focusing on 1) Group fairness, a specific aspect of OOD generalization, and 2) unsupervised domain adaptation. It is important to note, however, that SMA is not intended for deriving scientific findings beyond benchmarking purposes or for developing products that rely solely on this meta-dataset.


%% file: sections/StylizedMetaAlbum.tex
\section{Stylized Meta-Album (\SMA)}
\label{sec:SMA}

Style Transfer \cite{gatys2015neural} refers to a set of techniques in computer vision that aim to generate a new image by combining the content of one image with the style of another. The resulting image preserves the content of the first image while adopting the visual style of the second. A brief review of style transfer techniques is given in  Section \ref{sub-sec:style-transfer}. 
In SMA, we utilize style transfer to create stylized images by applying the style from our newly created Style Dataset to content images from the Meta-Album.


\begin{figure}[H]
    \centering
    \includegraphics[width=0.9\textwidth]{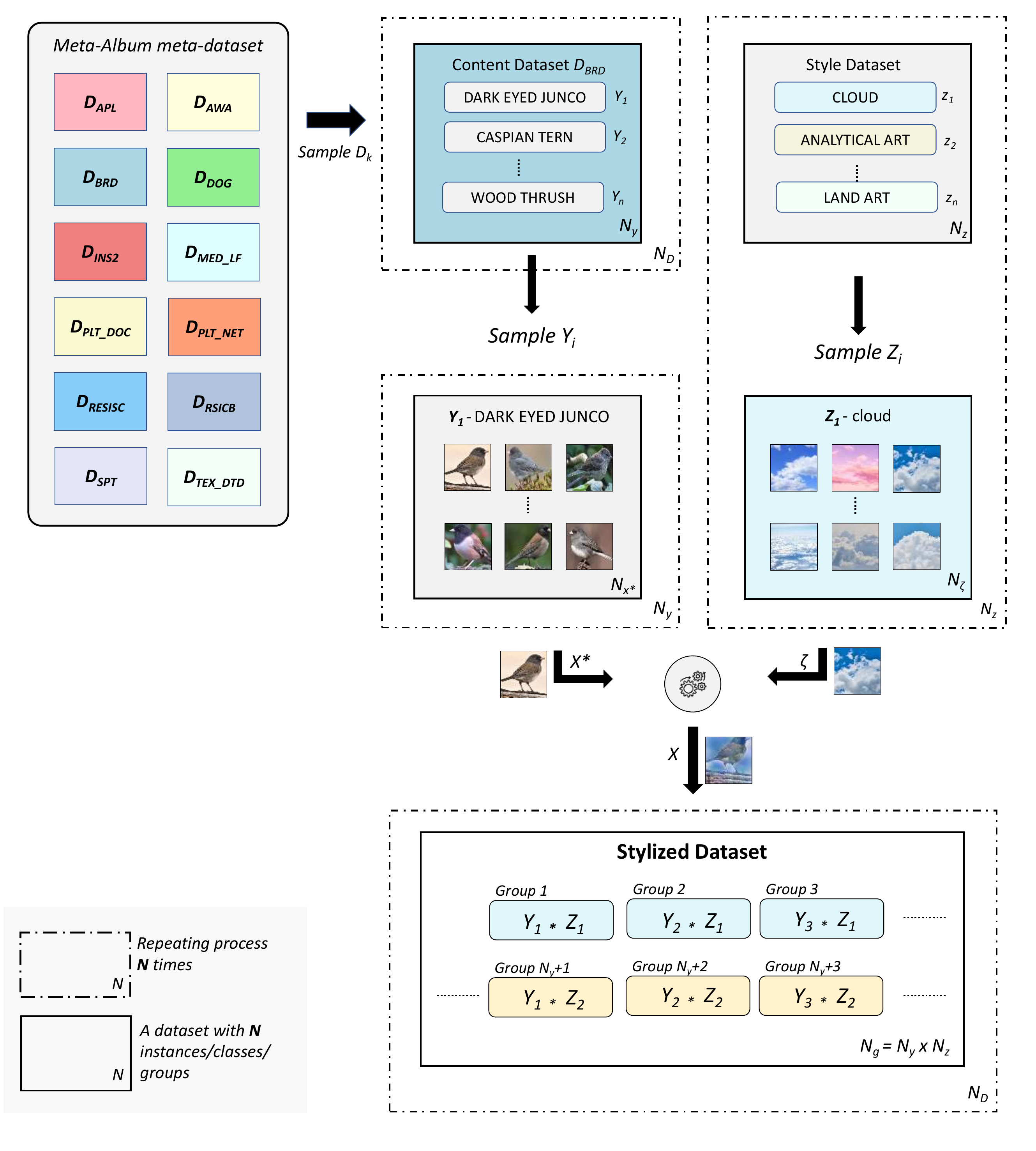}
    \caption{Stylized Meta-Album generation process}
    \label{fig:SMA-flow}
\end{figure}

As illustrated in Figure \ref{fig:SMA-flow}, the process of creating an SMA dataset is as follows:

\begin{enumerate}
    \item From the Meta-Album meta-dataset, sample a content dataset comprising \(N_y\) classes.
    \item From these \(N_y\) classes, select one class containing \(N_{x^*}\) content images.
    \item Simultaneously, select one style class from the Style Dataset, which consists of \(N_z\) style classes, each containing \(N_\zeta\) style images.
    \item Using the stylization process \(h\), generate a stylized image \(x\) by combining a content image \(x_*\) with a style image \(\zeta\), i.e., \(x = h(x_*, \zeta)\).
    \item Repeat 1-4 for all content and style classes, resulting in \(N_g = N_y \times N_z\) groups in the stylized dataset.
\end{enumerate}

The above procedure is applied to all selected \(N_D\) Meta-Album datasets, ultimately producing \(N_D\) SMA datasets. In this first version of SMA, \(N_x\) ranges from 50 to over 5,000 images, \(N_y = 20\) and \(N_z = 20\), resulting in \(N_g = 400\) groups per Meta-Album dataset. With \(N_D = 12\) Meta-Album datasets spanning various domains (including ecology, manufacturing, remote sensing, and human actions), the total number of groups is \(N_g(\text{total}) = N_g \times N_D = 4,800\). Each group represents a combination of content and style, e.g., DARK EYED JUNCO (a class of BIRD) - cloud, as shown in Figure \ref{fig:SMA-flow}, or CASPIAN TERN - analytical\_art, etc.

Stylized Meta-Album (SMA) is a meta-dataset in which each dataset consists of both stylized and content images. The content datasets are selected from the Meta-Album meta-dataset. However, Meta-Album only released preprocessed images at a resolution of $128 \times 128$, which is too small to support effective style transfer. Therefore, we reprocessed the original content images from scratch using the Meta-Album pipeline, generating images at a higher resolution of $256 \times 256$. Due to computational constraints, only a subset of the Meta-Album datasets was used as content datasets for SMA. Additionally, the use of style transfer allows for the creation of multiple stylized versions of the same content dataset. In particular, the design of the SMA datasets makes them naturally suited for tasks such as out-of-distribution learning and domain adaptation, which are beyond the scope of Meta-Album.

The composition of all 12 SMA datasets is summarized in Table \ref{tab:datasets-summary}, and Figure \ref{fig:sample_images} displays representative images from the content and stylized datasets for each SMA dataset. Detailed descriptions of the content and style datasets are provided in Sections \ref{sub-sec:content-datasets} and \ref{sub-sec:style-dataset}, respectively. Section \ref{sub-sec:style-transfer} discusses the methodology for transferring style onto content images, and Section \ref{sub-sec:release} provides instructions on SMA dataset access and its release details. Datasheet~\cite{gebru2021datasheets} for SMA datasets can be found in Appendix \ref{appendix:datasheet}.

\input{tables/data-summary-table}

\subsection{Content Datasets}
\label{sub-sec:content-datasets}

Content datasets consist of 12 \href{https://meta-album.github.io/}{Meta-Album} \cite{meta-album} datasets, selected from 8 domains: Large Animals, Small Animals, Plants, Plant Diseases, Vehicles, Manufacturing, Remote Sensing, and Human Actions. Selection focused on image quality and the quality of resulting stylized images, prioritizing datasets with a high number of images per class to support group fairness and related use cases. Images with watermarks or relying mainly on color for recognition were excluded to maintain quality and recognizability after stylization. For datasets with more than 20 classes, the top 20 classes with the most examples were retained. All images were resized to 256 $\times$ 256 pixels before stylization. For a comprehensive list of these datasets, see Table \ref{tab:datasets-summary}.

\subsection{Style Dataset}
\label{sub-sec:style-dataset}
The style dataset comprises images representing various styles, sourced from publicly accessible images on the internet using our scraping program described in the following section. This dataset includes 20 distinct classes (see Appendix \ref{appendix:style-dataset}), each containing 40 images, resulting in a total of 800 images. 

While we do not distribute the original images from the style dataset due to potential licensing restrictions, we provide metadata and the extracted feature maps of these images using our stylization process for the purpose of reproducibility. We have made every effort to ensure our use of these images adheres to the principles of fair and transformative use. Our work focuses on the generated SMA, which are significantly transformed stylized images distinct from the original source images. The datasets and benchmarks presented in this paper are solely for research and non-commercial purposes, protected by license in Appendix \ref{appendix:license}.

\subsubsection*{Data Scraping}
We curated a diverse collection of 50 keywords, comprising artistic and painting styles as well as natural scenes, with each keyword representing a distinct style. To compile our style collection, we developed a custom \textbf{Scraping Program} (code available in \href{https://github.com/ihsaan-ullah/stylized-meta-album/tree/master/style_dataset_preparation/style_image_scrapping}{our GitHub repository}) capable of downloading up to 100 images matching a reference keyword. By inputting the initial set of 50 keywords into our scraping program, we compiled a dataset of 5,000 style images, providing 100 images for each unique style. These images will undergo a quality filtering process to retain only those of high quality, as detailed in Section \ref{subsubsec:data-filtering}. For more information on the data scraping process, see Appendix \ref{appendix:style-dataset}.

\subsubsection*{Data Filtering}
\label{subsubsec:data-filtering}

Stylizing images can distort the image content arbitrarily, to the point that the main object can become non recognizable to the human eye. However, manual verification being tedious, we proceeded to put in place an automated filtering technique to select styles suitable for our purpose. Appendix \ref{appendix:style-dataset} gives more details about the data filtering process.

\subsubsection*{Quality control}
\label{sub-sub-section:quality-control}
To evaluate the quality of a style, in terms of its ability to preserve valuable information from the original content images, we compared the classification performance for stylized {\em vs.} original images using the embeddings from an EfficientNet (v2) model \cite{pmlr-v139-tan21a}, combined with a linear neural network. We propose a metric \[
Fidelity = \frac{\text{1}}{\text{N}} \sum_{i=1}^{N} \frac{\textit{Stylized Accuracy}_{i}}{\textit{Original Accuracy}_{i}}~,
\] which is the ratio between the classification performance of the original content image \textit{(Original Accuracy)} and that of the stylized image \textit{(Stylized Accuracy)}, averaged over $N$ experiments. We found that all styles chosen through our filtration process have the highest \textit{Fidelity} scores among all evaluated styles (all above 0.8). These observations serve as further confirmation of the effectiveness of our previous filtration process. We discuss the detailed quality control process in Appendix \ref{appendix:quality-control}.

\subsection{Style Transfer}
\label{sub-sec:style-transfer}

To transfer "style" to "content" images, various style transfer techniques have been developed, leveraging convolutional neural networks (CNNs) to blend the content of one image with the style of another. This technique separates style and content in the feature space of a CNN, treating style as a controlled nuisance factor $Z$ for thorough out-of-distribution analysis. Among the available methods, we focus on three key approaches to perform the style transfer.

\textbf{Neural Style} \cite{gatys2015neural}: Content representation of an image is derived from the feature responses in the higher layers of a CNN. The style representation, on the other hand, is captured by computing feature correlations across multiple layers of the same CNN. An image is then generated to minimize the distance between these two representations, enabling style transfer while preserving the content.

\textbf{Patch Based} \cite{chen2016fast}: Uses a pre-trained CNN to map content and style images into an activation space, splits them into overlapping patches, and matches each content patch with the closest style patch. Content activation patches are swapped with their corresponding style patches before reconstructing the complete content activations, enabling fast style transfer across various artistic styles. 

\textbf{AdaIN} \cite{AdaIN}: Introduces the use of an adaptive instance normalization to match the statistical properties (mean and variance) of content features to those of style features. The main goal is to offer fast style transfer over arbitrary styles. \\

\begin{figure}[ht]
    \centering
    \includegraphics[width=0.8\linewidth]{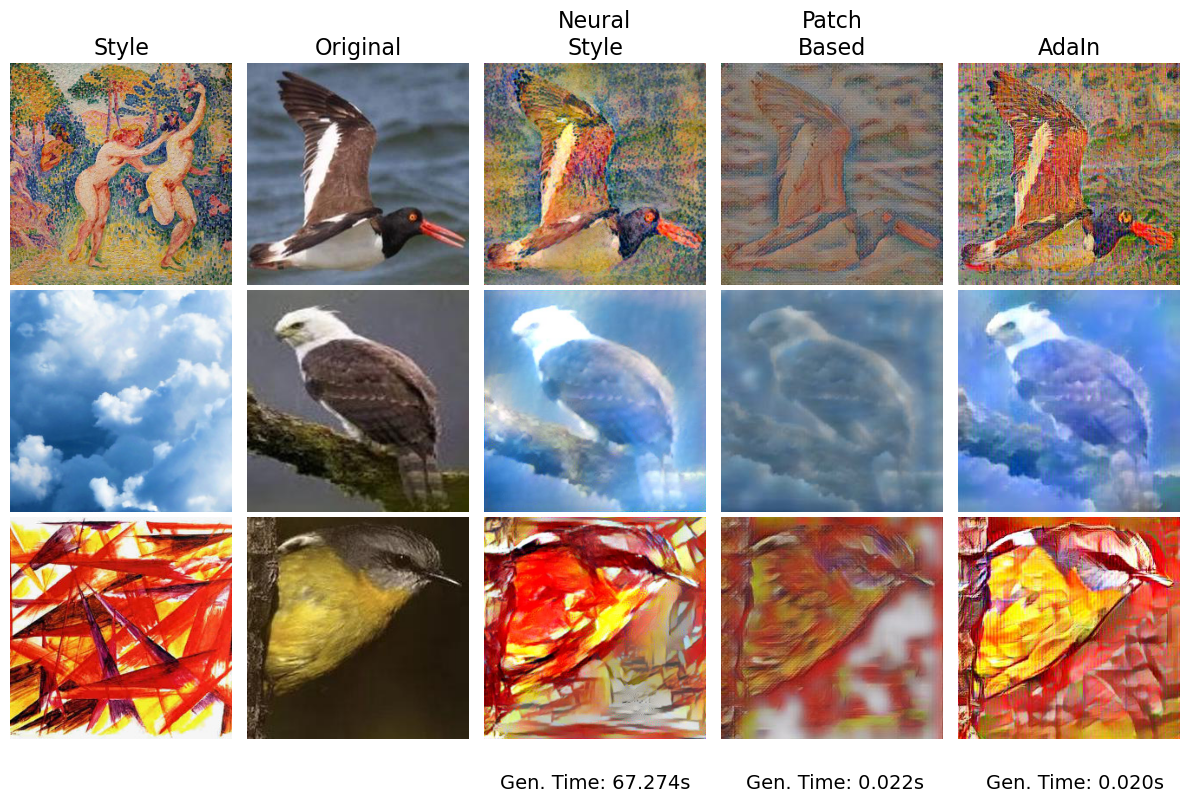}
    \caption{Style Transfer Comparison}
    \label{fig:model_beanchmark_quality}
\end{figure}

Compared to Neural Style, both AdaIN and Patch-Based Style Transfer offer faster computation during inference. AdaIN processes a style transfer in 0.020 seconds per image, while Style Swap (Patch-Based) takes 0.022 seconds per image on an NVIDIA RTX 4000 Ada Generation Laptop. In contrast, Neural Style Transfer takes over a minute per image, making it impractical for large-scale data generation despite its superior quality (Figure \ref{fig:model_beanchmark_quality}). AdaIN is the optimal choice due to its speed and higher quality images compared to Style Swap, which focuses less on the foreground. Therefore, we use AdaIN for generating the SMA datasets. For reproducibility, we adapted the original AdaIN algorithm to: 1) save the style feature maps extracted by the pretrained VGG model; and 2) use these feature maps as input, along with arbitrary content images. This adaptation allows us and other users to generate new stylized images with different content images and create new style feature maps. The adapted code is available in our GitHub repository.

\subsection{First-release}
\label{sub-sec:release}
We are releasing the first version of Stylized Meta-Album in \underline{{\bf August 2024}} on OpenML\footnote{OpenML: \url{https://www.openml.org/}} \cite{OpenML2013}. This first version consists of 12 content datasets, 12 stylized datasets, and 1 style dataset. These datasets are released with the same license as Meta-Album i.e. \href{https://creativecommons.org/licenses/by-nc/4.0/}{CC BY-NC 4.0} license. More details about license information in~Appendix \ref{appendix:license}.
We maintain a website \footnote{Stylized Meta-Album website: \url{https://stylized-meta-album.github.io/}} to host the datasets and to provide further release information. We have a dedicated GitHub repository\footnote{Stylized Meta-Album GitHub repo: \url{https://github.com/ihsaan-ullah/stylized-meta-album}} to release the code related to Stylized Meta-Album. Further, we are planning to prepare more datasets and add them to SMA. We also invite the community to contribute datasets to this meta-dataset. We will provide complete instructions for contributions on our GitHub repository.

%% file: tables/data-summary-table.tex
\begin{table}[t]
  \caption{Stylized Meta-Album: Datasets Summary (\textit{Extended} versions).}
  \label{tab:datasets-summary}
  \centering
  \begin{adjustbox}{width=\linewidth}
  \begin{tabular}{l l l c c c c}
    \toprule
    \makecell{{\bf Stylized Meta-Album}\\{\bf Dataset Name}} &
    \makecell{{\bf Meta-Album}\\{\bf Domain Name}} & 
    \makecell{{\bf Meta-Album}\\{\bf Dataset ID}} & 
    \makecell{{\bf \# Classes}\\(Content Dataset)} & 
    \makecell{{\bf \# Images}\\(Content Dataset)} & 
    \makecell{{\bf \# Classes}\\(Stylized Dataset)} & 
    \makecell{{\bf \# Images}\\(Stylized Dataset)}
    \\
    \midrule
    SMA\_APL & Vehicles & \href{https://meta-album.github.io/datasets/APL.html}{APL} & 20 & 10,205 & 400 & 204,100
    \\
    SMA\_AWA & Large Animals & \href{https://meta-album.github.io/datasets/AWA.html}{AWA} & 20 & 21,891 & 400 & 437,820
    \\
    SMA\_BRD & Small Animals & \href{https://meta-album.github.io/datasets/BRD.html}{BRD} & 20 & 4,531 & 400 & 90,620
    \\
    SMA\_DOG & Large Animals & \href{https://meta-album.github.io/datasets/DOG.html}{DOG} & 20 & 4,244 & 400 & 84,880
    \\
     SMA\_INS\_2 & Small Animals & \href{https://meta-album.github.io/datasets/INS_2.html}{INS\_2} & 20 & 30,000 & 400 & 600,000
    \\
     SMA\_MED\_LF & Plant Diseases & \href{https://meta-album.github.io/datasets/MED_LF.html}{MED\_LF} & 20 & 1,395 & 400 & 27,900
    \\
    SMA\_PLT\_DOC & Plant Diseases & \href{https://meta-album.github.io/datasets/PLT_DOC.html}{PLT\_DOC} & 20 & 2,135 & 400 & 42,620
    \\
    SMA\_PLT\_NET & Plants & \href{https://meta-album.github.io/datasets/PLT_NET.html}{PLT\_NET} & 20 & 30,000 & 400 & 600,000
    \\
    SMA\_RESISC & Remote Sensing & \href{https://meta-album.github.io/datasets/RESISC.html}{RESISC} & 20 & 14,000 & 400 & 280,000
    \\
    SMA\_RSICB & Remote Sensing & \href{https://meta-album.github.io/datasets/RSICB.html}{RSICB} & 20 & 18,691 & 400 & 373,820
    \\
    SMA\_SPT & Human Actions & \href{https://meta-album.github.io/datasets/SPT.html}{SPT} & 20 & 3,511 & 400 & 70,220
    \\
    SMA\_TEX\_DTD & Manufacturing & \href{https://meta-album.github.io/datasets/TEX_DTD.html}{TEX\_DTD} & 20 & 2,400 & 400 & 48,000
    \\
    \bottomrule
  \end{tabular}
  \end{adjustbox}
\end{table}

%% file: sections/benchmarks.tex
\section{Benchmarks using SMA}
\label{sec:Benchmarks-using-SMA}
In this section, we demonstrate the usage of SMA datasets using 2 benchmarks in  OOD generalization applied to group fairness and unsupervised domain adaptation. 

\subsection{Benchmark 1: Group Fairness}
\label{sec:Illustrative-Use-Cases}

The challenge of ensuring fairness in machine learning often involves addressing group fairness \footnote{Group fairness is related to out-of-distribution (OOD) generalization in that both aim to address distributional shifts. However, while OOD generalization targets robustness to any shift in data distribution, group fairness focuses specifically on reducing disparities across predefined groups.}, which focuses on ensuring equitable performance across predefined subgroups within a population, often defined by protected attributes such as gender, race, or age. Group fairness emphasizes reducing disparities between these known groups, such as ensuring fair treatment in loan approval for both males and females. However, acquiring real-world data that captures sensitive or protected attributes can be challenging due to privacy concerns, ethical constraints, and data availability. To address these challenges, the SMA datasets provide a synthetic environment for studying group fairness. Using style as a proxy for sensitive attributes, SMA enables systematic and ethical exploration of spurious correlations between content class Y and style Z. While styles do not fully capture the complexity of real-world attributes like gender or race, which often influence outcomes indirectly through interactions with other features, they allow researchers to evaluate how algorithms handle such correlations under controlled conditions. Variations in the distributions P(X|Y) and P(X|Z) between the training and validation/test sets enable the assessment of an algorithm’s ability to resist spurious correlations and deliver equitable predictions across subgroups during deployment.

Evaluating algorithms' ability to handle such data is crucial for ensuring equal treatment across all groups, regardless of the protected attribute Z. This is often guided by metrics such as worst-group accuracy to optimize fairness. \citet{idrissi2022simple} proposed a benchmark to compare state-of-the-art (SOTA) methods that optimize worst-group accuracy with simple methods using class/group balancing. Their findings revealed that these simple techniques achieved SOTA performance with faster training times and no additional hyperparameter tuning. However, the datasets used in this benchmark, such as Waterbirds and CelebA, were limited in the number of classes and featured only 2 or 3 styles. 

In this study, we extend this analysis using SMA datasets with up to 12 content and style classes at the same time, which offers a richer variety of domains, classes, and styles. This allows us to investigate how fairness algorithms generalize with a larger number of groups on different datasets and reassess the conclusions of the previous benchmark. Relevant real-world examples where fairness across multiple minority groups should be addressed include automated hiring tools, loan and housing application screening, and crime risk prediction — where multiple minority skin tones (e.g., Black, South Asian, Southeast Asian, or Indigenous), facial features (e.g., Middle Eastern, Pacific Islander, Afro-Caribbean, or Latin American), or cultural markers (e.g., hijabs, turbans, or tattoos) may each lead to biased outcomes.

\subsubsection{Experimental setup}

\textbf{Data:} Inspired by Waterbird \cite{Waterbirds} and CelebA \cite{CelebA}, we use a skewed training set with spurious correlations between class and style, and balanced validation/test sets. More specifically, for each SMA dataset, we select a number $K$ of content classes and $K$ styles classes from the 20 available, resulting in $K^2$ groups\footnote{Note that this choice is arbitrary and the number of classes and styles does not need to be the same.}. This setup allows each content class to be predominantly associated with exactly one style class, resulting in $K$ "dominant groups" where all examples are retained. In the training set, the remaining $K(K-1)$ "minority groups" are down-sampled, retaining only a fraction $\mu \in [0,1]$ of the training examples. This design emulates real-world scenarios where training data is often imbalanced across different groups, potentially leading to model biases that favor style over content for decision-making. By intentionally creating these imbalances, we aim to study how effectively fairness algorithms can address and mitigate such biases, ensuring that each content class is biased toward one specific style class.

Furthermore, we introduce content-level imbalance by randomly sampling 30\% to 100\% of examples per content class to reflect practical scenarios where there might be an imbalance in $Y$. A simple practical example can be found in Appendix \ref{appendix:example3x3}. The dataset is split into 50\%-20\%-30\% train-validation-test, while ensuring that stylized versions of the same image belong to the same split. This prevents data leakage by making sure that variations of the same image do not appear across different subsets, thereby maintaining the integrity of the training, validation, and testing processes.

\textbf{Benchmark Methods:} 
In our experiments, we address two types of methods based on the availability of group labels at training time. Firstly, group-aware methods that can be used when the group information  ($Y,Z|X$) is available during training, which allows methods to explicitly use group information to ensure fairness. Secondly, group-unaware training methods which only used class labels ($Y|X$), as in classical classification problems. Following the approach in \citet {idrissi2022simple}, group information is always used during validation to compute the worst-group accuracy metric for model selection.

We benchmark the seven methods from \citet{idrissi2022simple}, and adapt their code\footnote{\url{https://github.com/facebookresearch/BalancingGroups}} for SMA datasets. First, the classical Empirical Risk Minimization (\textbf{ERM}) minimizes average loss without access to group information. \textbf{JTT} \cite{liu2021just} retrains on hard examples to counteract spurious correlations without group labels. \textbf{DRO} \cite{Waterbirds} uses group labels to directly minimize the worst-case loss across groups. Simple subsampling methods like \textbf{SUBY} and \textbf{SUBG} adjust the dataset by reducing the number of examples based on class ($Y$) or group sizes. Similarly, \textbf{RWY} and \textbf{RWG} reweight the sampling probabilities of training examples to achieve balance across classes or groups. Note that SUBG, RWG, and DRO require access to group information which may not be accessible in real-world datasets.

\textbf{Validation and Evaluation: } Hyperparameters are optimized for each method and dataset using Bayesian optimization with DeepHyper \cite{deephyper_software} (more details in Appendix \ref{appendix:benchmark-fairness:Experimental-setup}). The hyperparameters set that achieves the highest optimization objective on the validation set is considered optimal. As a fairness metric, we use worst-group accuracy to maintain consistency with prior benchmarks and ensure comparability with existing work. This metric is widely used to evaluate fairness by reporting the accuracy on the samples of the most disadvantaged group. In previous works \cite{idrissi2022simple}, worst-group accuracy was also used for model selection during the hyperparameter optimization phase, to both guide the search for hyperparameters, and also to select the final model for testing. However, worst-group accuracy has notable limitations, particularly in high-diversity datasets like SMA. The metric is highly sensitive to outliers, where the performance of a single group may be disproportionately affected by statistical anomalies or small sample sizes, reducing its reliability in high-diversity contexts. Additionally, focusing on a single worst-performing group can fail to capture broader patterns of inequality, especially when multiple groups exhibit similar performance levels.
To overcome these challenges, in section \ref{subsubsec:Reevaluating-Fairness-Metrics-When-Increasing-Diversity}, we propose \textit{Top-M worst-group accuracy} as a new model selection metric. This metric, which averages performance across the M worst-performing groups, is used during hyperparameter optimization and final model selection to provide a more stable and reliable criterion in high-diversity contexts. However, the final fairness performance of models is still evaluated by worst-group accuracy to maintain comparability with prior work, and to fairly compare models, such as SUBG, tuned by Top-M worst-group accuracy against those tuned using worst-group accuracy.

Regarding the computational cost, conducting experiments on a single dataset involves testing 7 methods, each requiring hyperparameter optimization. For hyperparameters optimization, we performed 15 runs to select the best setting, followed by 5 additional runs with the optimized hyperparameters to compute error bars. With $\mu \in \{0.05, 0.1, 0.2\}$ and $K \in \{2, 4, 8, 12\}$, this resulted in 840 training sessions per dataset. By keeping these lower values of $K$ and $\mu$, the stylized datasets are heavily subsampled, which allowed us to maintain a relatively low computational cost. Notably, the training time depends quadratically on $K$. This amounts to 1,500 GPU hours for larger datasets and 50 GPU hours for smaller datasets, totaling approximately 5,000 GPU hours on an NVIDIA V100 to do the experiments of Section \ref{subsubsection:benchmark:fairness_result}.

\subsubsection{Fairness results on SMA Datasets}
\label{subsubsection:benchmark:fairness_result}
In the following section, we present our results and analysis using the {\it SMA\_PLT\_NET} dataset. This medium-sized plant dataset is representative of other SMA datasets in terms of size, statistical properties, and difficulty levels (see Figure \ref{violin_avgacc}), making it an appropriate choice for our illustration. However, our complete benchmark includes a broader collection of SMA datasets, encompassing a diverse range of image types such as textures, aerial shots, or preprocessed images. This variety allows for the evaluation of methods across different data distributions. For results on all other datasets, please refer to Appendix \ref{appendix:benchmark-fairness:Results-on-all-datasets}.

\paragraph{Simple balancing methods are still competitive}

\begin{figure}[ht]
    \centering
    \includegraphics[width=0.95\textwidth]{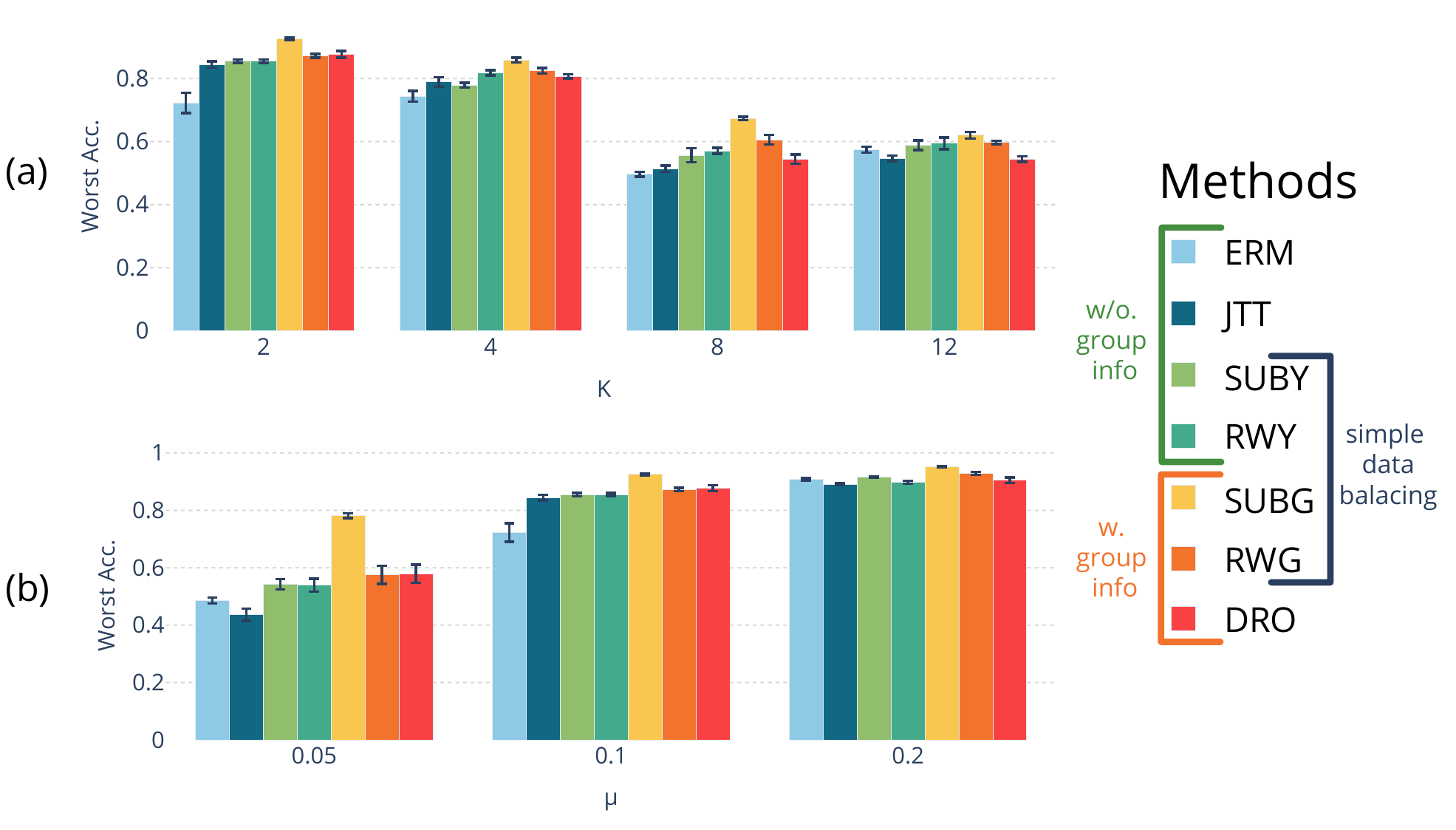}
    \caption{{\bf Group fairness:} Comparative analysis of methods for handling out-of-distribution (OOD) tasks, using the worst-group accuracy metric, exemplified by the \textit{SMA\_PLT\_NET} dataset. The evaluation is shown as a function of (a) $K$ (number of styles and content classes) with a fixed value of $\mu =10\%$ and (b) $\mu$ (lower values mean higher bias) with a fixed value of $K = 2$. Simple data balancing methods (in particular SUBG) are competitive with more complex methods across varying $K$ and $\mu$. When $K$ becomes larger, the ERM baseline achieves comparable performance to group-info-based methods, which shows that increasing diversity tends to improve fairness. Error bars represent the standard error over five runs with optimized hyperparameters.}
    \label{fig:plt-net}
\end{figure}

Our dataset subsampling strategy, as previously described, enables us to study the efficiency of methods with respect to $K$, the number of spurious features (styles), and content classes retained.  As shown in Figure \ref{fig:plt-net} (a) and first column of Figure \ref{fig:benchmark_all_worstgroupaccs}, when $\mu$ is fixed to $10\%$, the number of groups $K$ significantly impacts the relative performance of the different methods. Moreover, our findings reaffirm the quite surprising conclusion of the initial benchmark \cite{idrissi2022simple}: simple data balancing techniques are indeed competitive with more complex approaches. Specifically, among methods that do not utilize group information, simple balancing ones like SUBY and RWY, perform on par with JTT. Similarly, simple group-aware methods such as RWG and SUBG are competitive with DRO. These trends hold true across nearly all datasets.

Notably, SUBG, despite its simplicity, delivers robust performance across datasets. Unlike DRO, which dynamically adjusts group weights during training, SUBG employs a fixed and straightforward balancing strategy that decorrelates spurious features without the risk of overfitting to small or noisy groups. This advantage is particularly evident in SMA datasets, characterized by high group diversity and the presence of multiple minority groups. Here, SUBG outperforms DRO by maintaining stable performance and avoiding the pitfalls of dynamic weighting, such as overemphasis on noisy or underrepresented groups.

By easily subsampling minority groups, SMA datasets also allow to easily investigate the influence of inherent bias in the dataset, quantified by the parameter $\mu$, which represents the proportion of examples in each minority group relative to the majority group (a larger $\mu$ indicates a smaller bias). As observed in  Figure \ref{fig:plt-net} (b) and second columns of Figure \ref{fig:benchmark_all_worstgroupaccs}, when $\mu$ is smaller, methods using group information generally perform better, especially SUBG, which sometimes surpasses all other methods by a large margin. Conversely, when $\mu$ is higher, the number of examples in each group becomes more balanced, reducing the bias and leading to equal performance across all methods. Surprisingly, changes in $\mu$ not only affect the overall effectiveness of the methods but also shift their relative rankings. Each method exhibits distinct sensitivity to the proportion of minority class samples retained. This variability underscores the importance of understanding how different methods adapt to changes in dataset composition, particularly concerning bias.

\paragraph{Increasing diversity improves fairness}
Another notable finding from these experiments is that as $K$ increases, the ERM approach attains worst-group accuracy levels that become comparable to other methods, despite not explicitly addressing bias. This is the case for most of the datasets, including $\textit{SMA\_PLT\_NET}$ as illustrated in Figure \ref{fig:plt-net}.  Notably, when considering methods that do not use group information, such as JTT, SUBY, or RWG, we observe that none of these methods outperform significantly ERM at $K=12$ across any of all the 8 tested SMA datasets. This indicates that increased diversity may naturally mitigate bias. We hypothesize that as the number of styles and classes increases, the broader exposure to varied examples within each class helps the model to generalize better and rely less on existing spurious correlations in the training dataset.

\subsubsection{Reevaluating Fairness Metrics When Increasing Diversity}
\label{subsubsec:Reevaluating-Fairness-Metrics-When-Increasing-Diversity}

Experiments on SMA datasets have provided key insights into the limitations of using worst-group accuracy as the sole metric for assessing fairness, particularly as the diversity within the dataset increases (i.e. with higher values of $K$). As $K$ increases, the likelihood of encountering a subgroup with atypically poor data quality or representation also increases, especially in smaller datasets, which disproportionately affects the worst-group accuracy. This can lead to situations where, in the presence of a high number of groups, the worst-group accuracy metric may not accurately reflect the overall performance of the algorithm, leading to poor hyper-parameter optimization during the model selection phase This can lead to issues as K increases, when fair treatment should be ensured across multiple minority groups, resulting in the following flaws:

\begin{enumerate}
    \item {\bf Misleading Performance Indicators: } The worst-group accuracy, when extremely low, can paint a pessimistic picture of an otherwise decently performing model. Indeed, in several datasets within our analysis, we observed that the worst-group accuracy could approach nearly zero as $K$ increases, even though standard accuracy across all groups remained relatively high. More specifically, Figure \ref{fig:benchmark_all_worstgroupaccs} from the appendix demonstrates that increasing the value of $K$ on challenging datasets like \textit{SMA\_TEX\_DTD} or \textit{SMA\_AIRPLANES} makes the performance of most methods difficult to distinguish, particularly for methods that do not use group information. This is due to the very low accuracy scores among the worst-performing groups, which increase in number quadratically as $K$ increases.

    \item {\bf Impact on Hyperparameter Optimization:} In addition to the problem of comparing methods, relying on worst-group accuracy to tune hyperparameters can also be problematic, especially when this metric is notably low. A very low worst-group accuracy, which becomes statistically more likely as $K$ increases (more details in Appendix \ref{appendix:benchmark-fairness:statistical_property}), does not provide sufficient information about effective hyperparameters, leading to poor hyperparameter search and suboptimal model performance across diverse datasets. This is because our hyperparameter optimization approach relies on Bayesian optimization, which makes use of an acquisition function that depends on observed values to decide its next action. In challenging datasets, frequent occurrences of zero values or highly noisy values across the hyperparameter space can cause the objective function to appear nearly constant or noisy, making it difficult to pinpoint an accurate optimum. These effects hinder the ability of the hyperparameter optimization process to effectively identify the best settings.

\end{enumerate}

\begin{figure}[ht]
\centering
\includegraphics[width=.8\textwidth]{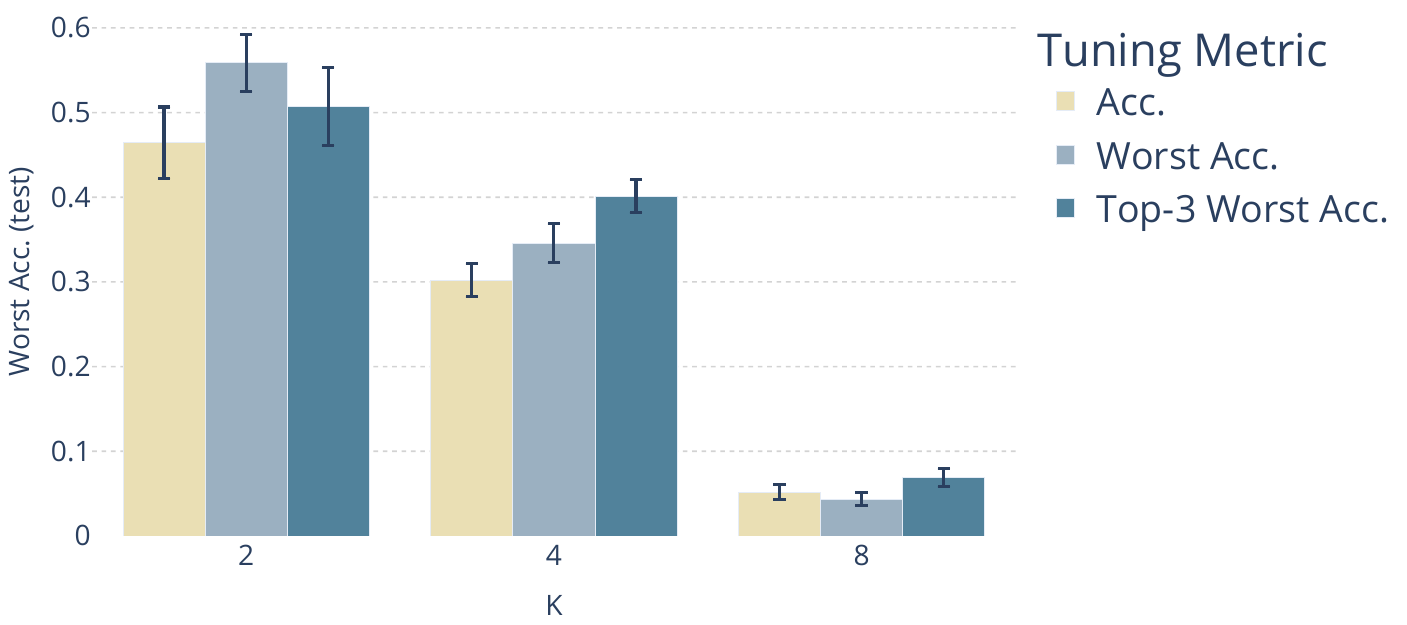}
\caption{Test worst-group accuracy on \textit{SMA\_TEX\_DTD} for the SUBG method based on different model selection metrics in the validation phase: simple accuracy (Acc.), worst-group accuracy (Worst Acc.), and the Top-3 worst-group accuracy (Top-3 Worst Acc.) for different values of K. Hyperparameter optimization based on our proposed Top-3 Worst Acc. significantly improves Worst Acc. performance in scenarios with higher number of groups ($K \geq 4$), demonstrating the ability of this tuning metric to ensure broader fairness and deliver better final results for large $K$. Each result corresponds to 10 distinct subsampling of the datasets (with different classes and styles selected), with 5 random seeds per subsampling, totaling 50 runs per configuration. The error bars represent the standard error.}
\label{fig:hyperparameter_tuning}
\end{figure}

\paragraph{Consider Top-M worst-group accuracy for model selection with larger K.}

\label{Consider-Top-M-worst-group-accuracy-for-larger-K}
To address the limitations of relying solely on worst-group accuracy for model selection on the evaluation set, we propose considering the average accuracy of the Top-M worst-performing groups when K becomes larger. This approach provides greater stability and reduces the high variability associated with focusing on a single worst group.
To demonstrate this, we propose a new set of experiments focusing on the SUBG method, which demonstrates the best performance on average across the SMA datasets in high-diversity settings. 

As illustrated in Figure~\ref{fig:hyperparameter_tuning}, tuning hyperparameters based on a broader assessment of the lower-performing groups like Top-3 worst-group accuracy, achieves better fairness at test time as measured by the classical worst-group accuracy when $K\geq 4$. This new tuning metric is expected to lead to more reliable and effective model optimization in scenarios with large group diversity. The Top-3 worst-group accuracy metric emphasizes the performance of lower-performing groups beyond just the absolute worst, providing a more comprehensive view of fairness during tuning. This approach results in models that exhibit higher worst-group accuracy at test time, demonstrating the effectiveness of this new tuning strategy for scenarios with large group diversity.  More details about these experiments and results covering more SMA datasets can be found in the Appendix  \ref{appendix:topm}.

Furthermore, for $K = 8$, we also observe that optimizing for classical accuracy, which is not a particularly good measure of fairness, results in a slightly superior worst-group accuracy compared to directly optimizing for worst-group accuracy. This finding further confirms that using worst-group accuracy as a metric for hyperparameter optimization may not be very effective in this scenario.

%% file: sections/domain_adaptation.tex
\subsection{Benchmark 2: Unsupervised Domain Adaptation}
\label{sec:benchmark2}



Domain adaptation (DA) trains a model using data from a source distribution, then adapts it for use on a related target distribution. In unsupervised domain adaptation (UDA), the source domain has labeled training data, while the target domain has unlabeled training data. To contrast with the previous benchmark on group fairness: in both cases (group fairness and DA), we always train and test within the same SMA dataset \(\mathcal{D}\).

However, in group fairness, we vary the distribution \((X|Y,Z)\) between the train \(\mathcal{D}_{tr}\) and validation \(\mathcal{D}_{va}\) / test \(\mathcal{D}_{te}\) sets, while keeping the set of values for \(Z\)-that is, the set of styles-consistent. On the other hand, in DA, \(\mathcal{D}\) is split into source \(\mathcal{D}_s\) and target \(\mathcal{D}_t\) domains by taking different values for \(Z\) (e.g., \(Z\) = cloud style in the source domain and \(Z\) = Chinese painting style in the target domain), while keeping the distribution \((X|Y)\) consistent across the train and test sets, for both source and target domains.

To use SMA for DA benchmarking purposes, domains can be identified to styles. SMA is a meta-dataset with 12 datasets, each containing 20 styles, providing a total of 20 domains for evaluating UDA algorithms. This diversity makes SMA an excellent testbed for DA (and particularly UDA), offering lower error bars on benchmark results, due to the large number of domains. 

In this study, we use 5 SMA datasets to benchmark state-of-the-art UDA methods, demonstrating the benefit of lower error bars compared to existing benchmarks from the literature. Initially, we started with 3 datasets but expanded to 5 to create a more robust benchmark (lower error bars) while maintaining reasonable computational costs. We selected these 5 datasets to cover a diverse range of tasks, including object classification, texture classification, and human activity recognition, and to account for the impact of varying dataset sizes and statistics. 
We carry out the evaluation in two scenarios: (1) closed-set DA, where source (\(Y_s\)) and target (\(Y_t\)) label sets are identical; and (2) universal DA (UniDA), where \(Y_s\) and \(Y_t\) can differ. Appendix \ref{appendix:usecase-UDA} offers a detailed introduction to DA, UDA, closed-set DA and universal DA.

\subsubsection{Benchmark Experiments}

\textbf{Data:} We selected the stylized versions of BIRDS (SMA\_BRD), DOGS (SMA\_DOG), SPORTS (SMA\_SPTS), PLT\_DOC (SMA\_PLT\_DOC), and APL (SMA\_APL) from the SMA benchmark datasets to encompass diverse classification tasks, including object classification, texture classification, and human activity recognition. These datasets also vary in size, from PLT\_DOC with 2,135 images to APL with 10,205 images, allowing us to examine how the amount of training data impacts model performance.

We randomly designated 25 scenarios (1 scenario is 1 transfer task from source to target) for domain adaptation training using these datasets. For UniDA, we structured each scenario with ten shared classes ($Y=Y_s \cap Y_t$), five private source classes ($\overline{Y}_s = Y_s \backslash Y$), and five private target classes ($\overline{Y}_t = Y_t \backslash Y$), arranged alphabetically. We used a 50/50 train/test split across all datasets, ensuring robustness and consistency in our evaluations, made possible by the ample data in the SMA benchmark.

\textbf{Evaluation:} We report results only on the target test set. In the case of closed-set experiments, we only report accuracy, because it is one of the metrics commonly used in domain adaptation \cite{deepJDOT, CDAN, DANN}. Though accuracy is a less effective metric for imbalanced datasets, it was chosen to enable comparison with Office31 and OfficeHome datasets for which only accuracy is reported. For UniDA, we consider the standard setting where each model can classify a sample as one of the K source classes or as an ``unknown'' class, thus extending the number of classes to (K+1).
The metric reported for UniDA is the H-score $= \frac{2A_cA_u}{A_c+A_u}$, which is the harmonic mean between the accuracy of known classes $A_c$ and the accuracy of target unknown classes $A_u$. This metric, introduced in \cite{CMU}, evaluates a model's ability to balance classification performance across known and unknown classes. However, we acknowledge its limitation: the H-score does not account for the imbalance between known and unknown samples, as it assigns equal weight to both regardless of their proportions in the dataset. Despite this drawback, it remains a standard metric in the field and is used here for consistency and comparison with prior works.\\

\textbf{Benchmark Models and Implementation: } For closed-set domain adaptation, we trained three state-of-the-art methods: \textbf{DANN} \cite{DANN}, \textbf{CDAN} \cite{CDAN}, \textbf{DeepJDOT} \cite{deepJDOT}. For reference, we also trained a source-only model (\textbf{NO\_ADAPT}) which is trained on source domain and tested on target domain without adaptation. For Universal Domain Adaptation (UniDA), we trained four methods: \textbf{UDA} \cite{UDA_you}, \textbf{OSBP} \cite{OSBP}, \textbf{OVANet} \cite{OVANet}, and \textbf{UniOT} \cite{uniOT}. Detailed reviews of these models are provided in Appendix \ref{appendix:usecase-UDA}. For implementation, we adapted code from Adatime \cite{adatime}, replacing the 1D CNN feature extractor with a pretrained ResNet50 for images and extending it for UniDA scenarios. We set the learning rate to \(1 \times 10^{-4}\), epochs to 10, and batch size to 32 (except UniOT, which used 16 due to memory constraints). Increasing the number of epochs to 30 did not improve performance. Additional hyperparameters were configured by Adatime or the original authors for each method. \\

\subsubsection{Benchmark results and analysis}

\begin{figure}[h]
    \centering
    \begin{subfigure}{0.5\textwidth}
        \centering
        \includegraphics[width=\linewidth]{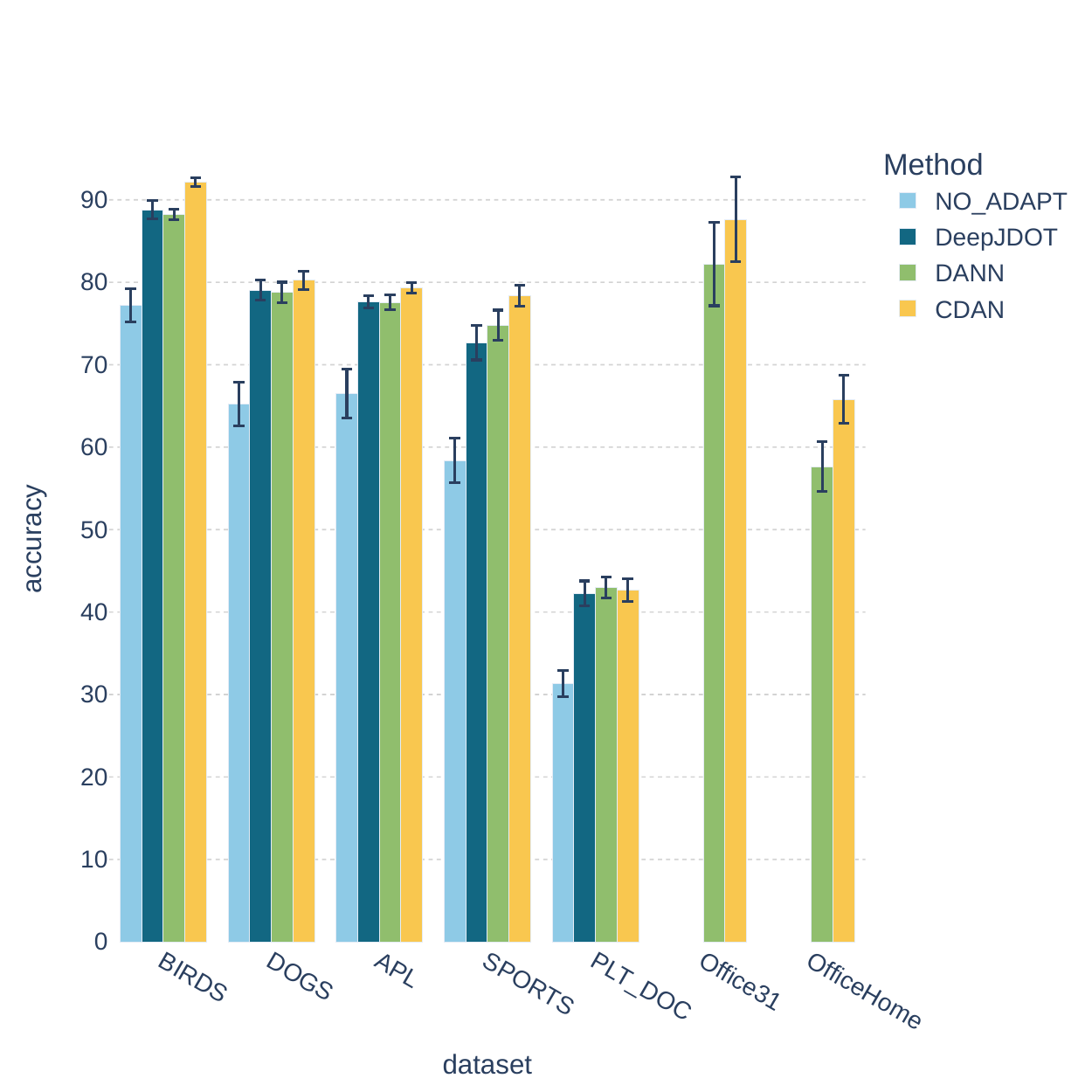}
        \caption{Mean accuracy for Closed Set Scenarios}
        \label{fig:usecase_uda}
    \end{subfigure}\begin{subfigure}{0.5\textwidth}
        \centering
        \includegraphics[width=\linewidth]{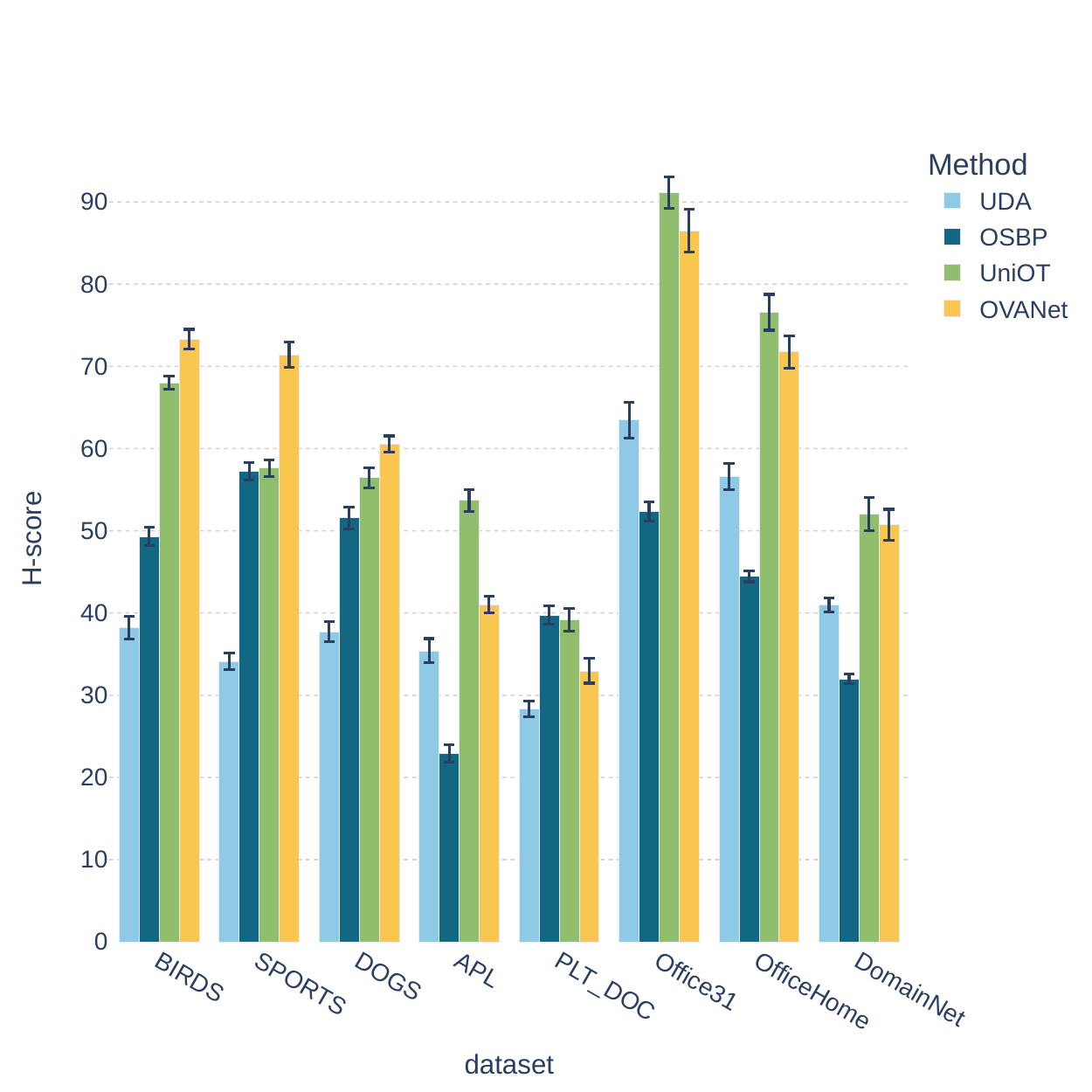}
        \caption{Mean H-score for Universal Domain Adaptation}
        \label{fig:usecase_unida}
    \end{subfigure}
    
    \caption{Mean accuracy and H-score over 25 domain adaptation tasks for each of the SMA datasets, 9 for OfficeHome, 6 for Office31 and 6 for DomainNet.}
    \label{fig:combined}
\end{figure}

\textbf{Performance Analysis: }
Figure \ref{fig:usecase_uda} shows the Closed-Set Results. The results for the non-SMA datasets (Office-31 and OfficeHome) are sourced from \cite{DSAN}, which only test 2 methods DANN and CDAN. The light blue bar indicates the outcome with no domain adaptation (NO\_ADAPT), where the model is trained on the source domain and evaluated on the target domain, illustrating the performance drop due to domain shift. Models such as DeepJDOT, DANN, and CDAN, which are designed to handle data shifts, consistently improve performance. Notably, CDAN outperforms other models, achieving the best results on all SMA datasets and also on the "office" datasets, confirming the results of previous studies.

Figure \ref{fig:usecase_unida} shows the UniDA Results. The results for the non-SMA datasets (Office-31, OfficeHome, DomainNet, and VisDA) are sourced from \cite{uniOT}. Notably, OVANet demonstrates remarkable efficacy, outperforming all other methods on three out of the five SMA datasets (BRD, SPORTS, DOGS, which are balanced). For the other two imbalanced datasets (PLT-DOC and APL), UniOT achieves the best performance. This might be due to its balancing components, specifically adaptive filling, which allows balancing the proportion of known and unknown samples in every mini-batch. 

In both cases, the error bars represent the \textbf{Standard Error of the Mean (SEM)} which is the precision of the sample mean estimate of a population mean. They are calculated as: $
\text{SEM} = \frac{\text{SD}}{\sqrt{n}}$, where \(\text{SD}\) is the sample standard deviation and \(n\) is the sample size (i.e. the number of scenarios). These error bars measure performance variability and the statistical significance of the obtained results. Smaller SEM indicate greater precision and are crucial in Domain Adaptation, where new state-of-the-art methods often outperform their competitors by small margins, as shown in Figure \ref{fig:combined}.\\

\textbf{Improved Benchmark Robustness: } By sampling 25 scenarios for each SMA dataset in both settings, we achieved a significant average reduction in the error bars.

This implies an improvement in the statistical significance of the reported results. Specifically, for each group comprising SMA and non-SMA datasets, we computed the average SEM across all datasets and models for both SMA and non-SMA datasets. In the case of the closed-set scenario, we excluded the No Adapt and DeepJDOT models, as results for these models were not available for the non-SMA datasets. The ratio was calculated as: $\text{Ratio} = \frac{\text{SEM of SMA}}{\text{SEM of non-SMA}}$. This calculation yielded a ratio of 0.27 in the closed-set scenario and 0.72 for UniDA. To interpret these ratios, we calculated the percentage reduction in SEM as $(1-\text{Ratio}) \times 100$. This corresponds to a 73\% reduction in SEM in the closed-set scenario and a 28\% reduction for UniDA. This extensive sampling approach enhances the robustness and reliability of the benchmarking results. The standard error over scenarios is notably smaller compared to traditional benchmarks like Office-31 (6 scenarios) and OfficeHome (12 scenarios), demonstrating the advantage of the SMA dataset in providing a broader evaluation spectrum.


%
\textbf{SMA Offers Diverse Difficulties: } SMA encompasses a wide range of domain adaptation datasets, covering various classification tasks such as human activity recognition, object classification, and texture classification under different conditions. This diversity allows us to benchmark state-of-the-art (SOTA) methods across tasks with varying levels of difficulty, as indicated by the average accuracy across benchmarked methods. demonstrate this range of difficulty, with average accuracies from the easiest 85\% (BIRDS) to the most challenging 40\% (PLT\_DOC) in Closed-Set (Figure \ref{fig:usecase_uda}), and from 55\% (BIRDS) to 35\% (PLT\_DOC) in UniDA (Figure \ref{fig:usecase_unida}). Expanding this benchmark with additional SMA datasets will further enhance our ability to investigate the performance of SOTA methods across a richer spectrum of difficulties.

%% file: sections/discussion.tex
\section{Conclusion and Broader Impact Discussion}
\label{sec:Conclusion}
In this paper, we introduced the Stylized Meta-Album (SMA), a comprehensive meta-dataset featuring a wide array of classes and styles, designed to facilitate extensive research in fields such as out-of-distribution (OOD) generalization and related topics. We demonstrated SMA's utility through two benchmarks in group fairness and unsupervised domain adaptation, providing significant insights beyond existing efforts and highlighting the substantial impact of increased variability in the nuisance factor (i.e., style) $Z$. We plan to release future versions of SMA with more data generated using faster algorithms, potentially establishing a benchmark for style transfer algorithms. This will produce stylized images of varying difficulty levels, offering a diverse and challenging dataset for researchers. Additionally, we aim to explore further use cases, such as shortcut learning and bias mitigation, enhancing SMA's versatility and stimulating advancements across various research domains. We also acknowledge recent methods such as last-layer retraining \cite{kirichenko2023last}, which adapts only the final classifier layer to mitigate spurious correlations, and Correct-N-Contrast \cite{zhang2024correctncontrastcontrastiveapproachimproving}, which uses a contrastive learning objective to learn group-robust representations without requiring group annotations. While these approaches were not included in our current benchmarks, incorporating them is part of our future work.

While SMA offers significant advantages, it also has limitations. The AdaIN technique used for stylization can introduce artifacts that affect image quality. Our preliminary experiments in group fairness were limited by computational costs, preventing a finer variation in the number of groups. Although we do not foresee negative societal consequences from our work, the high computational cost of using SMA for extensive experiments may limit accessibility for researchers with fewer resources. Another limitation lies in the reliance on worst-group accuracy as a primary fairness metric. While this choice aligns with prior benchmarks, it has inherent constraints, particularly in datasets with high group diversity. Metrics such as equality of opportunity or equality of odds could offer alternative perspectives on fairness but would require methods specifically optimized for these criteria. Enforcing equality constraints across many groups also introduces statistical noise and variability, especially for smaller or underrepresented groups, further complicating fairness evaluations.
These limitations highlight areas for future research. Addressing these challenges will enhance SMA’s utility and maximize its positive impact on the research community.

%% file: appendices/datasheet.tex
\section{Datasheet for Stylized Meta-Album Datasets}
\label{appendix:datasheet}

\subsection*{Motivation}

\textcolor{blue}{\textbf{For what purpose was the dataset created?} Was there a specific task in mind? Was there a specific gap that needed to be filled? Please provide a description.}

\noindent Stylized Meta-Album (SMA) datasets are created for benchmark purposes. The recommended use of Stylized Meta-Album is to conduct fundamental research on machine learning algorithms and conduct benchmarks, particularly in: transfer learning, image classification, domain adaptation, out-of-distribution generalization, etc.\\

\noindent \textcolor{blue}{\textbf{Who created the dataset (e.g., which team, research group) and on behalf of which entity (e.g., company, institution, organization)?}}

\noindent SMA meta-dataset is created from Meta-Album meta-dataset. Ihsan Ullah, Romain Mussard and Thanh Gia Hieu Khuong contributed in the creation of the meta-dataset. This work was supervised by Professor Isabelle Guyon. The work was performed at LISN laboratory, Université Paris-Saclay, France, in the TAU team, as part of the HUMANIA project, funded by the French research agency ANR. ChaLearn also supported the creation of the meta-dataset.\\

\noindent \textcolor{blue}{\textbf{Who funded the creation of the dataset? } If there is an associated grant, please provide the name of the grantor and the grant name and number.}

\noindent ANR (Agence Nationale de la Recherche, National Agency for Research\footnote{\url{https://anr.fr/}}, grant number 20HR0134 and ChaLearn\footnote{\url{http://www.chalearn.org/}} a 501(c)(3) non-for-profit California organization.
\\

\subsection*{Composition}

\textcolor{blue}{\textbf{What do the instances that comprise the dataset represent (e.g., documents, photos, people, countries)?} Are there multiple types of instances (e.g., movies, users, and ratings; people and interactions between them; nodes and edges)? Please provide a description.}

\noindent The instances are 256×256 RGB images.
\\

\noindent \textcolor{blue}{\textbf{How many instances are there in total (of each type, if appropriate)?}}

\noindent The meta-dataset consists of 24 datasets(12 content datasets from Meta-Album and 12 stylized datasets), each with two versions \textit{(Extended and Mini)}. Each dataset has at least 20 classes and 40 images per class. 
\\

\noindent \textcolor{blue}{\textbf{Does the dataset contain all possible instances or is it a sample (not necessarily random) of instances from a larger set?} If the dataset is a sample, then what is the larger set? Is the sample representative of the larger set (e.g., geographic coverage)? If so, please describe how this representativeness was validated/verified. If it is not representative of the larger set, please describe why not (e.g., to cover a more diverse range of instances, because instances were withheld or unavailable).}

\noindent Each dataset in the SMA meta-dataset has two versions. The Extended version consists of all possible instances while the Mini version has a subset of the instances.
\\

\noindent \textcolor{blue}{\textbf{What data does each instance consist of?} “Raw” data (e.g., unprocessed text or images) or features? In either case, please provide a description.}

\noindent Each instance is 256×256 RGB image. The instances are preprocessed i.e. resized into 256×256 with an anti-aliasing filter.
\\

\noindent \textcolor{blue}{\textbf{Is there a label or target associated with each instance?} If so, please provide a description.}

\noindent Yes, each instance has a category/label which is provided with the images in meta-data. 
\\

\noindent \textcolor{blue}{\textbf{Is any information missing from individual instances?} If so, please provide a description, explaining why this information is missing (e.g., because it was unavailable). This does not include intentionally removed information, but might include, e.g., redacted text.}

\noindent No, all information is provided for each instance.
\\

\noindent \textcolor{blue}{\textbf{Are relationships between individual instances made explicit (e.g., users' movie ratings, social network links)?} If so, please describe how these relationships are made explicit.}

\noindent All relationships are contained in categories/classes.
\\

\noindent \textcolor{blue}{\textbf{Are there recommended data splits (e.g., training, development/validation, testing)?}  If so, please provide a description of these splits, explaining the rationale behind them.}

\noindent The data has no splits.
\\

\noindent \textcolor{blue}{\textbf{Are there any errors, sources of noise, or redundancies in the dataset?}  If so, please provide a description.}

\noindent No, there are no suspected errors, sources of noise, or redundancies.
\\

\noindent \textcolor{blue}{\textbf{Is the dataset self-contained, or does it link to or otherwise rely on external resources (e.g., websites, tweets, other datasets)?}  If it links to or relies on external resources, a) are there guarantees that they will exist, and remain constant, over time; b) are there official archival versions of the complete dataset (i.e., including the external resources as they existed at the time the dataset was created); c) are there any restrictions (e.g., licenses, fees) associated with any of the external resources that might apply to a future user? Please provide descriptions of all external resources and any restrictions associated with them, as well as links or other access points, as appropriate.}

\noindent Each dataset in the SMA meta-dataset is self-contained. It will exist, and remain constant, over time.\\

\noindent \textcolor{blue}{\textbf{Does the dataset contain data that might be considered confidential (e.g., data that is protected by legal privilege or by doctor–patient confidentiality, data that includes the content of individuals’ non-public communications)?} If so, please provide a description.}

\noindent No, all the datasets are released and are free to use with the provided license.\\

\noindent \textcolor{blue}{\textbf{Does the dataset contain data that, if viewed directly, might be offensive, insulting, threatening, or might otherwise cause anxiety?} If so, please describe why.}

\noindent No.
\\

\noindent \textcolor{blue}{\textbf{Does the dataset relate to people?}  If not, you may skip the remaining questions in this section.}

\noindent No.
\\

\noindent \textcolor{blue}{\textbf{Does the dataset identify any subpopulations (e.g., by age, gender)?}  If so, please describe how these subpopulations are identified and provide a description of their respective distributions within the dataset.}

\noindent No.
\\

\noindent \textcolor{blue}{\textbf{Is it possible to identify individuals (i.e., one or more natural persons), either directly or indirectly (i.e., in combination with other data) from the dataset?} If so, please describe how.}

\noindent No.
\\

\noindent \textcolor{blue}{\textbf{Does the dataset contain data that might be considered sensitive in any way (e.g., data that reveals racial or ethnic origins, sexual orientations, religious beliefs, political opinions or union memberships, or locations; financial or health data; biometric or genetic data; forms of government identification, such as social security numbers; criminal history)?} If so, please provide a description.}

\noindent No.
\\

\subsection*{Collection Process}

\textcolor{blue}{\textbf{How was the data associated with each instance acquired?} Was the data directly observable (e.g., raw text, movie ratings), reported by subjects (e.g., survey responses), or indirectly inferred/derived from other data (e.g., part-of-speech tags, model-based guesses for age or language)? If data were reported by subjects or indirectly inferred/derived from other data, was the data validated/verified? If so, please describe how.}

\noindent Each instance is an image and is directly observable.
\\

\noindent \textcolor{blue}{\textbf{What mechanisms or procedures were used to collect the data (e.g., hardware apparatus or sensor, manual human curation, software program, software API)? } How were these mechanisms or procedures validated?}

\noindent 12 content datasets are taken from Meta-Album meta-dataset from the internet while 12 stylized datasets are generated using these content datasets.
\\

\noindent \textcolor{blue}{\textbf{If the dataset is a sample from a larger set, what was the sampling strategy (e.g., deterministic, probabilistic with specific sampling probabilities)?}}

\noindent N/A.\\

\noindent \textcolor{blue}{\textbf{Who was involved in the data collection process (e.g., students, crowdworkers, contractors) and how were they compensated (e.g., how much were crowdworkers paid)?}}

\noindent Ihsan Ullah, Romain Mussard and Thanh Gia Hieu Khuong were involved in the data collection process.
\\

\noindent \textcolor{blue}{\textbf{Over what timeframe was the data collected?} Does this timeframe match the creation timeframe of the data associated with the instances (e.g., recent crawl of old news articles)? If not, please describe the timeframe in which the data associated with the instances was created.}

\noindent The data were collected between June 2023 and April 2024.
\\

\noindent \textcolor{blue}{\textbf{Were any ethical review processes conducted (e.g., by an institutional review board)?} If so, please provide a description of these review processes, including the outcomes, as well as a link or other access point to any supporting documentation.}

\noindent N/A
\\

\noindent \textcolor{blue}{\textbf{Does the dataset relate to people?} If not, you may skip the remaining questions in this section.}

\noindent No.
\\

\noindent \textcolor{blue}{\textbf{Did you collect the data from the individuals in question directly, or obtain it via third parties or other sources (e.g., websites)?}}

\noindent N/A
\\

\noindent \textcolor{blue}{\textbf{Were the individuals in question notified about the data collection?} If so, please describe (or show with screenshots or other information) how notice was provided, and provide a link or other access point to, or otherwise reproduce, the exact language of the notification itself.}

\noindent N/A
\\

\noindent \textcolor{blue}{\textbf{Did the individuals in question consent to the collection and use of their data?} If so, please describe (or show with screenshots or other information) how consent was requested and provided, and provide a link or other access point to, or otherwise reproduce, the exact language to which the individuals consented.}

\noindent N/A
\\

\noindent \textcolor{blue}{\textbf{If consent was obtained, were the consenting individuals provided with a mechanism to revoke their consent in the future or for certain uses?} If so, please provide a description, as well as a link or other access point to the mechanism (if appropriate).}

\noindent N/A
\\

\noindent \textcolor{blue}{\textbf{Has an analysis of the potential impact of the dataset and its use on data subjects (e.g., a data protection impact analysis) been conducted?} If so, please provide a description of this analysis, including the outcomes, as well as a link or other access point to any supporting documentation. }

\noindent N/A
\\

\subsection*{Preprocessing/cleaning/labeling}

\textcolor{blue}{\textbf{Was any preprocessing/cleaning/labeling of the data done (e.g., discretization or bucketing, tokenization, part-of-speech tagging, SIFT feature extraction, removal of instances, processing of missing values)?} If so, please provide a description. If not, you may skip the remainder of the questions in this section.}

\noindent Yes, the data is preprocessed: the images are resized into 256x256 with an anti-aliasing filter.
\\

\noindent \textcolor{blue}{\textbf{Was the “raw” data saved in addition to the preprocessed/cleaned/labeled data (e.g., to support unanticipated future uses)?} If so, please provide a link or other access point to the “raw” data.}

\noindent The raw data is not released with the preprocessed datasets however it can be accessed from its original sources.\\

\noindent \textcolor{blue}{\textbf{Is the software used to preprocess/clean/label the instances available?} If so, please provide a link or other access point.}

\noindent Yes, the preprocessing software is available in the Stylized Meta-Album Github repository. Details are provided on the Stylized Meta-Album website: \url{https://stylized-meta-album.github.io/}.\\

\subsection*{Uses} 
\noindent \textcolor{blue}{\textbf{Has the dataset been used for any tasks already?} If so, please provide a description.}

\noindent These datasets are already used in the described tasks in this paper.\\

\noindent \textcolor{blue}{\textbf{Is there a repository that links to any or all papers or systems that use the dataset?} If so, please provide a link or other access point.}

\noindent Yes, a dedicated GitHub repository will be active once the meta-dataset is publicly released. Details are provided on the Stylized Meta-Album website (\url{https://stylized-meta-album.github.io/}). This website will also be used to announce any necessary information related to the meta-dataset.

\noindent \textcolor{blue}{\textbf{What (other) tasks could the dataset be used for?}}

\noindent Besides transfer learning, these datasets could be used for classification tasks, out-of-distribution generalization, domain adaptation, etc.\\

\noindent \textcolor{blue}{\textbf{Is there anything about the composition of the dataset or the way it was collected and preprocessed/cleaned/labeled that might impact future uses? }For example, is there anything that a future user might need to know to avoid uses that could result in unfair treatment of individuals or groups (e.g., stereotyping, quality of service issues) or other undesirable harms (e.g., financial harms, legal risks) If so, please provide a description. Is there anything a future user could do to mitigate these undesirable harms?}

\noindent All datasets in the SMA meta-dataset have been prepared for benchmarks in machine learning and no other purposes. We do not make any warranties that are appropriate for conducting scientific research other than research on machine learning algorithms nor that they are fit for developing products, whether commercial or not. In particular, these datasets may include biases that could render them unfit for such other purposes.\\

\noindent \textcolor{blue}{\textbf{Are there tasks for which the dataset should not be used? }If so, please provide a description.}

\noindent Until possible biases are further investigated, the datasets should not be used for any other purpose than their primary intended purpose (benchmarks).\\

\subsection*{Distribution} 
\noindent \textcolor{blue}{\textbf{Will the dataset be distributed to third parties outside of the entity (e.g., company, institution, organization) on behalf of which the dataset was created? }If so, please provide a description.}

\noindent SMA datasets will be made available to everyone. More details about distribution can be found on the Stylized Meta-Album website (\url{https://stylized-meta-album.github.io/}).\\

\noindent \textcolor{blue}{\textbf{How will the dataset be distributed (e.g., tarball on website, API, GitHub)? }Does the dataset have a digital object identifier (DOI)?}

\noindent The access information and any necessary updates will be announced via the Stylized Meta-Album website (\url{https://stylized-meta-album.github.io/}).\\

\noindent \textcolor{blue}{\textbf{When will the dataset be distributed?}}

\noindent SMA datasets will be distributed in August 2024.\\

\noindent \textcolor{blue}{\textbf{Will the dataset be distributed under a copyright or other intellectual property (IP) license, and/or under applicable terms of use (ToU)? }If so, please describe this license and/or ToU, and provide a link or other access point to, or otherwise reproduce, any relevant licensing terms or ToU, as well as any fees associated with these restrictions.}

\noindent SMA datasets are public for research and are released with Stylized Meta-Album license CC-BY-NC 4.0. Further information about licenses can be found in the 'info.json' meta-data file. The license information is also mentioned on the website (\url{https://stylized-meta-album.github.io/}).
\\

\noindent \textcolor{blue}{\textbf{Have any third parties imposed IP-based or other restrictions on the data associated with the instances? }If so, please describe these restrictions, and provide a link or other access point to, or otherwise reproduce, any relevant licensing terms, as well as any fees associated with these restrictions.}

\noindent No.\\

\noindent \textcolor{blue}{\textbf{Do any export controls or other regulatory restrictions apply to the dataset or to individual instances? }If so, please describe these restrictions, and provide a link or other access point to, or otherwise reproduce, any supporting documentation.}

\noindent No.\\

\subsection*{Maintenance} 
\noindent \textcolor{blue}{\textbf{Who will be supporting/hosting/maintaining the dataset?}}

\noindent The authors of \href{https://stylized-meta-album.github.io/paper.html}{Stylized Meta-Album paper} will be responsible for supporting the meta-dataset.\\

\noindent \textcolor{blue}{\textbf{How can the owner/curator/manager of the dataset be contacted (e.g., email address)?}}

\noindent The preferred way to contact the maintainers is to raise issues on the Github repository, details are provided on the Stylized Meta-Album website(\url{https://stylized-meta-album.github.io/}). In case of emergency, the authors of the Stylized Meta-Album paper can be contacted via email: stylized-meta-album@chalearn.org.\\

\noindent \textcolor{blue}{\textbf{Is there an erratum? If so, please provide a link or other access point.}}

\noindent Any necessary information or updates will be accessible via the corresponding website (\url{https://stylized-meta-album.github.io/}).\\

\noindent \textcolor{blue}{\textbf{Will the dataset be updated (e.g., to correct labeling errors, add new instances, delete instances)? }If so, please describe how often, by whom, and how updates will be communicated to users (e.g., mailing list, GitHub)?}

\noindent We have no intention to update the datasets unless required. In any case, updates will be available on the website (\url{https://stylized-meta-album.github.io/}).\\

\noindent \textcolor{blue}{\textbf{If the dataset relates to people, are there applicable limits on the retention of the data associated with the instances (e.g., were individuals in question told that their data would be retained for a fixed period of time and then deleted)? }If so, please describe these limits and explain how they will be enforced.}

\noindent N/A\\

\noindent \textcolor{blue}{\textbf{Will older versions of the dataset continue to be supported/hosted/maintained? }If so, please describe how. If not, please describe how its obsolescence will be communicated to users.}

\noindent Any necessary information or updates will be accessible via the website (\url{https://stylized-meta-album.github.io/}).\\

\noindent \textcolor{blue}{\textbf{If others want to extend/augment/build on/contribute to the dataset, is there a mechanism for them to do so? }If so, please provide a description. Will these contributions be validated/verified? If so, please describe how. If not, why not? Is there a process for communicating/distributing these contributions to other users? If so, please provide a description.}

\noindent We have provided a complete protocol of how such datasets can be produced in the Stylized Meta-Album paper and the procedures and code for verification/validation of newly constructed datasets using the defined protocol are given in the GitHub repository (details on our website: \url{https://stylized-meta-album.github.io/}). All updates will be available on the website and the authors can be contacted via  email: stylized-meta-album@chalearn.org.\\

%% file: appendices/glossary.tex
\section{Glossary}
\label{appendix:glossary}
\begin{itemize}

\item \textbf{Out-of-Distribution (OOD) Generalization:} Refers to a model’s ability to perform well on new or unseen data distributions that differ from the training set. OOD generalization addresses robustness to any distributional shift, focusing on how models handle changes in the underlying data patterns while maintaining predictive performance.

\item \textbf{Group Fairness:} Ensures equitable performance across predefined subgroups within a population, often defined by protected attributes such as gender, race, or age. Unlike OOD generalization, which targets robustness to any distributional shift, group fairness emphasizes reducing disparities in performance across these known groups.

\item \textbf{Robustness:} Refers to the reliability and consistency of a model’s performance across varying conditions, evaluated with statistical confidence. In the context of SMA, robustness is assessed in two ways: 

\begin{itemize} 

\item In group fairness studies, robustness reflects the model’s ability to resist spurious correlations during training and maintain equitable performance across challenging subgroups. This is typically measured through metrics like worst-group accuracy and top-$M$ worst-group accuracy. 

\item In unsupervised domain adaptation (UDA) studies, robustness refers to the stability and reliability of benchmark results achieved through extensive sampling of transfer tasks (scenarios). Increasing the number of sampled scenarios reduces variability (e.g., SEM) and improves the statistical significance of the results, which is referred to as an improvement in robustness. 
\end{itemize}

\item \textbf{Unsupervised Domain Adaptation (UDA):} Field that Involves adapting a model trained on a labeled source domain to an unlabeled target domain. In SMA, domain shifts are represented by varying styles ($Z$), simulating domain adaptation scenarios. 

\item \textbf{Meta-dataset:} A collection of datasets sharing a unified structure. The SMA meta-dataset consists of 12 content datasets paired with their stylized counterparts, forming a total of 24 datasets.

\item \textbf{Content Dataset:} A dataset comprising labeled images (Y) from domains such as animals, plants, vehicles, and human actions. These datasets, sourced from Meta-Album, serve as the base content for stylization.

\item \textbf{Style Dataset:} A dataset containing images representing 20 distinct visual styles (Z), sourced from publicly available images. These styles are used to transform content datasets into stylized datasets, introducing controlled variability for benchmarking and experimentation.

\item \textbf{Stylized Dataset:} A stylized dataset consists of stylized images generated by applying style transformations to the content dataset.


\item \textbf{Dataset:} An SMA dataset refers to a specific dataset within the SMA meta-dataset. It consists of: (1) a stylized dataset: stylized images organized into folders named according to the labels specified in the provided labels.csv file; (2) a content dataset: a labels.csv file containing image labels and associated metadata, the actual content images can be downloaded separately.

\item \textbf{Worst-Group Accuracy:} A metric used to evaluate the fairness of a model by measuring its performance on the group with the lowest accuracy. This metric is particularly useful in scenarios where equitable performance across all groups is critical, as it highlights the subgroup where the model struggles the most, ensuring that efforts to improve fairness focus on minimizing disparities.

\item \textbf{Top-M Worst-Group Accuracy:} A metric we propose as a refinement of worst-group accuracy. It computes the average accuracy of the $M$ lowest-performing groups. This metric provides a more stable and less noisy measure of fairness in datasets with high group diversity or small group sizes. By averaging over multiple low-performing groups, it reduces the sensitivity to outliers and better reflects overall fairness across challenging subgroups.

\item \textbf{Accuracy:} A metric used to evaluate the overall performance of a model by measuring the proportion of correctly classified samples across all categories.

\item \textbf{H-Score:} A metric specifically designed for Universal Domain Adaptation to balance performance across known and unknown classes. It is calculated as the harmonic mean of the accuracy on known classes (source-aligned categories) and the accuracy on unknown classes. The H-Score ensures that improvements in recognizing unknown categories do not come at the expense of degrading performance on shared classes, thereby promoting a more balanced and robust evaluation of UDA models.


\end{itemize}

%% file: appendices/style-dataset.tex
\section{Style Dataset}
\label{appendix:style-dataset}
In this section, we provide details about the creation of our style dataset, used for generating the SMA datasets. As mentioned in the main paper, all usage of these style images is ensured under the fair and transformative use rules. Our work is protected by license in Appendix \ref{appendix:license}.
\subsection*{Style dataset categories}
Style dataset is created by searching some keywords on the internet. We identify high quality images for each keyword and keep the link to these images. Later on these links are used as input of the scrapping program. Each keyword is considered a class. Style dataset consists of 20 classes/categories and 40 images per class. In total it consists of 800 images.

Following are the 20 categories for style dataset:
\begin{AutoMultiColItemize}
    \item prehistorian painting
    \item analytical art
    \item water lilies
    \item swamp
    \item american barbizon school
    \item arts and crafts movement
    \item aurora
    \item fauvism painting
    \item cloud
    \item chinese art
    \item wave
    \item american impressionism
    \item land art
    \item abstract illusionism
    \item hurricane
    \item meadow
    \item dawn
    \item rayonism
    \item autumn leaves
    \item thunder
\end{AutoMultiColItemize}


\subsection*{Scraping process}

\textit{Initial Collection}

The project started with a manually curated set of 50 style images, encompassing a range of artistic and painting styles as well as natural scenes. Each image was carefully chosen based on specific keywords that best represent the desired styles. These images served as the seed for further image collection through the scraping program.

\noindent\textit{Scraping Program}

The \href{https://github.com/ihsaan-ullah/stylized-meta-album/tree/master/style_dataset_preparation/style_image_scrapping}{Scraping Program} is capable of downloading up to 100 images per style. It uses the associated keywords of each seed image to find and download images from Google Image Search that match the style of the reference image.

\noindent\textit{Steps of the Scraping Process}
\begin{enumerate}
    \item \textit{Keyword Selection:} Manually select distinctive keywords to yield high-quality style images.
    
    \item \textit{Image Downloading:} Input these keywords into Google Image Search to download a large volume of similar style images.
    
    \item \textit{Image Filtering:} Filter the downloaded style images to select the top 40 images. The criteria for filtering include image resolution, relevance to the style keyword, and aesthetic quality.
\end{enumerate}

\begin{figure}[t]
\centering
\hspace*{-1.5cm}
\includegraphics[width=1.2\linewidth]{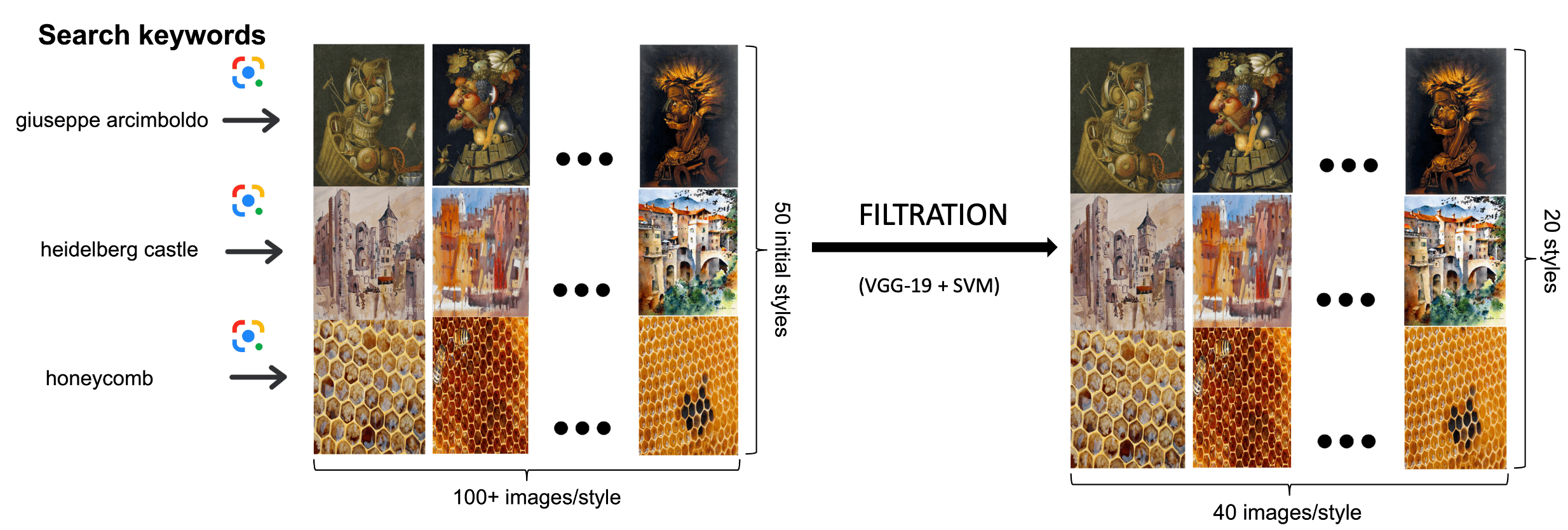}
\caption{The overall scrapping process. On the left, 
process of scraping images corresponding to the Search keywords. On the right, the final dataset after Filtration.}
\label{fig:scraping}
\end{figure}

The top 40 images from each style are then used as inputs for style transfer algorithms. Figure \ref{fig:scraping} shows the scrapping process.\\

\noindent\textit{Data Filtering} A filtration process is applied to our collected style dataset to segregate styles into "good" and "bad" categories. We randomly selected 200 style images and, for each, generated 20 stylized images by applying the style to 20 randomly chosen content images from the Meta-Album, resulting in stylized images (details in Section \ref{sub-sec:style-transfer}). A style is classified as "bad" if any of its 20 stylized images are unrecognizable to the human eye. Conversely, if all stylized images are recognizable, the style is classified as "good". This manual classification process was repeated for the 200 randomly selected styles.

During our investigation of stylized images, we noticed that some styles consistently resulted in multiple unrecognizable stylized images. To tackle this issue, we initiated a manual labeling process. We randomly selected 200 style images and generated 20 stylized images for each. If any of the 20 stylized images created from a single style image were unrecognizable upon human inspection, we classified that style image as "bad". If all 20 stylized images were recognizable, we classified the style image as "good". This procedure resulted in a dataset of 200 style images, each manually labeled as either "good" or "bad".

Subsequently, we utilized our 200 labeled style images to further classify more style images. We generated embeddings for each image by processing them through a pre-trained VGG-19 model \cite{simonyan2015deep}. We then computed the average of the output from the layers: \textit{conv1\_1}, \textit{conv2\_1}, \textit{conv3\_1}, \textit{conv4\_1}, \textit{and conv5\_1}. Using these embeddings, we trained a Support Vector Machine (SVM) model \cite{boser1992training} to classify each image as either "good" or "bad" style.

Finally, using the trained SVM, we filtered the initial pool of 5000 scraped images, retaining only the top 40 images for each style class. From these, we manually selected the 20 best style classes to comprise our final Style dataset.

%% file: appendices/quality-control.tex
\section{Quality Control}
\label{appendix:quality-control}

To ensure quality control, we assessed the degree of information loss (i.e., obscurity) following the application of stylization. We employed a baseline model featuring an EfficientNet V2 backbone combined with a linear projection head.

We created an evaluation set comprising all 25 crawled styles. We conducted ${N}=100$ trials, each involving 5 main steps below:

\begin{enumerate} 
    \item Select randomly three style images from the evaluation set and pairing them with three different content classes from the Meta Album datasets.
    \item Generate stylized images from these combinations and train our baseline model to classify the stylized images into the three content classes. 
    \item Measure the model's test accuracy for classifying stylized images into the above three content classes, which we termed as "\textbf{Stylized Accuracy}".
    \item Train another baseline model to classify the original (non-stylized) content images into the same three content classes.
    \item  Measure the model's test accuracy for classifying original content images into the above three content classes, referring to this metric as "\textbf{Original Accuracy}".
\end{enumerate}

\begin{figure}[ht]
    \centering
    \includegraphics[width=\linewidth]{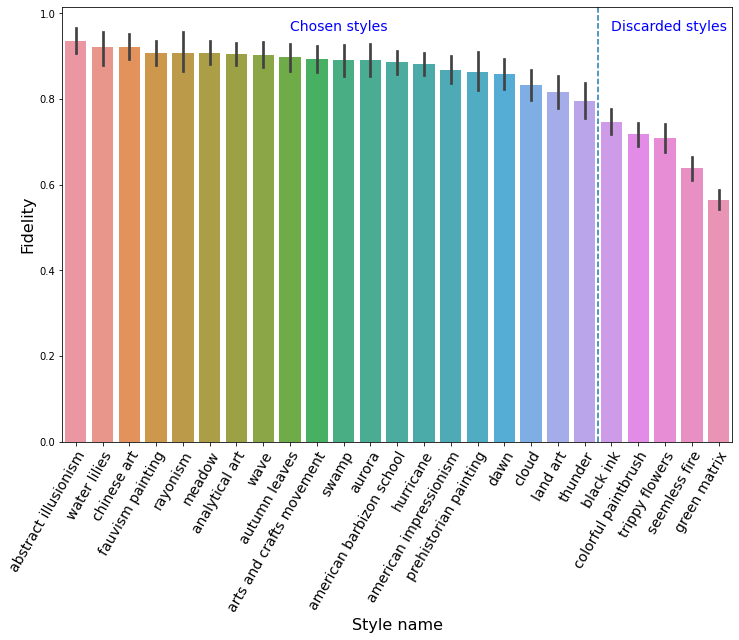}
    \caption{Fidelity of all styles. Blue line indicates separation between Chosen and Discard styles. Left side of blue line consists of all 20 chosen styles for the final dataset.}
    \label{fig:template-verification}
\end{figure}

The ratio:
\[
\frac{\textit{Stylized Accuracy}}{\textit{Original Accuracy}}
\]
indicates the amount of information preserved after style transfer. We quantified the fidelity of each style by averaging all ratios across experiments:
\[
Fidelity = \frac{\text{1}}{\text{N}} \sum_{i=1}^{N} \frac{\textit{Stylized Accuracy}_{i}}{\textit{Original Accuracy}_{i}}
\]
The results are depicted in Figure \ref{fig:template-verification}.

Based on the final results, we selected the top 20 styles with the highest fidelity scores and discarded the remainder. Consequently, our final Stylized Meta Album dataset comprises 20 style classes, each containing 40 images.

%% file: appendices/benchmark-fairness.tex
\section{Group Fairness Benchmark}
\label{appendix:benchmark-fairness}

\subsection{Experimental setup}
\label{appendix:benchmark-fairness:Experimental-setup}
\begin{figure}[ht]
    \centering
    \includegraphics[width=.7\linewidth]{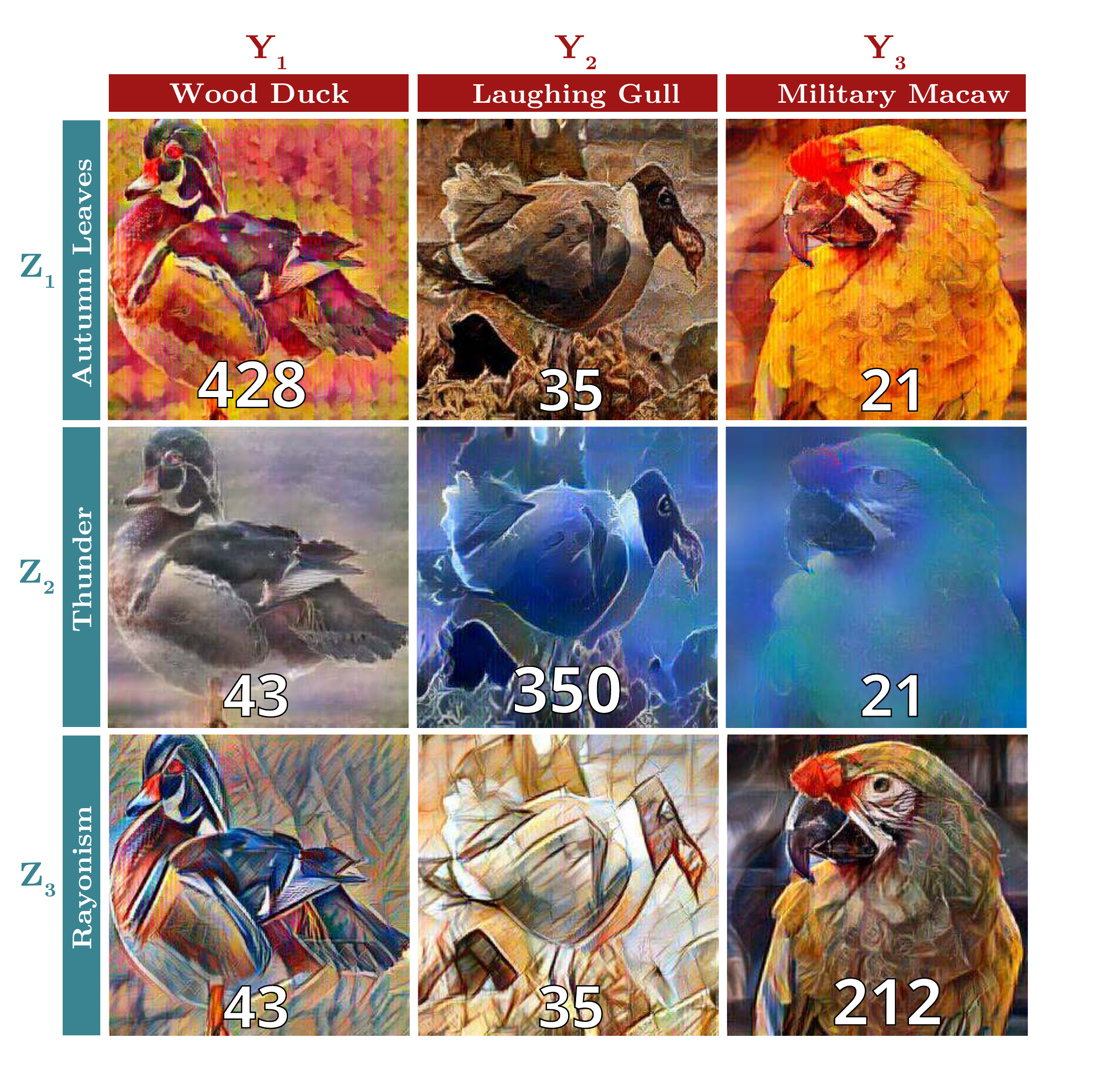}
    \vspace{-20px}
    {\caption{Subsampled {\it SMA\_BRD} training set for $K=3$ and $\mu=10\%$. Each of the 3 content classes has 1 dominant and 2 minority styles (minority examples are $\mu$ of the majority). Example counts are shown in white, with font size proportional to the value. For instance, `WOOD DUCK' has 428 examples in `autumn leaves' style and 43 in each minority style (`Thunder' and `Rayonism'). Majority and minority group sizes vary due to the extra random class content subsampling to create imbalance.}
    \label{appendix:example3x3}}
    
\end{figure}

In this section, we give more details about our experimental setup. As explained, to make the plot of Figure \ref{fig:plt-net} and \ref{fig:benchmark_all_worstgroupaccs}, we create group imbalances in the training set while keeping the validation and test sets balanced. To introduce imbalances within each original SMA dataset, we select only $K$ classes and $K$ styles out of the 20 available for each dataset, resulting in $K^2$ groups (cross-sections of classes and styles). For each class, we designate one style as the "dominant style", creating $K$ "dominant groups" for which we retain all examples. For the remaining $K(K-1)$ "minority groups," we down-sample the dataset by keeping only a fraction $\mu$ of the examples. To mimic real-world scenarios where datasets might have an imbalance at the class level, we introduce another layer of imbalance by uniformly sampling a random subset of examples for each class, ranging from 30\% to 100\%. The figure \ref{appendix:example3x3} provides an example of the type of dataset obtained.

\paragraph{Hyperparameter Optimization:}
For each configuration of the parameters $K$ and $\mu$, and for each method, we use Bayesian optimization with DeepHyper \cite{deephyper_software} to search for optimal hyperparameters. In alignment with the original benchmark paper \cite{idrissi2022simple}, we optimized hyperparameters such as learning rate (between $10^{-4}$ and $5\times 10^{-3}$), weight decay (between $10^{-4}$ and $1$), and batch size (between $8$ and $64$)). Additionally, for JTT, we also optimized two hyperparameters: $T$ (in ${1,3,5}$) and $\lambda_{up}$ (in ${4,20,50,100}$). We use 20 epochs of fine-tuning for all datasets with a Resnet50 backbone initialized with the ImageNet pre-trained weights.

Specifically, we use the validation set to identify the best hyperparameters through 15 runs of the optimization process for experiments of section \ref{subsubsection:benchmark:fairness_result} and 50 runs are used in \ref{subsubsec: Reevaluating Fairness Metrics When Increasing Diversity} to plot Figure \ref{fig:hyperparameter_tuning} to better demonstrate the impact of the hyperparameter optimization phase. This approach allows us to efficiently navigate the hyperparameter space and converge on a set of values appropriate for each method and dataset subsampling.

Once the best hyperparameters are identified using the validation set, we conduct five additional training runs using the previously found hyperparameters to ensure robustness and mitigate the effects of any random fluctuations. Additionally, to plot Figure \ref{fig:hyperparameter_tuning}, we use 10 different subsamples of the $\textit{SMA\_TEX\_DTD}$ dataset (totaling 50 runs), featuring various styles and content classes of the dataset to ensure better statistical significance. By leveraging these diverse dataset subsamples, we aim to demonstrate that tuning our model on top-3 worst-group accuracy is generalizable across different styles and content classes. The final performance is then assessed on the test set, where we report the metric that yielded the best result during the validation phase. All the error bars on our graphs correspond to the standard error over these final runs on the test set.

\subsection{Fairness results on all datasets}

\label{appendix:benchmark-fairness:Results-on-all-datasets}
\begin{figure}[htbp]
    \centering
    \begin{tabular}{c c c}
        \multirow{2}{*}[5em]{\rotatebox[origin=c]{90}{Airplanes}} &
        \begin{subfigure}{0.45\textwidth}
            \includegraphics[width=\linewidth]{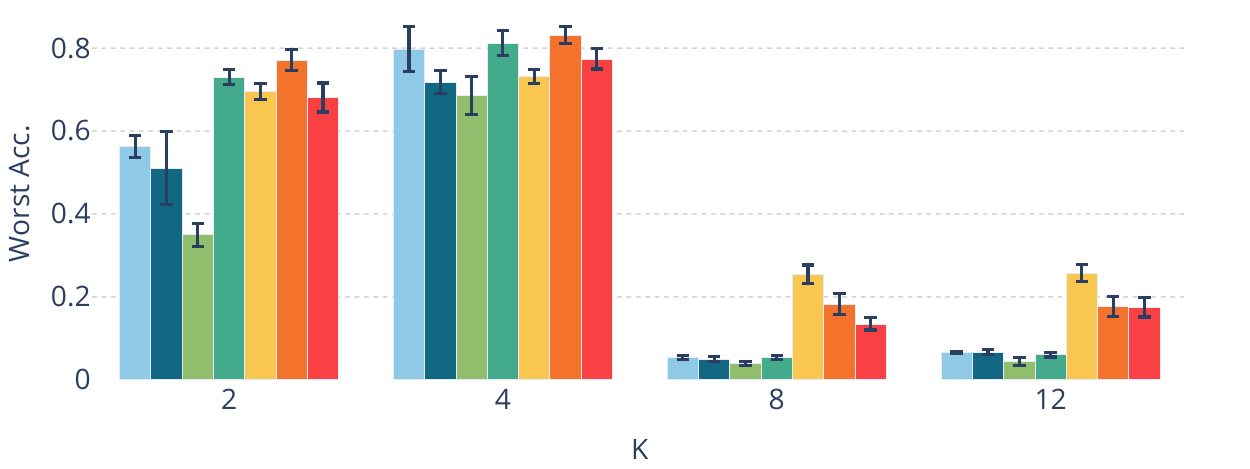}
        \end{subfigure} &
        \begin{subfigure}{0.45\textwidth}
            \includegraphics[width=\linewidth]{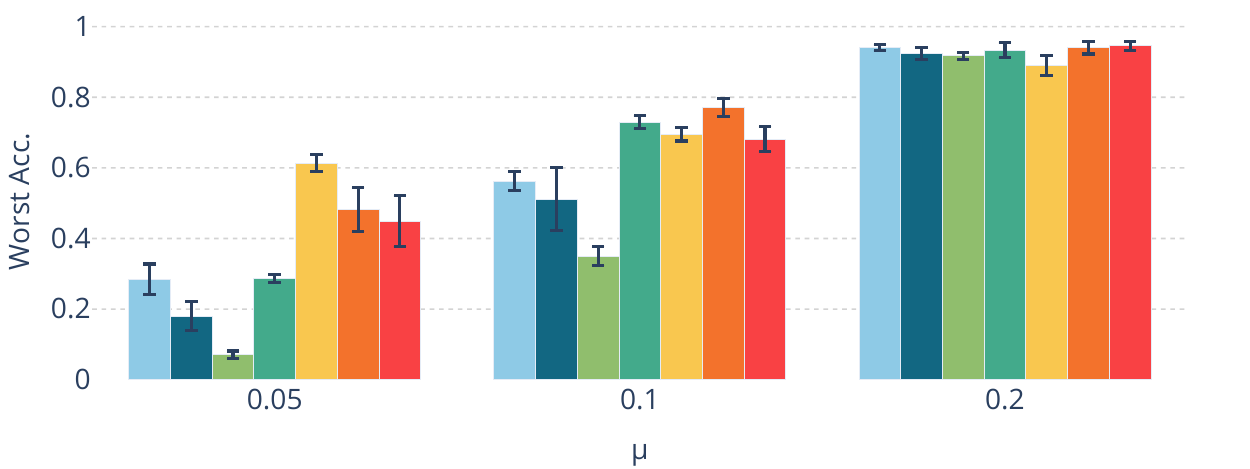}
        \end{subfigure} \\
    \end{tabular}

    \begin{tabular}{c c c}
        \multirow{2}{*}[4em]{\rotatebox[origin=c]{90}{Bird}} &
        \begin{subfigure}{0.45\textwidth}
            \includegraphics[width=\linewidth]{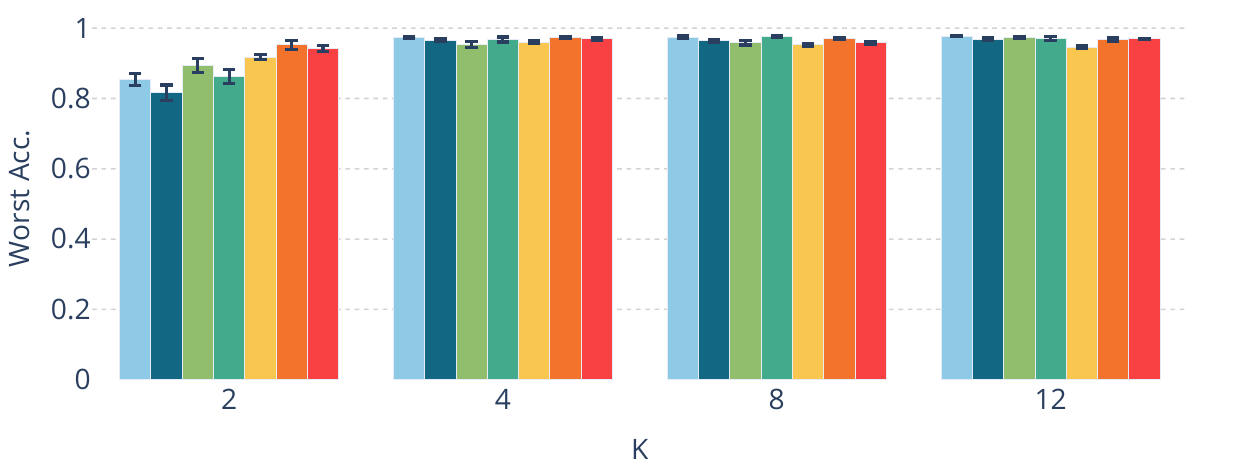}
        \end{subfigure} &
        \begin{subfigure}{0.45\textwidth}
            \includegraphics[width=\linewidth]{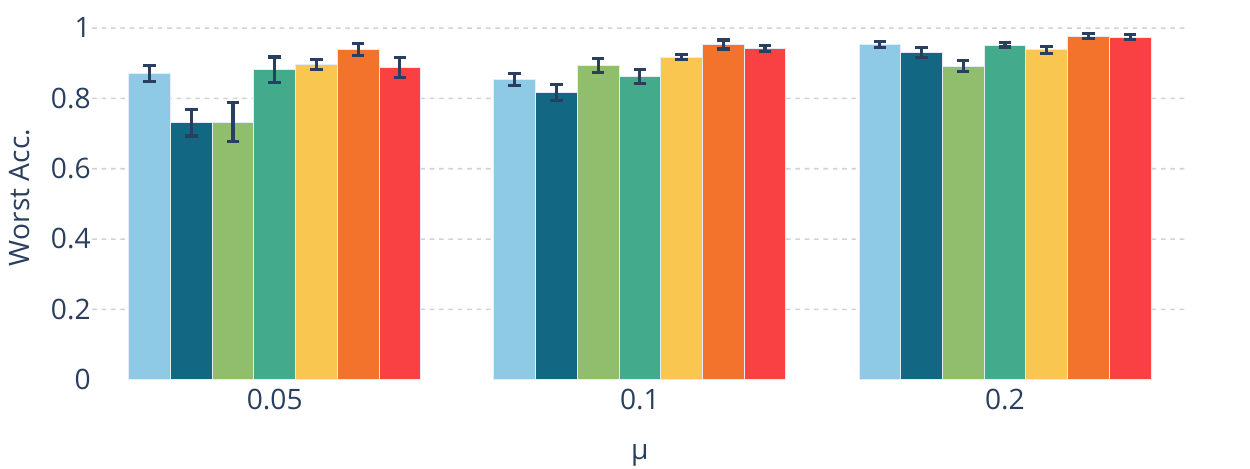}
        \end{subfigure} \\
    \end{tabular}

    \begin{tabular}{c c c}
        \multirow{2}{*}[4em]{\rotatebox[origin=c]{90}{Dogs}} &
        \begin{subfigure}{0.45\textwidth}
            \includegraphics[width=\linewidth]{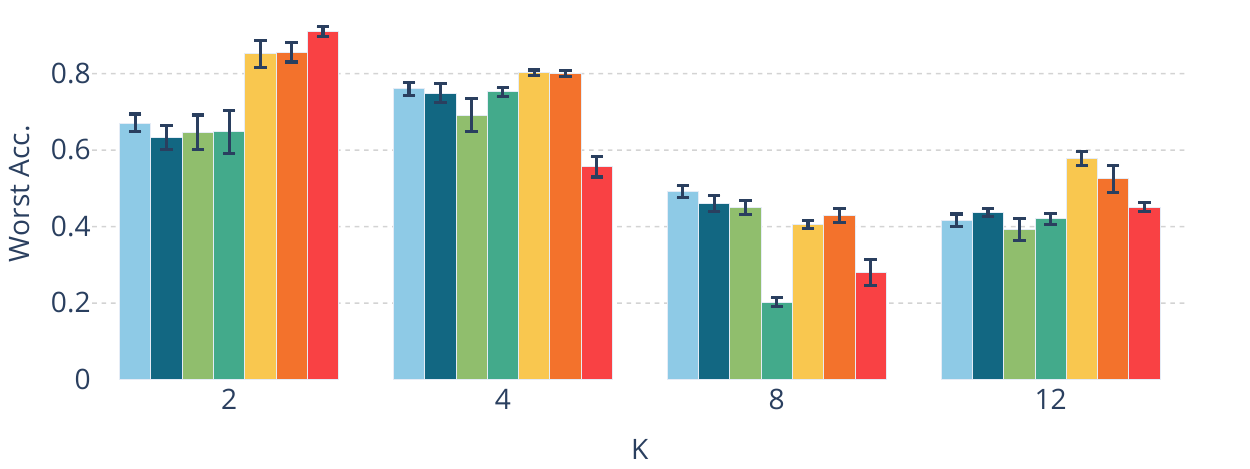}
        \end{subfigure} &
        \begin{subfigure}{0.45\textwidth}
            \includegraphics[width=\linewidth]{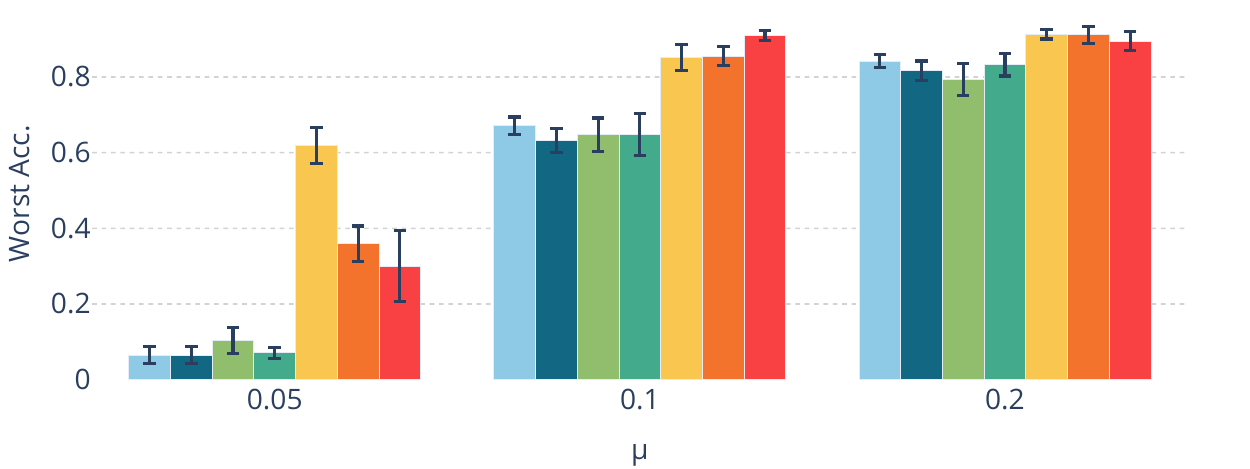}
        \end{subfigure} \\
    \end{tabular}

    \begin{tabular}{c c c}
        \multirow{2}{*}[5.5em]{\rotatebox[origin=c]{90}{Medical-Leaf}} &
        \begin{subfigure}{0.45\textwidth}
            \includegraphics[width=\linewidth]{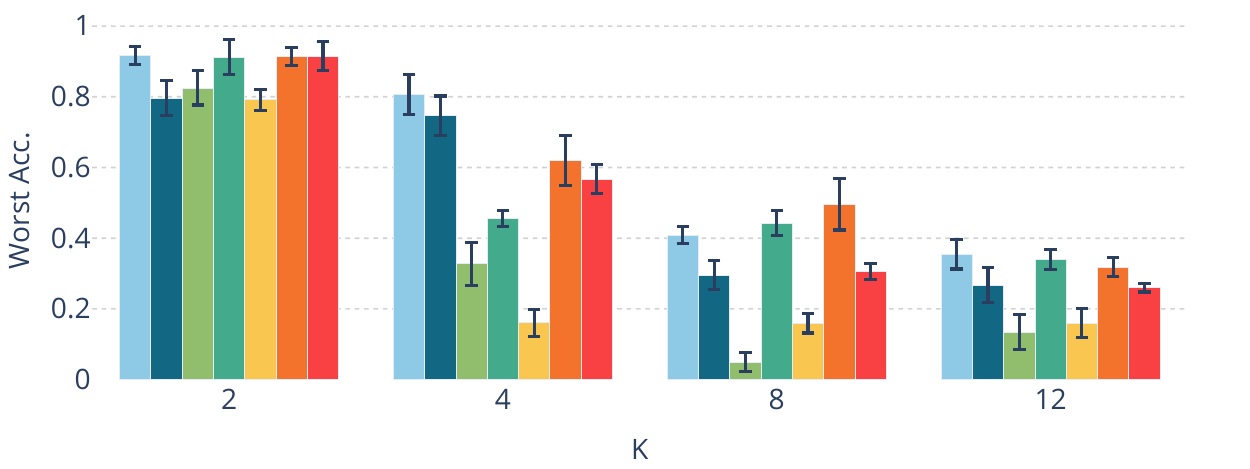}
        \end{subfigure} &
        \begin{subfigure}{0.45\textwidth}
            \includegraphics[width=\linewidth]{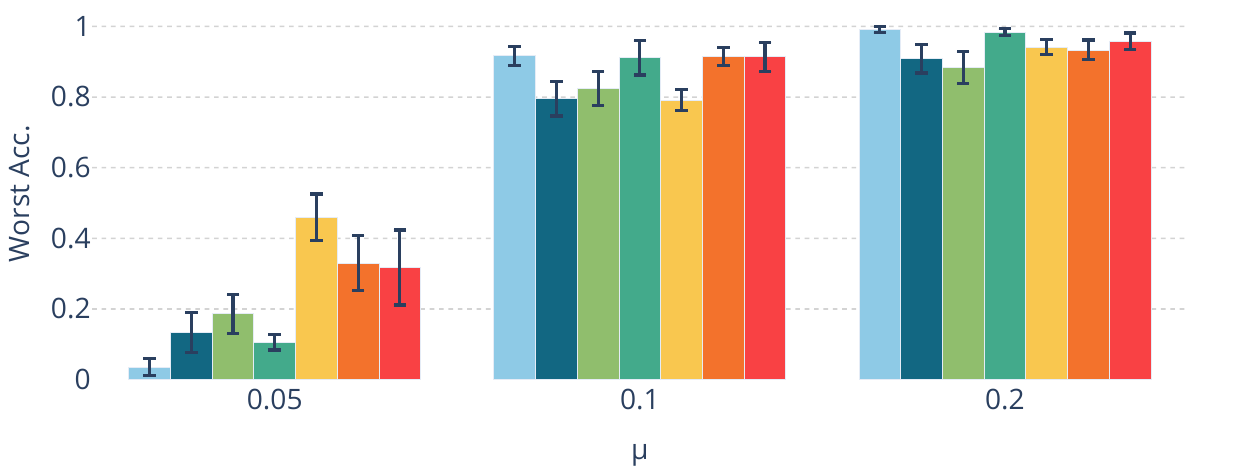}
        \end{subfigure} \\
    \end{tabular}

    \begin{tabular}{c c c}
        \multirow{2}{*}[4.5em]{\rotatebox[origin=c]{90}{Plt-Doc}} &
        \begin{subfigure}{0.45\textwidth}
            \includegraphics[width=\linewidth]{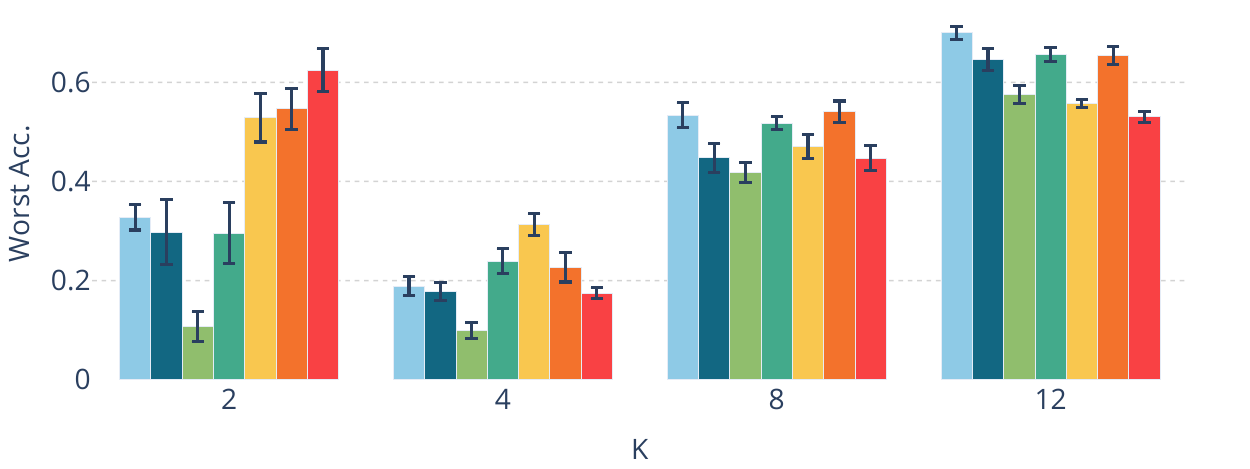}
        \end{subfigure} &
        \begin{subfigure}{0.45\textwidth}
            \includegraphics[width=\linewidth]{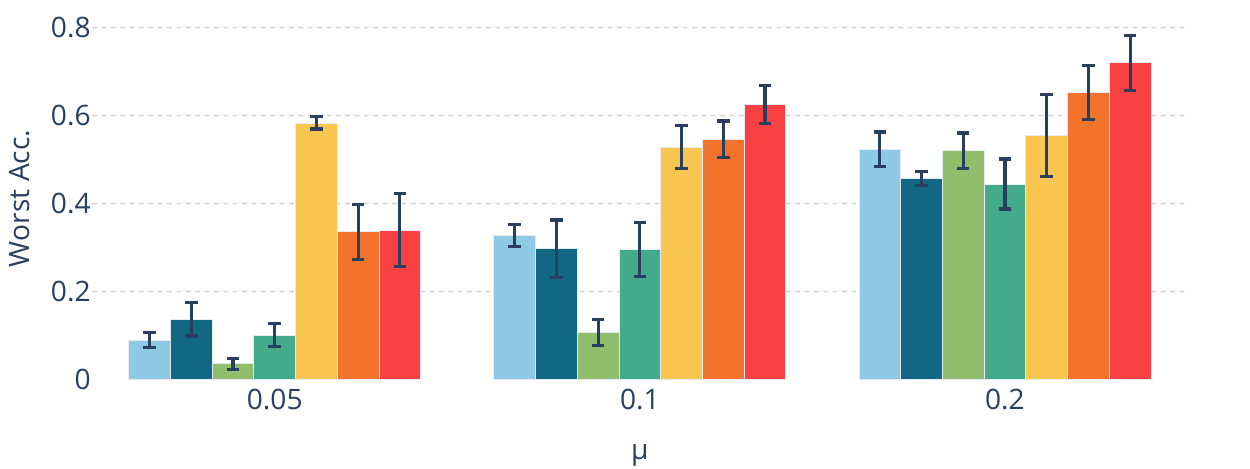}
        \end{subfigure} \\
    \end{tabular}

    \begin{tabular}{c c c}
        \multirow{2}{*}[4em]{\rotatebox[origin=c]{90}{Resisc}} &
        \begin{subfigure}{0.45\textwidth}
            \includegraphics[width=\linewidth]{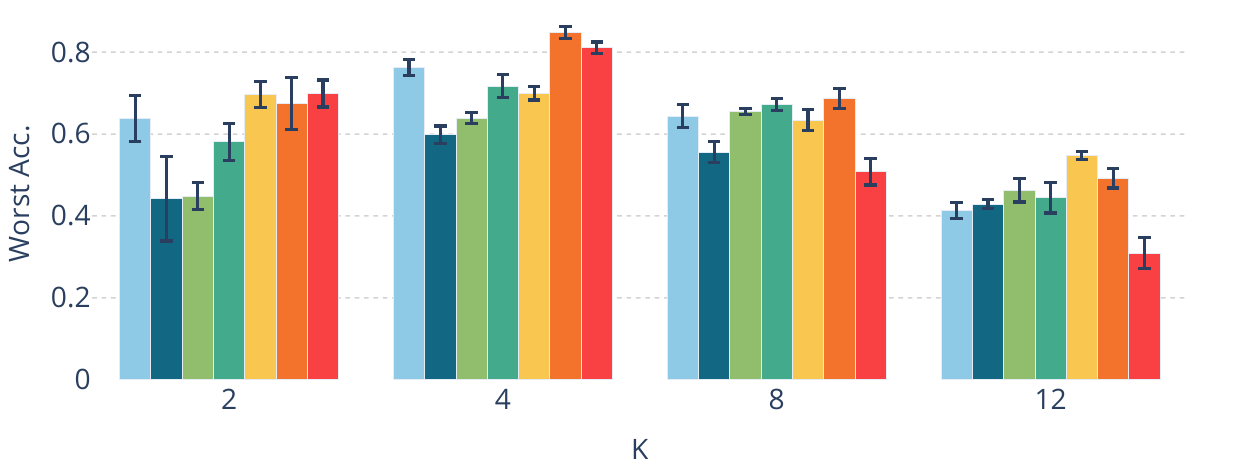}
        \end{subfigure} &
        \begin{subfigure}{0.45\textwidth}
            \includegraphics[width=\linewidth]{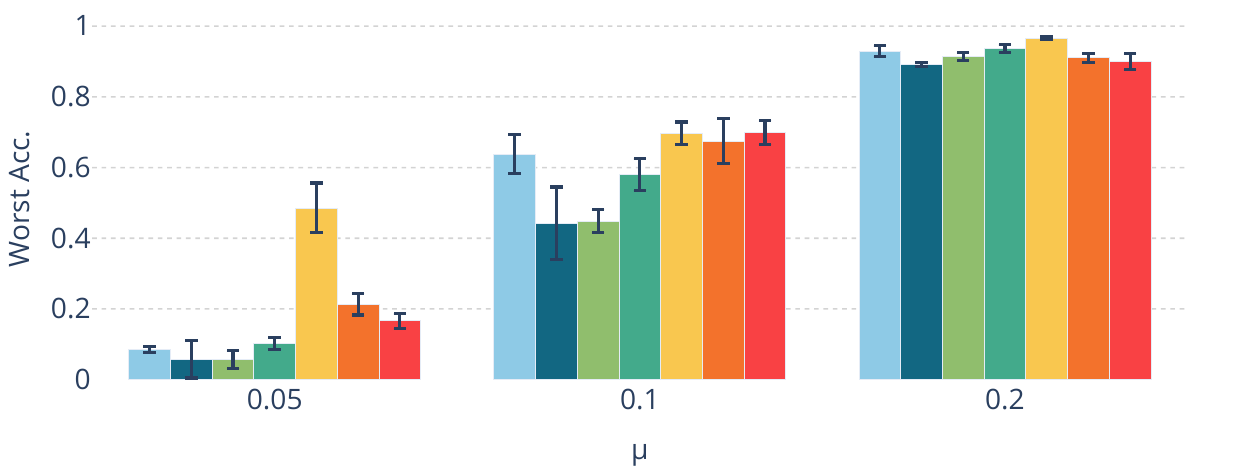}
        \end{subfigure} \\
    \end{tabular}


    \begin{tabular}{c c c}
        \multirow{2}{*}[5em]{\rotatebox[origin=c]{90}{Texture\_DTD}} &
        \begin{subfigure}{0.45\textwidth}
            \includegraphics[width=\linewidth]{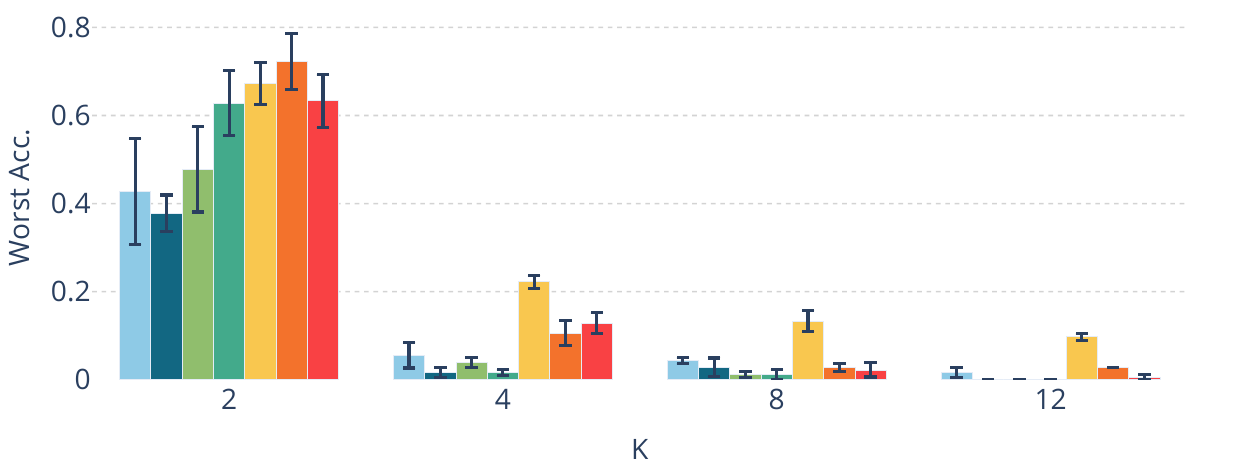}
        \end{subfigure} &
        \begin{subfigure}{0.45\textwidth}
            \includegraphics[width=\linewidth]{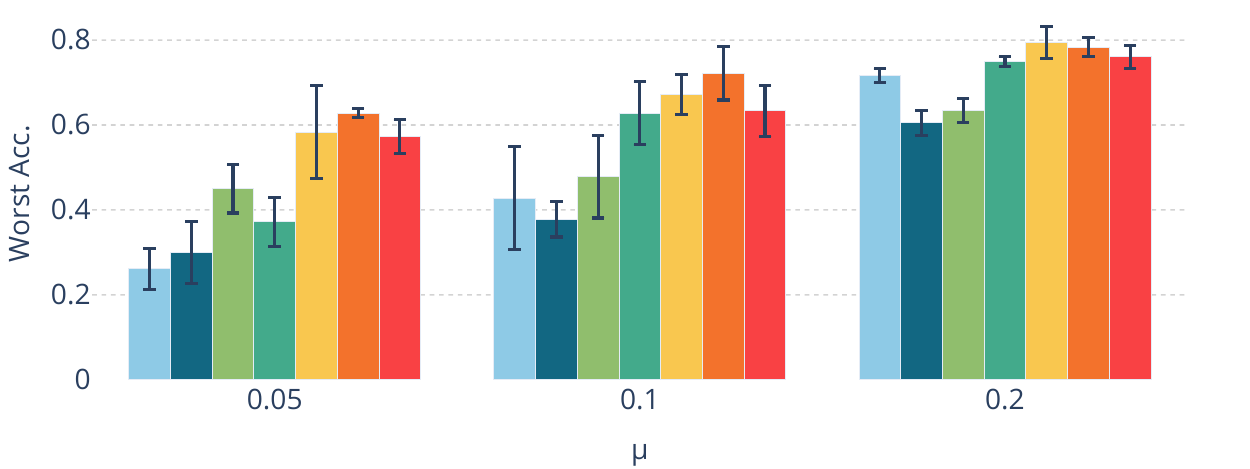}
        \end{subfigure} \\
    \end{tabular}
    
    \caption{Worst-group accuracy for different SMA datasets: (left) varying $K$ values with $\mu=10\%$ and (right) varying $\mu$ values with $K=2$. Experiments are conducted with seven methods (from left bars to right: ERM, JTT, SUBY, RWY, SUBG, RWG, DRO).}
    \label{fig:benchmark_all_worstgroupaccs}
\end{figure}

\begin{figure}[ht]
    \centering
    \includegraphics[width=\linewidth]{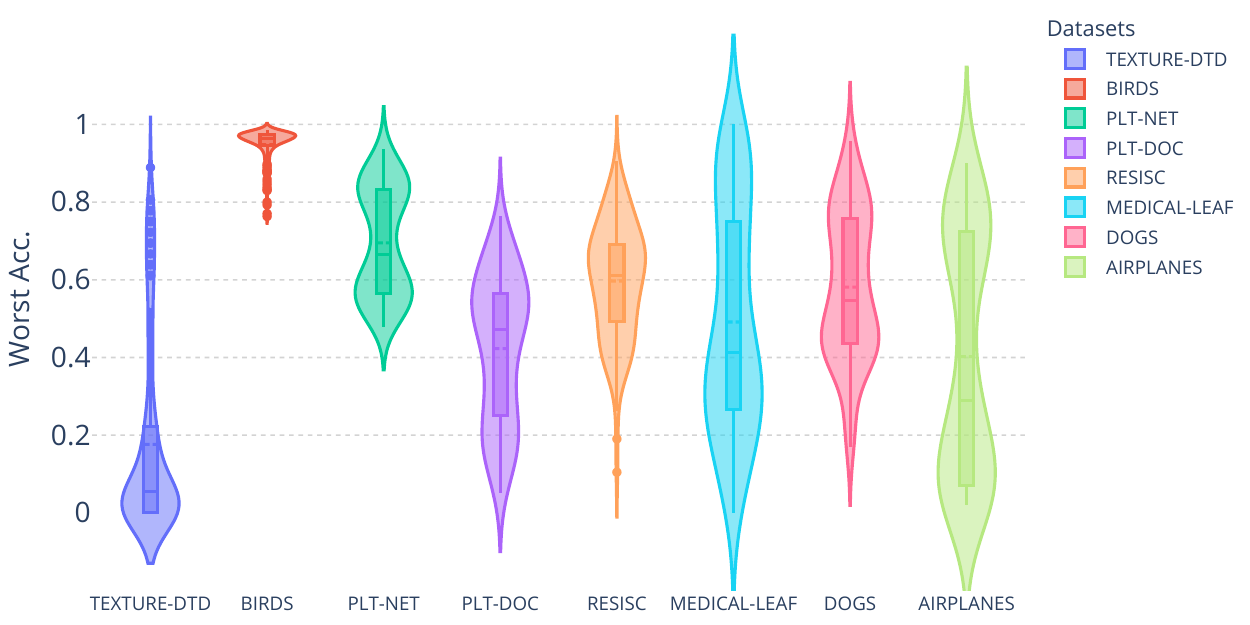}
    \caption{Worst-group accuracy distribution across datasets for different subsampling of the datasets with $K\in\{2,4,8,12\}$ for all 7 methods and $\mu=10\%$. Each SMA dataset exhibits intrinsic complexities that affect the difficulty of group fairness problem}
    \label{violin_avgacc}
\end{figure}

We experimented with a total of 8 SMA datasets.  Figure \ref{fig:benchmark_all_worstgroupaccs} presents the results (worst-group accuracy) as a function of $K$ and $\mu$ for the remaining 7 datasets. Figure \ref{violin_avgacc} illustrates the diverse difficulty levels of these datasets, demonstrating our effort to ensure the benchmark encompasses a variety of dataset domains and difficulty levels.We now give more details and add extra precisions about the analysis of the results of these experiments across 8 SMA datasets. Firstly, we can see in the second column of Figure \ref{fig:benchmark_all_worstgroupaccs}, that for all SMA datasets, when $\mu$ is smaller (e.g., $\mu = 0.05$), performance drops sharply across all methods, indicating a strong bias of the models toward the majority classes. In these cases, methods using group information generally outperform others, with SUBG almost always achieving the best results (6 datasets over 8), often by a large margin. Conversely, as $\mu$ increases and the dataset becomes more balanced, bias is reduced, leading to more uniform performance across methods. Interestingly, changes in $\mu$ not only affect the overall performance of the methods but also alter their relative rankings. For example, although DRO achieves the best worst-group accuracy for \textit{SMA\_DOGS} and \textit{SMA\_PLT-DOC}, it is outperformed by SUBG when $K=8$. This variability highlights the importance of understanding how different methods respond to changes in dataset composition, particularly regarding class and group imbalances.

Similarly $\mu$ is fixed at $10\%$, the number of groups $K$ (first column of Figure \ref{fig:benchmark_all_worstgroupaccs}) has a substantial impact on the relative performance of the various methods. Notably, for datasets such as \textit{SMA\_TEXTURE-DTD} and \textit{SMA\_AIRPLANES}, we observe that the worst-group accuracy drops to near or exactly 0 as diversity ($K$) increases. This effect is likely attributable to the smaller size of these datasets, which exhibit greater sensitivity to group-level variability and higher variance in group performance. Moreover, our findings reinforce the surprising conclusion of \cite{idrissi2022simple} that simple data balancing techniques remain competitive with more complex approaches. Specifically, among the methods that do not utilize group information, simple balancing ones like SUBY and RWY, perform comparably or better than JTT. 
Indeed, across the seven datasets shown in this figure, JTT never significantly outperforms the two simple balancing methods or ERM. 
Similarly, among group-aware methods, simple balancing approaches like RWG and SUBG achieve performance comparable to methods like DRO, which explicitly optimize for worst-case group performance, across nearly all datasets.

Our second key observation was that increasing diversity improves fairness. Indeed, as $K$ grows, we can observe that ERM achieves worst-group accuracy levels comparable to other methods, despite not explicitly addressing bias. Notably, for methods that do not use group information, such as JTT, SUBY, or RWY,
none significantly outperform ERM at $K=12$ across the 8 tested SMA datasets. We hypothesize that the broader exposure to diverse examples within each class, brought about by higher $K$, encourages models to generalize better and reduces their reliance on spurious correlations present in the training data. These findings collectively highlight the intricate dynamics between dataset diversity, group imbalance, and method performance, underscoring the importance of systematically exploring such factors to optimize fairness and robustness in machine learning models.

\subsection{Top-M worst-group accuracy on all datasets }
\label{appendix:topm}

\begin{table}[ht]
\caption{Test
worst-group accuracy across SMA datasets for varying values of K for the SUBG methods, based on different model selection metrics during the validation phase: simple accu-
racy (Acc.), worst-group accuracy (Worst Acc.), and Top-3 worst-group accuracy (Top-3 Worst Acc.) Each result corresponds to 10 distinct subsampling of the datasets (with different classes and styles selected), with 5 random seeds per subsampling, totaling 50 runs per configuration. This ensures that each variant of a dataset represents 10  unique dataset configurations, with their own specific class-style associations and imbalances. The mean performance is reported with the standard error of the mean (SEM) indicated with $\pm$ and the best results for each configuration are highlighted in \textbf{bold}. Notably, for a high diversity scenario (K=8), our proposed Top-3 worst-group accuracy metric almost always leads to better fairness results.}

\centering
\renewcommand{\arraystretch}{1.2} 
\begin{tabular}{ccccc}
\toprule
\multicolumn{2}{c}{} & \multicolumn{3}{c}{\textbf{Model Selection Metrics}} \\
\cmidrule(lr){3-5}
\textbf{Dataset} & \textbf{$K$} & \textbf{Acc.} & \textbf{Worst Acc.} & \textbf{Top-3 Worst Acc.} \\
\midrule
\multirow{3}{*}{Airplanes}    
& 2 & 50.87 $\pm$ \scriptsize{4.3} & 52.64 $\pm$ \scriptsize{4.0} & \textbf{55.22 $\pm$ \scriptsize{5.0}} \\
& 4 & \textbf{76.40 $\pm$ \scriptsize{1.7}} & 69.44 $\pm$ \scriptsize{2.0} & 70.74 $\pm$ \scriptsize{2.7} \\
& 8 & 19.84 $\pm$ \scriptsize{2.1} & 21.46 $\pm$ \scriptsize{1.1} & \textbf{22.73 $\pm$ \scriptsize{2.3}} \\
\cline{1-5}
\multirow{3}{*}{Birds}      
& 2 & 84.04 $\pm$ \scriptsize{2.4} & 84.15 $\pm$ \scriptsize{2.3} & \textbf{84.86 $\pm$ \scriptsize{2.4}} \\
& 4 & 92.94 $\pm$ \scriptsize{0.6} & \textbf{93.51 $\pm$ \scriptsize{0.5}} & 92.05 $\pm$ \scriptsize{0.8} \\
& 8 & 85.33 $\pm$ \scriptsize{3.4} & 84.17 $\pm$ \scriptsize{3.2} & \textbf{94.16 $\pm$ \scriptsize{0.2}} \\
\cline{1-5}
\multirow{3}{*}{Dogs}      
& 2 & 45.20 $\pm$ \scriptsize{4.3} & 46.54 $\pm$ \scriptsize{4.0} & \textbf{48.30 $\pm$ \scriptsize{4.0}} \\
& 4 & \textbf{53.11 $\pm$ \scriptsize{2.7}} & 50.09 $\pm$ \scriptsize{3.9} & 49.67 $\pm$ \scriptsize{3.9} \\
& 8 & 31.15 $\pm$ \scriptsize{1.4} & \textbf{34.86 $\pm$ \scriptsize{1.5}} & 34.34 $\pm$ \scriptsize{1.4} \\
\cline{1-5}
\multirow{3}{*}{Plt-Net}      
& 2 & 81.63 $\pm$ \scriptsize{0.9} & 78.58 $\pm$ \scriptsize{1.6} & \textbf{82.50 $\pm$ \scriptsize{1.0}} \\
& 4 & 80.65 $\pm$ \scriptsize{1.5} & \textbf{83.23 $\pm$ \scriptsize{0.4}} & 82.95 $\pm$ \scriptsize{0.9} \\
& 8 & 64.33 $\pm$ \scriptsize{1.6} & 59.53 $\pm$ \scriptsize{1.0} & \textbf{66.87 $\pm$ \scriptsize{1.0}} \\
\cline{1-5}
\multirow{3}{*}{Resisc}      
& 2 & 65.94 $\pm$ \scriptsize{1.9} & \textbf{66.95 $\pm$ \scriptsize{2.6}} & 63.64 $\pm$ \scriptsize{3.2} \\
& 4 & 61.96 $\pm$ \scriptsize{2.6} & 58.25 $\pm$ \scriptsize{2.7} & \textbf{63.20 $\pm$ \scriptsize{2.4}} \\
& 8 & 56.86 $\pm$ \scriptsize{1.1} & 55.52 $\pm$ \scriptsize{1.6} & \textbf{59.05 $\pm$ \scriptsize{1.2}} \\
\cline{1-5}
\multirow{3}{*}{Texture-DTD}      
& 2 & 46.32 $\pm$ \scriptsize{5.4} &  \textbf{54.33 $\pm$ \scriptsize{4.7}}&50.11 $\pm$ \scriptsize{5.1} \\
& 4 & 30.05 $\pm$ \scriptsize{2.9} & 33.33 $\pm$ \scriptsize{3.1} & \textbf{40.02 $\pm$ \scriptsize{2.5}} \\
& 8 & 5.42 $\pm$ \scriptsize{0.8} & 4.3 $\pm$ \scriptsize{0.9} & \textbf{6.63 $\pm$ \scriptsize{1.0}} \\
\bottomrule
\end{tabular}
\label{tab:selector}
\end{table}

We extend the findings presented in Figure  \ref{fig:hyperparameter_tuning} to additional SMA datasets, as summarized in Table \ref{tab:selector}. As we did previously, for each dataset, we perform 10 subsampling for each dataset, to account for the significant variability in results that can arise depending on the specific selected combinations of classes and styles. We can observe in Table \ref{tab:selector} that for high-diversity scenarios (larger $K$), using Top-3 Worst-Group Accuracy as the model selection metric consistently leads to improved fairness outcomes except on Dogs dataset. Surprisingly, even on datasets with more examples like PLT-Net or Birds, using the Top-3 worst group accuracy also improves significantly the results for K=8. This suggests that averaging over the three worst-performing groups provides a more stable and less noisy criterion for model selection in high-diversity settings. For $K=2$,  there are only 4 groups, so the Top-3 worst-group accuracy closely approximates the standard accuracy, which averages performance across all four groups. This explains why no significant differences are observed between the two metrics across all datasets in this setting.

\subsection{Worst-group accuracy statistical properties}
\label{appendix:benchmark-fairness:statistical_property}

In our study, using higher values of $K$ for both styles and classes results in $K^2$ different groups. We try to show the statistical effect through which, the worst-group accuracy, which focuses only on the minimum over a quadratically increasing number of groups is inherently subject to a significant drop in expectation as K increases. We can model the result of the accuracy of the $K^2$ groups by i.i.d.  variables $X_1, \ldots, X_{K^2} $. We consider $Y_k=\min (X_1,...,X_{K^2})$.
 
An initial straightforward argument consists in noticing that $Y_K$ is a decreasing function of $K$ and it will thus converge in probability towards the infimum of $X$. This shows that increasing the number of groups can only hinder worst-group accuracy.

For instance, if we chose to model the result of the accuracy of the $ K^2 $ groups by i.i.d.  random variables $ X_1, \ldots, X_{K^2} $, with the same expectation $ \mu $, standard deviation $ \sigma $,  it can be shown that the expectation of the minimum is the following:

\begin{equation}
    \mathbb{E}(Y_K)= \mu - \sigma E_K
\end{equation}
where $E_K = \sqrt{2*\log(K^2)}$, which makes the expectation a decreasing function of $K$. In our experimental setup, we can have $K=20$. 

Moreover, the assumption of uniformity is often not met in practice. The likelihood that one group is highly biased and harder to classify also increases as $K$ grows, meaning it only takes one particularly difficult-to-classify group to significantly lower the worst-group accuracy, which is even more likely to happen for smaller datasets with a relatively low number of examples per group.

As a result, metrics overly sensitive to the worst-case performance may not reliably reflect the algorithm's overall fairness, especially when the number of groups is large and the group sizes are small. To mitigate this issue and ensure a more stable and representative assessment of fairness, one could use other metrics like top-M worst-group accuracy.

%% file: appendices/UDA-detailed-results.tex
\section{UDA benchmark}
\label{appendix:usecase-UDA}
\subsection{Detailed introduction of UDA}
To tackle learning amidst distribution shift, UDA algorithms utilize a labeled source domain \( \mathcal{D}_s = \{(x_i^s, y_i^s)\}_{i=0}^{n_s} \) and an unlabeled target domain \( \mathcal{D}_t = \{x_i^t\}_{i=0}^{n_t} \). These domains are sampled from joint distributions \( p_s(x, y) \) and \( p_t(x, y) \), respectively, representing different styles within the SMA. Most UDA models \cite{CDAN, DANCE, DANN, uniOT, DSAN} consist of a feature extractor \( g \) and a classifier \( f \), aiming to align the source and target feature spaces by minimizing the discrepancy between the source features \( g(x^s) \) and target features \( g(x^t) \). 

Various approaches achieve this alignment; for instance, one can minimize the distance between feature distributions using distances like Maximum Mean Discrepancy (MMD) \cite{da_mmd, DSAN}, Wasserstein between marginals \cite{OTDA}, or even between joint distributions as seen in \textbf{DeepJDOT} \cite{JDOT, deepJDOT}. The \textbf{DANN} approach \cite{DANN} employs adversarial learning to foster domain-invariant features, thereby reducing the discrepancy between source and target features. In \textbf{CDAN} \cite{CDAN}, authors enhance this method by incorporating the cross-covariance between feature representations and classifier predictions as input for the feature extractor.

\textbf{Closed-Set Domain Adaptation vs. Universal Domain Adaptation} Most existing methods in domain adaptation operate under a closed-set assumption, where it's assumed that the source label set \( \mathcal{C}_s \) and the target label set \( \mathcal{C}_t \) are identical (\( \mathcal{C}_s = \mathcal{C}_t \)). However, this assumption becomes untenable when dealing with unlabeled target domains, as we cannot guarantee such label set equality. Thus, universal domain adaptation (UniDA) emerges, with no prior assumption made regarding the distribution of the label set. This induces the existence of multiple label subset : 

\begin{itemize}
    \item Common Label Set \( Y = Y_s \cap Y_t \) are the set of shared label between source and target.
    \item Private Source Label Set \( \overline{Y}_s = Y_s \backslash Y\) are label only present in source domain.
    \item Private Target Label Set : \( \overline{Y}_t = Y_t \backslash Y \) are label only present in target domain. Often referred as unknown classes as the labels are not available in target domain.
\end{itemize}

The objective shifts to classifying target samples into \( |Y_s|+1 \) classes: either one of the \( |Y_s| \) source classes or as an unknown class.

\textbf{UDA} \cite{UDA_you} pioneered the concept of UniDA, employing an adversarial approach to align common classes. It introduced a separate non-adversarial discriminator to compute sample-level transferability scores, facilitating the differentiation of common classes from private ones. \textbf{OSBP} \cite{OSBP} also adopts an adversarial approach, directly instructing the classifier to classify target data either as one of the common classes or as unknown, a departure from methods that initially classify all targets and then exclude out-of-distribution classes. \textbf{OVANet} \cite{OVANet} introduces a One-vs-All classifier to ensure that private classes are never misclassified as positive instances by any binary One-VS-All classifiers. \textbf{UniOT} \cite{uniOT} leverages optimal transport principles, using Unbalanced OT to identify samples deemed too costly to transport as unknown target classes.\\

In the case of SMA one could consider the styles as domains, thus the 12 datasets of the SMA could be used separately to evaluate UDA and UniDA algorithms performances with a total of 20 Domains since we have 20 styles for every content dataset.

\subsection{UDA Detailed Results}
\label{appendix:UDA-detailed-results}

\subsubsection{Closed-Set Results}
\input{tables/close-set-UDA}

\subsubsection{UniDA Results}
\input{tables/Universal-DA}

%% file: tables/close-set-UDA.tex
\begin{table}[h]
\centering
\resizebox{\linewidth}{!}{
\begin{tabular}{|c|c|c|c|c|}
\hline
\textbf{Scenario} & \textbf{NO\_ADAPT} & \textbf{DeepJDOT} & \textbf{DANN} & \textbf{CDAN}\\
\hline 
$BIRDS~~american\_imp... \rightarrow american\_bar...$ & 83.58 & 93.12 & \textbf{93.56} & 92.72 \\
$BIRDS~~american\_imp... \rightarrow hurricane$ & 65.98 & \textbf{86.32} & \textbf{86.32} & 86.01 \\
$BIRDS~~analytical\_a... \rightarrow autumn\_leave...$ & 88.48 & 94.13 & 86.58 & \textbf{94.35} \\
$BIRDS~~analytical\_a... \rightarrow thunder$ & 64.39 & 80.54 & 83.50 & \textbf{88.04} \\
$BIRDS~~arts\_and\_cra... \rightarrow dawn$ & 40.16 & 68.14 & 77.93 & \textbf{86.89} \\
$BIRDS~~aurora \rightarrow american\_imp...$ & 75.60 & 86.05 & 90.69 & \textbf{92.81} \\
$BIRDS~~autumn\_leave... \rightarrow abstract\_ill...$ & 82.83 & 90.20 & 90.95 & \textbf{92.37} \\
$BIRDS~~chinese\_art \rightarrow land\_art$ & 84.25 & 87.42 & 86.28 & \textbf{93.65} \\
$BIRDS~~dawn \rightarrow abstract\_ill...$ & 76.65 & 93.29 & 89.28 & \textbf{93.51} \\
$BIRDS~~dawn \rightarrow american\_bar...$ & 79.04 & 91.35 & 88.35 & \textbf{91.88} \\
$BIRDS~~fauvism\_pain... \rightarrow abstract\_ill...$ & 84.82 & 93.95 & 92.85 & \textbf{96.56} \\
$BIRDS~~fauvism\_pain... \rightarrow rayonism$ & 83.36 & 89.45 & 88.08 & \textbf{94.44} \\
$BIRDS~~fauvism\_pain... \rightarrow swamp$ & 71.76 & 88.53 & 89.50 & \textbf{91.39} \\
$BIRDS~~hurricane \rightarrow aurora$ & 83.45 & 90.78 & 91.26 & \textbf{93.38} \\
$BIRDS~~hurricane \rightarrow autumn\_leave...$ & 71.80 & 86.89 & 84.25 & \textbf{94.40} \\
$BIRDS~~hurricane \rightarrow cloud$ & 71.32 & \textbf{89.81} & 89.10 & 86.94 \\
$BIRDS~~hurricane \rightarrow thunder$ & 70.04 & 84.02 & 88.17 & \textbf{91.17} \\
$BIRDS~~land\_art \rightarrow abstract\_ill...$ & 87.91 & 92.10 & 89.10 & \textbf{94.97} \\
$BIRDS~~land\_art \rightarrow american\_imp...$ & 85.53 & 90.20 & 91.09 & \textbf{92.19} \\
$BIRDS~~land\_art \rightarrow cloud$ & 74.23 & 88.97 & 85.17 & \textbf{89.98} \\
$BIRDS~~land\_art \rightarrow wave$ & 83.72 & 86.23 & 92.23 & \textbf{93.82} \\
$BIRDS~~meadow \rightarrow abstract\_ill...$ & 82.74 & 92.10 & 86.14 & \textbf{94.88} \\
$BIRDS~~meadow \rightarrow chinese\_art$ & 81.82 & \textbf{92.94} & 86.76 & 92.81 \\
$BIRDS~~meadow \rightarrow prehistorian$ & 79.35 & 90.34 & 88.92 & \textbf{92.67} \\
$BIRDS~~prehistorian \rightarrow american\_bar...$ & 76.92 & \textbf{92.81} & 89.59 & 92.32 \\
\hdashline
BIRDS mean & 77.19 $\pm$ 2.04 & 88.79 $\pm$ 1.09 & 88.23 $\pm$ 0.68 & \textbf{92.17 $\pm$ 0.54} \\
\hline
\end{tabular}
}
\caption{Closed Set Domain Adaptation Accuracy over BIRDS}
\label{tab:acc-BIRDS}
\end{table}

\begin{table}[h]
\centering
\resizebox{\linewidth}{!}{
\begin{tabular}{|c|c|c|c|c|}
\hline
\textbf{Scenario} & \textbf{NO\_ADAPT} & \textbf{DeepJDOT} & \textbf{DANN} & \textbf{CDAN}\\
\hline 
$DOGS~~american\_imp... \rightarrow american\_bar...$ & 78.65 & 83.74 & \textbf{83.93} & 83.36 \\
$DOGS~~american\_imp... \rightarrow hurricane$ & 67.53 & 76.15 & 73.66 & \textbf{78.13} \\
$DOGS~~analytical\_a... \rightarrow autumn\_leave...$ & 66.12 & \textbf{88.78} & 87.09 & 85.86 \\
$DOGS~~analytical\_a... \rightarrow thunder$ & 31.62 & 67.11 & \textbf{68.00} & 64.18 \\
$DOGS~~arts\_and\_cra... \rightarrow dawn$ & 28.23 & 61.55 & 64.04 & \textbf{69.32} \\
$DOGS~~aurora \rightarrow american\_imp...$ & 73.75 & \textbf{83.18} & 77.62 & 83.13 \\
$DOGS~~autumn\_leave... \rightarrow abstract\_ill...$ & 69.51 & 80.16 & 78.75 & \textbf{81.06} \\
$DOGS~~chinese\_art \rightarrow land\_art$ & 72.90 & 79.45 & \textbf{79.64} & 79.12 \\
$DOGS~~dawn \rightarrow abstract\_ill...$ & 65.50 & 82.47 & 82.05 & \textbf{85.63} \\
$DOGS~~dawn \rightarrow american\_bar...$ & 74.60 & \textbf{83.46} & 82.89 & 82.14 \\
$DOGS~~fauvism\_pain... \rightarrow abstract\_ill...$ & 75.07 & \textbf{83.84} & 83.65 & 83.55 \\
$DOGS~~fauvism\_pain... \rightarrow rayonism$ & 68.61 & 79.45 & 81.34 & \textbf{84.07} \\
$DOGS~~fauvism\_pain... \rightarrow swamp$ & 62.77 & 79.69 & 78.28 & \textbf{83.03} \\
$DOGS~~hurricane \rightarrow aurora$ & 67.44 & 79.12 & \textbf{82.85} & 82.14 \\
$DOGS~~hurricane \rightarrow autumn\_leave...$ & 53.35 & \textbf{83.51} & 78.93 & 82.52 \\
$DOGS~~hurricane \rightarrow cloud$ & 58.29 & 73.28 & 72.10 & \textbf{75.54} \\
$DOGS~~hurricane \rightarrow thunder$ & 50.19 & 67.95 & 66.35 & \textbf{72.57} \\
$DOGS~~land\_art \rightarrow abstract\_ill...$ & 74.74 & 83.22 & \textbf{84.97} & 83.08 \\
$DOGS~~land\_art \rightarrow american\_imp...$ & 62.68 & 82.61 & 79.31 & \textbf{85.30} \\
$DOGS~~land\_art \rightarrow cloud$ & 53.16 & \textbf{74.13} & 70.97 & 70.97 \\
$DOGS~~land\_art \rightarrow wave$ & 70.45 & 75.21 & \textbf{81.62} & 79.22 \\
$DOGS~~meadow \rightarrow abstract\_ill...$ & 76.53 & 82.00 & \textbf{82.61} & 81.95 \\
$DOGS~~meadow \rightarrow chinese\_art$ & 80.21 & 83.27 & \textbf{83.79} & 82.52 \\
$DOGS~~meadow \rightarrow prehistorian$ & 73.85 & 80.49 & 82.66 & \textbf{84.59} \\
$DOGS~~prehistorian \rightarrow american\_bar...$ & 75.12 & 82.52 & 82.70 & \textbf{83.60} \\
\hdashline
DOGS mean & 65.23 $\pm$ 2.66 & 79.05 $\pm$ 1.25 & 78.79 $\pm$ 1.23 & \textbf{80.26 $\pm$ 1.12} \\
\hline
\end{tabular}
}
\caption{Closed Set Domain Adaptation Accuracy over DOGS}
\label{tab:acc-DOGS}
\end{table}

\begin{table}[h]
\centering
\resizebox{\linewidth}{!}{
\begin{tabular}{|c|c|c|c|c|}
\hline
\textbf{Scenario} & \textbf{NO\_ADAPT} & \textbf{DeepJDOT} & \textbf{DANN} & \textbf{CDAN}\\
\hline 
$SPORTS~~american\_imp... \rightarrow american\_bar...$ & 61.05 & \textbf{83.03} & 72.61 & 82.57 \\
$SPORTS~~american\_imp... \rightarrow hurricane$ & 45.79 & 60.99 & 61.67 & \textbf{76.20} \\
$SPORTS~~analytical\_a... \rightarrow autumn\_leave...$ & 72.21 & \textbf{85.99} & 84.34 & 81.61 \\
$SPORTS~~analytical\_a... \rightarrow thunder$ & 41.34 & 56.66 & 64.41 & \textbf{69.93} \\
$SPORTS~~arts\_and\_cra... \rightarrow dawn$ & 21.81 & 51.65 & 47.04 & \textbf{56.09} \\
$SPORTS~~aurora \rightarrow american\_imp...$ & 63.95 & 74.94 & 72.72 & \textbf{80.81} \\
$SPORTS~~autumn\_leave... \rightarrow abstract\_ill...$ & 59.57 & 76.88 & \textbf{82.97} & 82.46 \\
$SPORTS~~chinese\_art \rightarrow land\_art$ & 62.59 & 68.96 & \textbf{80.87} & 80.52 \\
$SPORTS~~dawn \rightarrow abstract\_ill...$ & 71.30 & 77.16 & 77.45 & \textbf{85.48} \\
$SPORTS~~dawn \rightarrow american\_bar...$ & 76.59 & 58.14 & \textbf{83.77} & 81.61 \\
$SPORTS~~fauvism\_pain... \rightarrow abstract\_ill...$ & 66.74 & \textbf{88.33} & 87.64 & 82.00 \\
$SPORTS~~fauvism\_pain... \rightarrow rayonism$ & 56.83 & \textbf{80.69} & 75.63 & 79.50 \\
$SPORTS~~fauvism\_pain... \rightarrow swamp$ & 44.25 & 66.46 & \textbf{80.35} & 77.16 \\
$SPORTS~~hurricane \rightarrow aurora$ & 46.58 & 75.40 & 73.01 & \textbf{82.35} \\
$SPORTS~~hurricane \rightarrow autumn\_leave...$ & 48.63 & 73.97 & 79.33 & \textbf{83.26} \\
$SPORTS~~hurricane \rightarrow cloud$ & 48.35 & 63.61 & 75.00 & \textbf{78.02} \\
$SPORTS~~hurricane \rightarrow thunder$ & 42.71 & 63.21 & \textbf{70.67} & 70.56 \\
$SPORTS~~land\_art \rightarrow abstract\_ill...$ & 71.41 & 85.19 & 84.11 & \textbf{85.76} \\
$SPORTS~~land\_art \rightarrow american\_imp...$ & 59.40 & 74.77 & 66.46 & \textbf{80.24} \\
$SPORTS~~land\_art \rightarrow cloud$ & 55.13 & 58.66 & 61.73 & \textbf{67.31} \\
$SPORTS~~land\_art \rightarrow wave$ & 67.65 & 77.33 & \textbf{83.88} & 77.16 \\
$SPORTS~~meadow \rightarrow abstract\_ill...$ & 66.46 & \textbf{84.34} & 80.01 & 79.61 \\
$SPORTS~~meadow \rightarrow chinese\_art$ & 69.65 & \textbf{84.62} & 75.40 & 81.55 \\
$SPORTS~~meadow \rightarrow prehistorian$ & 60.88 & 73.58 & 71.98 & \textbf{79.16} \\
$SPORTS~~prehistorian \rightarrow american\_bar...$ & \textbf{78.87} & 71.92 & 76.59 & 78.70 \\
\hdashline
SPORTS mean & 58.39 $\pm$ 2.67 & 72.66 $\pm$ 2.07 & 74.79 $\pm$ 1.84 & \textbf{78.38 $\pm$ 1.29} \\
\hline
\end{tabular}
}
\caption{Closed Set Domain Adaptation Accuracy over SPORTS}
\label{tab:acc-SPORTS}
\end{table}

\begin{table}[h]
\centering
\resizebox{\linewidth}{!}{
\begin{tabular}{|c|c|c|c|c|}
\hline
\textbf{Scenario} & \textbf{NO\_ADAPT} & \textbf{DeepJDOT} & \textbf{DANN} & \textbf{CDAN}\\
\hline 
$PLT\_DOC~~american\_imp... \rightarrow american\_bar...$ & 38.11 & \textbf{50.84} & 46.44 & 48.31 \\
$PLT\_DOC~~american\_imp... \rightarrow hurricane$ & 27.43 & 41.01 & \textbf{41.85} & 40.07 \\
$PLT\_DOC~~analytical\_a... \rightarrow autumn\_leave...$ & 42.60 & 49.16 & \textbf{50.75} & 48.03 \\
$PLT\_DOC~~analytical\_a... \rightarrow thunder$ & 16.10 & \textbf{31.46} & 31.09 & 31.37 \\
$PLT\_DOC~~arts\_and\_cra... \rightarrow dawn$ & 13.58 & 21.44 & \textbf{26.40} & 19.38 \\
$PLT\_DOC~~aurora \rightarrow american\_imp...$ & 30.24 & 37.55 & \textbf{40.36} & 38.95 \\
$PLT\_DOC~~autumn\_leave... \rightarrow abstract\_ill...$ & 36.14 & \textbf{49.44} & 48.97 & 47.10 \\
$PLT\_DOC~~chinese\_art \rightarrow land\_art$ & 35.58 & 38.48 & 38.48 & \textbf{40.07} \\
$PLT\_DOC~~dawn \rightarrow abstract\_ill...$ & 31.84 & 45.13 & 46.07 & \textbf{47.10} \\
$PLT\_DOC~~dawn \rightarrow american\_bar...$ & 37.45 & 44.01 & 47.38 & \textbf{49.06} \\
$PLT\_DOC~~fauvism\_pain... \rightarrow abstract\_ill...$ & 42.51 & \textbf{53.75} & 53.56 & 51.50 \\
$PLT\_DOC~~fauvism\_pain... \rightarrow rayonism$ & 34.64 & 43.45 & \textbf{47.85} & 44.19 \\
$PLT\_DOC~~fauvism\_pain... \rightarrow swamp$ & 33.99 & 46.16 & \textbf{47.75} & 46.16 \\
$PLT\_DOC~~hurricane \rightarrow aurora$ & 27.62 & 38.67 & 42.60 & \textbf{44.10} \\
$PLT\_DOC~~hurricane \rightarrow autumn\_leave...$ & 20.04 & 42.32 & \textbf{44.10} & 42.98 \\
$PLT\_DOC~~hurricane \rightarrow cloud$ & 23.60 & 38.39 & 38.30 & \textbf{39.23} \\
$PLT\_DOC~~hurricane \rightarrow thunder$ & 20.97 & 29.21 & 32.21 & \textbf{36.24} \\
$PLT\_DOC~~land\_art \rightarrow abstract\_ill...$ & 39.04 & \textbf{50.75} & 47.57 & 48.31 \\
$PLT\_DOC~~land\_art \rightarrow american\_imp...$ & 25.37 & 43.45 & \textbf{44.76} & 44.01 \\
$PLT\_DOC~~land\_art \rightarrow cloud$ & 28.84 & 34.55 & \textbf{34.93} & 34.46 \\
$PLT\_DOC~~land\_art \rightarrow wave$ & 40.82 & \textbf{45.32} & 44.19 & 42.32 \\
$PLT\_DOC~~meadow \rightarrow abstract\_ill...$ & 27.81 & \textbf{49.06} & 48.03 & 46.25 \\
$PLT\_DOC~~meadow \rightarrow chinese\_art$ & 35.77 & \textbf{48.88} & 45.79 & 46.35 \\
$PLT\_DOC~~meadow \rightarrow prehistorian$ & 36.99 & \textbf{45.97} & 41.85 & 42.23 \\
$PLT\_DOC~~prehistorian \rightarrow american\_bar...$ & 35.67 & 38.01 & 43.82 & \textbf{48.88} \\
\hdashline
PLT\_DOC mean & 31.31 $\pm$ 1.60 & 42.26 $\pm$ 1.52 & \textbf{43.00 $\pm$ 1.29} & 42.67 $\pm$ 1.39 \\
\hline
\end{tabular}
}
\caption{Closed Set Domain Adaptation Accuracy over PLT\_DOC}
\label{tab:acc-PLT-DOC}
\end{table}

\begin{table}[h]
\centering
\resizebox{\linewidth}{!}{
\begin{tabular}{|c|c|c|c|c|}
\hline
\textbf{Scenario} & \textbf{NO\_ADAPT} & \textbf{DeepJDOT} & \textbf{DANN} & \textbf{CDAN}\\
\hline 
$APL~~american\_imp... \rightarrow american\_bar...$ & 72.66 & \textbf{81.29} & 79.99 & 79.80 \\
$APL~~american\_imp... \rightarrow hurricane$ & 55.58 & 75.96 & \textbf{76.25} & 73.78 \\
$APL~~analytical\_a... \rightarrow autumn\_leave...$ & 72.84 & 77.93 & 78.52 & \textbf{82.95} \\
$APL~~analytical\_a... \rightarrow thunder$ & 16.36 & 68.76 & 70.12 & \textbf{71.19} \\
$APL~~arts\_and\_cra... \rightarrow dawn$ & 27.28 & 66.02 & 61.06 & \textbf{70.82} \\
$APL~~aurora \rightarrow american\_imp...$ & 74.00 & 79.19 & 78.46 & \textbf{81.46} \\
$APL~~autumn\_leave... \rightarrow abstract\_ill...$ & 68.90 & 76.41 & 78.60 & \textbf{79.42} \\
$APL~~chinese\_art \rightarrow land\_art$ & 71.00 & 76.90 & 77.35 & \textbf{78.03} \\
$APL~~dawn \rightarrow abstract\_ill...$ & 60.44 & 80.74 & 81.36 & \textbf{81.74} \\
$APL~~dawn \rightarrow american\_bar...$ & 75.80 & 80.27 & \textbf{81.21} & 80.78 \\
$APL~~fauvism\_pain... \rightarrow abstract\_ill...$ & 75.56 & 80.11 & 79.76 & \textbf{81.32} \\
$APL~~fauvism\_pain... \rightarrow rayonism$ & 73.72 & 79.35 & 80.50 & \textbf{81.81} \\
$APL~~fauvism\_pain... \rightarrow swamp$ & 66.41 & 76.43 & 76.86 & \textbf{81.44} \\
$APL~~hurricane \rightarrow aurora$ & 68.00 & 77.27 & 78.29 & \textbf{80.34} \\
$APL~~hurricane \rightarrow autumn\_leave...$ & 59.02 & 79.82 & 79.40 & \textbf{81.54} \\
$APL~~hurricane \rightarrow cloud$ & 72.11 & 77.09 & 77.01 & \textbf{78.99} \\
$APL~~hurricane \rightarrow thunder$ & 62.63 & 74.90 & 73.37 & \textbf{76.35} \\
$APL~~land\_art \rightarrow abstract\_ill...$ & 79.74 & \textbf{81.78} & 80.76 & 81.56 \\
$APL~~land\_art \rightarrow american\_imp...$ & 71.76 & 77.99 & 79.60 & \textbf{80.89} \\
$APL~~land\_art \rightarrow cloud$ & 70.66 & 75.19 & 73.33 & \textbf{76.72} \\
$APL~~land\_art \rightarrow wave$ & 77.31 & 78.31 & 80.31 & \textbf{80.74} \\
$APL~~meadow \rightarrow abstract\_ill...$ & 66.33 & 79.91 & 79.50 & \textbf{80.78} \\
$APL~~meadow \rightarrow chinese\_art$ & 75.13 & 79.35 & 78.95 & \textbf{80.25} \\
$APL~~meadow \rightarrow prehistorian$ & 74.25 & \textbf{80.17} & 79.23 & 79.89 \\
$APL~~prehistorian \rightarrow american\_bar...$ & 75.27 & 79.42 & 79.54 & \textbf{79.87} \\
\hdashline
APL mean & 66.51 $\pm$ 2.96 & 77.62 $\pm$ 0.73 & 77.57 $\pm$ 0.87 & \textbf{79.30 $\pm$ 0.64} \\
\hline
\end{tabular}
}
\caption{Closed Set Domain Adaptation Accuracy over APL}
\label{tab:acc-APL}
\end{table}

%% file: tables/Universal-DA.tex
\begin{table}[h]
\centering
\resizebox{\linewidth}{!}{
\begin{tabular}{|c|c|c|c|c|}
\hline
\textbf{Scenario} & \textbf{UDA} & \textbf{OSBP} & \textbf{OVANet} & \textbf{UniOT}\\
\hline 
$BIRDS~~american\_imp... \rightarrow american\_bar...$ & 37.51 & 56.11 & 69.69 & \textbf{72.13} \\
$BIRDS~~american\_imp... \rightarrow hurricane$ & 30.06 & 53.35 & \textbf{79.41} & 66.73 \\
$BIRDS~~analytical\_a... \rightarrow autumn\_leave...$ & 42.55 & 39.57 & 63.16 & \textbf{69.00} \\
$BIRDS~~analytical\_a... \rightarrow thunder$ & 21.64 & 52.75 & \textbf{81.53} & 64.55 \\
$BIRDS~~arts\_and\_cra... \rightarrow dawn$ & 35.21 & 48.13 & \textbf{68.66} & 62.86 \\
$BIRDS~~aurora \rightarrow american\_imp...$ & 41.35 & 52.73 & \textbf{72.85} & 71.59 \\
$BIRDS~~autumn\_leave... \rightarrow abstract\_ill...$ & 42.03 & 49.17 & \textbf{76.56} & 67.77 \\
$BIRDS~~chinese\_art \rightarrow land\_art$ & 35.64 & 45.11 & \textbf{79.07} & 71.79 \\
$BIRDS~~dawn \rightarrow abstract\_ill...$ & 39.90 & 44.08 & \textbf{70.32} & 68.56 \\
$BIRDS~~dawn \rightarrow american\_bar...$ & 51.16 & 53.83 & \textbf{79.47} & 69.66 \\
$BIRDS~~fauvism\_pain... \rightarrow abstract\_ill...$ & 43.30 & 52.90 & \textbf{77.63} & 67.07 \\
$BIRDS~~fauvism\_pain... \rightarrow rayonism$ & 51.15 & 51.52 & \textbf{77.68} & 73.00 \\
$BIRDS~~fauvism\_pain... \rightarrow swamp$ & 40.02 & 52.53 & \textbf{76.21} & 66.01 \\
$BIRDS~~hurricane \rightarrow aurora$ & 43.66 & 50.12 & 69.71 & \textbf{77.05} \\
$BIRDS~~hurricane \rightarrow autumn\_leave...$ & 39.73 & 57.59 & \textbf{67.00} & 62.95 \\
$BIRDS~~hurricane \rightarrow cloud$ & 27.69 & 49.61 & \textbf{78.50} & 63.73 \\
$BIRDS~~hurricane \rightarrow thunder$ & 33.81 & 55.23 & \textbf{75.46} & 68.72 \\
$BIRDS~~land\_art \rightarrow abstract\_ill...$ & 31.87 & 34.35 & \textbf{75.66} & 67.52 \\
$BIRDS~~land\_art \rightarrow american\_imp...$ & 38.93 & 48.51 & 57.45 & \textbf{67.84} \\
$BIRDS~~land\_art \rightarrow cloud$ & 34.15 & 47.07 & \textbf{76.52} & 64.41 \\
$BIRDS~~land\_art \rightarrow wave$ & 32.17 & 54.14 & \textbf{80.76} & 68.81 \\
$BIRDS~~meadow \rightarrow abstract\_ill...$ & 36.69 & 44.80 & \textbf{66.57} & 58.99 \\
$BIRDS~~meadow \rightarrow chinese\_art$ & 36.49 & 46.92 & 71.15 & \textbf{73.58} \\
$BIRDS~~meadow \rightarrow prehistorian$ & 43.29 & 42.96 & \textbf{74.65} & 67.53 \\
$BIRDS~~prehistorian \rightarrow american\_bar...$ & 45.51 & 49.57 & 66.53 & \textbf{68.65} \\
\hdashline
BIRDS mean & 38.22 $\pm$ 1.36 & 49.31 $\pm$ 1.08 & \textbf{73.29 $\pm$ 1.21} & 68.02 $\pm$ 0.79 \\
\hline
\end{tabular}
}
\caption{Universal Domain Adaptation H-score over BIRDS}
\label{tab:H-score-BIRDS}
\end{table}

\begin{table}[h]
\centering
\resizebox{\linewidth}{!}{
\begin{tabular}{|c|c|c|c|c|}
\hline
\textbf{Scenario} & \textbf{UDA} & \textbf{OSBP} & \textbf{OVANet} & \textbf{UniOT}\\
\hline 
$DOGS~~american\_imp... \rightarrow american\_bar...$ & 47.79 & 58.32 & \textbf{63.10} & 59.71 \\
$DOGS~~american\_imp... \rightarrow hurricane$ & 34.48 & 54.20 & \textbf{55.68} & 51.84 \\
$DOGS~~analytical\_a... \rightarrow autumn\_leave...$ & 33.61 & 54.08 & \textbf{60.23} & 58.94 \\
$DOGS~~analytical\_a... \rightarrow thunder$ & 31.26 & 42.60 & \textbf{54.89} & 52.04 \\
$DOGS~~arts\_and\_cra... \rightarrow dawn$ & 32.61 & 27.18 & \textbf{55.61} & 36.76 \\
$DOGS~~aurora \rightarrow american\_imp...$ & 42.98 & 56.46 & \textbf{58.72} & 55.55 \\
$DOGS~~autumn\_leave... \rightarrow abstract\_ill...$ & 28.84 & 52.82 & \textbf{65.51} & 57.75 \\
$DOGS~~chinese\_art \rightarrow land\_art$ & 31.40 & 48.25 & \textbf{62.07} & 54.51 \\
$DOGS~~dawn \rightarrow abstract\_ill...$ & 38.66 & 47.55 & \textbf{65.82} & 63.73 \\
$DOGS~~dawn \rightarrow american\_bar...$ & 27.08 & 54.08 & 50.72 & \textbf{65.18} \\
$DOGS~~fauvism\_pain... \rightarrow abstract\_ill...$ & 42.70 & 57.31 & \textbf{62.65} & 60.53 \\
$DOGS~~fauvism\_pain... \rightarrow rayonism$ & 38.30 & 52.75 & \textbf{60.53} & 54.39 \\
$DOGS~~fauvism\_pain... \rightarrow swamp$ & 39.09 & 49.73 & \textbf{59.01} & 58.91 \\
$DOGS~~hurricane \rightarrow aurora$ & 33.24 & 57.64 & \textbf{65.21} & 59.20 \\
$DOGS~~hurricane \rightarrow autumn\_leave...$ & 44.20 & 54.33 & \textbf{68.31} & 56.53 \\
$DOGS~~hurricane \rightarrow cloud$ & 39.54 & 48.60 & \textbf{51.83} & 48.67 \\
$DOGS~~hurricane \rightarrow thunder$ & 34.90 & 51.34 & \textbf{59.22} & 48.07 \\
$DOGS~~land\_art \rightarrow abstract\_ill...$ & 36.87 & 52.90 & 56.25 & \textbf{63.88} \\
$DOGS~~land\_art \rightarrow american\_imp...$ & 34.20 & 50.13 & 57.91 & \textbf{60.24} \\
$DOGS~~land\_art \rightarrow cloud$ & 34.29 & 49.17 & \textbf{65.63} & 50.50 \\
$DOGS~~land\_art \rightarrow wave$ & 41.84 & 45.75 & \textbf{64.26} & 57.61 \\
$DOGS~~meadow \rightarrow abstract\_ill...$ & 37.08 & 56.88 & \textbf{68.68} & 57.60 \\
$DOGS~~meadow \rightarrow chinese\_art$ & 46.23 & 56.95 & \textbf{62.84} & 59.59 \\
$DOGS~~meadow \rightarrow prehistorian$ & 52.68 & 50.78 & \textbf{62.20} & 55.90 \\
$DOGS~~prehistorian \rightarrow american\_bar...$ & 38.38 & 59.45 & 57.46 & \textbf{63.93} \\
\hdashline
DOGS mean & 37.69 $\pm$ 1.22 & 51.57 $\pm$ 1.32 & \textbf{60.57 $\pm$ 0.96} & 56.46 $\pm$ 1.24 \\
\hline
\end{tabular}
}
\caption{Universal Domain Adaptation H-score over DOGS}
\label{tab:H-score-DOGS}
\end{table}

\begin{table}[h]
\centering
\resizebox{\linewidth}{!}{
\begin{tabular}{|c|c|c|c|c|}
\hline
\textbf{Scenario} & \textbf{UDA} & \textbf{OSBP} & \textbf{OVANet} & \textbf{UniOT}\\
\hline 
$SPORTS~~american\_imp... \rightarrow american\_bar...$ & 31.66 & 59.14 & \textbf{77.14} & 58.64 \\
$SPORTS~~american\_imp... \rightarrow hurricane$ & 31.75 & 50.74 & \textbf{63.23} & 55.70 \\
$SPORTS~~analytical\_a... \rightarrow autumn\_leave...$ & 37.30 & 59.73 & \textbf{75.99} & 61.60 \\
$SPORTS~~analytical\_a... \rightarrow thunder$ & 22.32 & 43.98 & \textbf{56.90} & 48.30 \\
$SPORTS~~arts\_and\_cra... \rightarrow dawn$ & 28.13 & \textbf{53.22} & 50.91 & 45.24 \\
$SPORTS~~aurora \rightarrow american\_imp...$ & 37.44 & 66.32 & \textbf{74.37} & 64.38 \\
$SPORTS~~autumn\_leave... \rightarrow abstract\_ill...$ & 37.93 & \textbf{62.04} & 59.45 & 59.15 \\
$SPORTS~~chinese\_art \rightarrow land\_art$ & 35.20 & 53.67 & \textbf{72.84} & 56.75 \\
$SPORTS~~dawn \rightarrow abstract\_ill...$ & 38.76 & 60.92 & \textbf{77.60} & 59.97 \\
$SPORTS~~dawn \rightarrow american\_bar...$ & 39.55 & 62.57 & \textbf{79.43} & 62.36 \\
$SPORTS~~fauvism\_pain... \rightarrow abstract\_ill...$ & 40.71 & 59.14 & \textbf{79.89} & 66.81 \\
$SPORTS~~fauvism\_pain... \rightarrow rayonism$ & 39.99 & 61.83 & \textbf{76.61} & 60.26 \\
$SPORTS~~fauvism\_pain... \rightarrow swamp$ & 34.37 & 60.57 & \textbf{75.57} & 56.15 \\
$SPORTS~~hurricane \rightarrow aurora$ & 33.28 & 56.28 & \textbf{75.19} & 57.52 \\
$SPORTS~~hurricane \rightarrow autumn\_leave...$ & 30.00 & 57.34 & \textbf{74.01} & 57.42 \\
$SPORTS~~hurricane \rightarrow cloud$ & 33.71 & 51.99 & \textbf{71.75} & 54.54 \\
$SPORTS~~hurricane \rightarrow thunder$ & 26.00 & 55.91 & \textbf{66.96} & 54.20 \\
$SPORTS~~land\_art \rightarrow abstract\_ill...$ & 29.27 & 58.11 & \textbf{79.83} & 64.82 \\
$SPORTS~~land\_art \rightarrow american\_imp...$ & 30.57 & 52.52 & \textbf{73.87} & 55.62 \\
$SPORTS~~land\_art \rightarrow cloud$ & 39.85 & 50.82 & \textbf{64.39} & 49.39 \\
$SPORTS~~land\_art \rightarrow wave$ & 28.87 & 47.94 & \textbf{74.54} & 58.45 \\
$SPORTS~~meadow \rightarrow abstract\_ill...$ & 39.67 & 60.48 & \textbf{78.41} & 57.21 \\
$SPORTS~~meadow \rightarrow chinese\_art$ & 33.56 & 64.15 & \textbf{72.02} & 55.83 \\
$SPORTS~~meadow \rightarrow prehistorian$ & 32.75 & 62.61 & \textbf{70.96} & 57.00 \\
$SPORTS~~prehistorian \rightarrow american\_bar...$ & 40.70 & 58.60 & 63.35 & \textbf{63.36} \\
\hdashline
SPORTS mean & 34.13 $\pm$ 1.01 & 57.22 $\pm$ 1.09 & \textbf{71.41 $\pm$ 1.54} & 57.63 $\pm$ 1.01 \\
\hline
\end{tabular}
}
\caption{Universal Domain Adaptation H-score over SPORTS}
\label{tab:H-score-SPORTS}
\end{table}

\begin{table}[h]
\centering
\resizebox{\linewidth}{!}{
\begin{tabular}{|c|c|c|c|c|}
\hline
\textbf{Scenario} & \textbf{UDA} & \textbf{OSBP} & \textbf{OVANet} & \textbf{UniOT}\\
\hline 
$PLT\_DOC~~american\_imp... \rightarrow american\_bar...$ & 26.45 & \textbf{44.69} & 32.10 & 38.71 \\
$PLT\_DOC~~american\_imp... \rightarrow hurricane$ & 30.70 & \textbf{36.39} & 25.62 & 30.71 \\
$PLT\_DOC~~analytical\_a... \rightarrow autumn\_leave...$ & 30.38 & 42.01 & \textbf{42.71} & 42.14 \\
$PLT\_DOC~~analytical\_a... \rightarrow thunder$ & 25.67 & \textbf{35.57} & 20.84 & 24.91 \\
$PLT\_DOC~~arts\_and\_cra... \rightarrow dawn$ & 18.68 & 23.83 & 22.05 & \textbf{24.18} \\
$PLT\_DOC~~aurora \rightarrow american\_imp...$ & 31.22 & \textbf{45.29} & 40.61 & 41.87 \\
$PLT\_DOC~~autumn\_leave... \rightarrow abstract\_ill...$ & 25.12 & \textbf{46.60} & 40.02 & 41.46 \\
$PLT\_DOC~~chinese\_art \rightarrow land\_art$ & 34.15 & 37.79 & \textbf{37.90} & 36.40 \\
$PLT\_DOC~~dawn \rightarrow abstract\_ill...$ & 26.24 & 42.64 & 33.02 & \textbf{44.24} \\
$PLT\_DOC~~dawn \rightarrow american\_bar...$ & 30.71 & 39.76 & 38.35 & \textbf{41.24} \\
$PLT\_DOC~~fauvism\_pain... \rightarrow abstract\_ill...$ & 31.99 & 41.33 & 38.81 & \textbf{46.83} \\
$PLT\_DOC~~fauvism\_pain... \rightarrow rayonism$ & 36.92 & 42.48 & 39.37 & \textbf{47.01} \\
$PLT\_DOC~~fauvism\_pain... \rightarrow swamp$ & 30.64 & 40.46 & 37.79 & \textbf{45.27} \\
$PLT\_DOC~~hurricane \rightarrow aurora$ & 31.02 & 38.32 & 27.43 & \textbf{39.27} \\
$PLT\_DOC~~hurricane \rightarrow autumn\_leave...$ & 34.27 & 32.83 & 25.06 & \textbf{43.45} \\
$PLT\_DOC~~hurricane \rightarrow cloud$ & 22.51 & \textbf{43.07} & 21.61 & 34.68 \\
$PLT\_DOC~~hurricane \rightarrow thunder$ & 18.65 & \textbf{31.06} & 22.19 & 26.10 \\
$PLT\_DOC~~land\_art \rightarrow abstract\_ill...$ & 28.29 & 44.18 & 41.83 & \textbf{46.75} \\
$PLT\_DOC~~land\_art \rightarrow american\_imp...$ & 22.43 & \textbf{42.35} & 38.55 & 39.58 \\
$PLT\_DOC~~land\_art \rightarrow cloud$ & 23.11 & 35.20 & 21.34 & \textbf{37.67} \\
$PLT\_DOC~~land\_art \rightarrow wave$ & 26.22 & \textbf{37.15} & 34.39 & 36.25 \\
$PLT\_DOC~~meadow \rightarrow abstract\_ill...$ & 31.16 & \textbf{51.64} & 32.63 & 44.47 \\
$PLT\_DOC~~meadow \rightarrow chinese\_art$ & 33.46 & \textbf{41.87} & 33.24 & 37.39 \\
$PLT\_DOC~~meadow \rightarrow prehistorian$ & 27.64 & 38.25 & 34.25 & \textbf{39.46} \\
$PLT\_DOC~~prehistorian \rightarrow american\_bar...$ & 30.27 & 38.92 & 42.26 & \textbf{49.30} \\
\hdashline
PLT\_DOC mean & 28.32 $\pm$ 0.95 & \textbf{39.75 $\pm$ 1.12} & 32.96 $\pm$ 1.50 & 39.17 $\pm$ 1.37 \\
\hline
\end{tabular}
}
\caption{Universal Domain Adaptation H-score over PLT\_DOC}
\label{tab:H-score-PLT-DOC}
\end{table}

\begin{table}[h]
\centering
\resizebox{\linewidth}{!}{
\begin{tabular}{|c|c|c|c|c|}
\hline
\textbf{Scenario} & \textbf{UDA} & \textbf{OSBP} & \textbf{OVANet} & \textbf{UniOT}\\
\hline 
$APL~~american\_imp... \rightarrow american\_bar...$ & 41.65 & 26.07 & 47.83 & \textbf{57.07} \\
$APL~~american\_imp... \rightarrow hurricane$ & 30.99 & 26.31 & 44.93 & \textbf{52.43} \\
$APL~~analytical\_a... \rightarrow autumn\_leave...$ & 38.34 & 22.25 & 41.67 & \textbf{60.48} \\
$APL~~analytical\_a... \rightarrow thunder$ & 24.64 & 28.74 & 43.42 & \textbf{46.75} \\
$APL~~arts\_and\_cra... \rightarrow dawn$ & 24.27 & 31.34 & \textbf{55.00} & 42.25 \\
$APL~~aurora \rightarrow american\_imp...$ & 32.52 & 17.12 & 34.57 & \textbf{45.04} \\
$APL~~autumn\_leave... \rightarrow abstract\_ill...$ & 28.26 & 18.14 & 40.47 & \textbf{59.80} \\
$APL~~chinese\_art \rightarrow land\_art$ & 40.20 & 29.57 & 44.81 & \textbf{48.17} \\
$APL~~dawn \rightarrow abstract\_ill...$ & 37.69 & 22.81 & 33.12 & \textbf{55.81} \\
$APL~~dawn \rightarrow american\_bar...$ & 37.39 & 19.78 & 39.97 & \textbf{58.44} \\
$APL~~fauvism\_pain... \rightarrow abstract\_ill...$ & 25.78 & 22.62 & 40.04 & \textbf{60.12} \\
$APL~~fauvism\_pain... \rightarrow rayonism$ & 36.90 & 21.73 & 46.08 & \textbf{50.04} \\
$APL~~fauvism\_pain... \rightarrow swamp$ & 41.16 & 19.15 & 34.96 & \textbf{63.84} \\
$APL~~hurricane \rightarrow aurora$ & 28.47 & 22.96 & 45.53 & \textbf{50.24} \\
$APL~~hurricane \rightarrow autumn\_leave...$ & 42.13 & 17.31 & 38.08 & \textbf{44.82} \\
$APL~~hurricane \rightarrow cloud$ & 35.10 & 14.80 & \textbf{48.24} & 44.84 \\
$APL~~hurricane \rightarrow thunder$ & 27.69 & 28.91 & 42.90 & \textbf{46.75} \\
$APL~~land\_art \rightarrow abstract\_ill...$ & 37.31 & 31.86 & 39.31 & \textbf{54.46} \\
$APL~~land\_art \rightarrow american\_imp...$ & 25.13 & 14.27 & 34.98 & \textbf{48.07} \\
$APL~~land\_art \rightarrow cloud$ & 31.51 & 20.07 & 35.27 & \textbf{60.12} \\
$APL~~land\_art \rightarrow wave$ & 36.08 & 13.42 & 41.58 & \textbf{57.45} \\
$APL~~meadow \rightarrow abstract\_ill...$ & 36.58 & 27.25 & 38.59 & \textbf{55.85} \\
$APL~~meadow \rightarrow chinese\_art$ & 48.23 & 21.61 & 36.05 & \textbf{56.77} \\
$APL~~meadow \rightarrow prehistorian$ & 47.40 & 28.92 & 36.48 & \textbf{55.82} \\
$APL~~prehistorian \rightarrow american\_bar...$ & 49.72 & 25.32 & 41.71 & \textbf{67.01} \\
\hdashline
APL mean & 35.40 $\pm$ 1.48 & 22.89 $\pm$ 1.08 & 41.02 $\pm$ 1.05 & \textbf{53.70 $\pm$ 1.33} \\
\hline
\end{tabular}
}
\caption{Universal Domain Adaptation H-score over APL}
\label{tab:H-score-APL}
\end{table}

%% file: appendices/license.tex
\section{Stylized Meta-Album License}
\label{appendix:license}

All datasets in Stylized Meta-Album are curated from Meta-Album meta-dataset. Meta-Album is distributed with \href{https://creativecommons.org/licenses/by-nc/4.0/}{CC BY-NC 4.0} license. We are releasing Stylized Meta-Album (12 content datasets and 12 stylized datasets) with the same license as Meta-Album.
\\

\noindent CC BY-NC license allows to freely:
\begin{itemize}
    \item {\bf Share} — copy and redistribute the material in any medium or format
    \item {\bf Adapt} — remix, transform, and build upon the material
\end{itemize}

Under the terms:
\begin{itemize}
    \item {\bf Attribution} — You must give appropriate credit, provide a link to the license, and indicate if changes were made. You may do so in any reasonable manner, but not in any way that suggests the licensor endorses you or your use.
    \item {\bf NonCommercial} — You may not use the material for commercial purposes.
\end{itemize}

\noindent More details about this license can be found here: \url{https://creativecommons.org/licenses/by-nc/4.0/}

%% file: appendices/links.tex
\section*{Important Links}
\label{appendix:links}

Stylized Meta-Album Website:\\\url{https://stylized-meta-album.github.io/}
\\~\\
Stylized Meta-Album GitHub repository:\\\url{https://github.com/ihsaan-ullah/stylized-meta-album}
\\~\\

\subsection*{Contact}
For any query about the Stylized Meta-Album meta-dataset, reach us by email\\
\href{mailto:stylized-meta-album@chalearn.org}{stylized-meta-album@chalearn.org}. 